\documentclass[journal=mamobx, manuscript=article]{achemso}
 
\SectionNumbersOn
 
\usepackage[T1]{fontenc}
\usepackage{palatino}
 
\usepackage{geometry}
\usepackage{setspace}
\usepackage{xr}
 
\makeatletter
\newcommand*{\addFileDependency}[1]{
    \typeout{(#1)}
    \@addtofilelist{#1}
    \IfFileExists{#1}{}{\typeout{No file #1.}}
}
\makeatother
 
\newcommand*{\myexternaldocument}[1]{%
    \externaldocument{#1}%
    \addFileDependency{#1.tex}%
    \addFileDependency{#1.aux}%
}
\myexternaldocument{si}
 
\usepackage{fancyhdr}
\usepackage{amsfonts, amssymb, mathtools, bm, bbm}
\usepackage[version=4]{mhchem}
\usepackage{textgreek}
\usepackage{stackengine}
\usepackage{siunitx}
\usepackage{multirow}
\usepackage{enumitem}
 
\usepackage{booktabs}
\usepackage{tabularx}

\usepackage{graphicx}
\usepackage{afterpage}
 
\usepackage{xcolor, pagecolor}
\definecolor{myadded}{HTML}{2495C1} 
\definecolor{mydeleted}{HTML}{E4256D}
\definecolor{myreplaced}{HTML}{F5A623}
\usepackage{changes}
\usepackage{ulem}
\definechangesauthor[name=Bruce, color=myadded]{bruce}
\setaddedmarkup{\textcolor{myadded}{{#1}}}
\setdeletedmarkup{\textcolor{mydeleted}{\sout{#1}}}
 
\usepackage[colorlinks=true, linkcolor=blue, citecolor=blue, urlcolor=blue]{hyperref}
 
\usepackage{nameref, zref-xr}
\zxrsetup{toltxlabel}
\zexternaldocument{si}
 
\usepackage{algorithm}
\usepackage{algpseudocode}
\usepackage{dsfont}
\usepackage{amsthm}
 
\usepackage[symbol]{footmisc}
\usepackage{tcolorbox}
\usepackage{verbatim}
 
\usepackage{caption}
\usepackage{microtype}
\newcommand{\note}[1]{}

\author{Shengli Jiang}
\author{Jason Wu}
\author{Charles M. Schroeder}
\author{Michael A. Webb}
\email{mawebb@princeton.edu}
\affiliation{Department of Chemical and Biological Engineering, Princeton University, Princeton, NJ 08540, USA}
 
\title{Range-Aware Bayesian Optimization for Discovering Diverse Designs within Target Property Windows}


\begin{document}

\begin{abstract}
In many materials and product design problems, desirable candidates exhibit properties that fall within an acceptable range rather than achieve a single optimum. Recovering multiple, distinct solutions that satisfy such specifications is also practically valuable, as some candidates may be preferred for reasons of cost, processability, or robustness that are difficult to encode directly in an objective function.
Here, we develop a range-aware Bayesian optimization (BO) framework in which the acquisition function directly scores the posterior probability that a candidate satisfies a target range.
The framework naturally extends to parallel pursuit of multiple distinct specifications over a shared candidate space.
Across benchmark tasks, range-aware acquisition consistently recovers larger and more diverse sets of valid designs than standard BO baselines and recent goal-seeking methods.
Its utility is further demonstrated in two practically motivated design case studies involving optimizing reaction conditions for polymer synthesis and sequence-defined oligomer discovery for prescribed optical absorption bands, supported by quantum chemical calculations.
These results suggest that range-aware BO can provide a practical and sample-efficient foundation for specification-driven design, particularly when design flexibility and solution diversity are important considerations.
\end{abstract}

\section{Introduction}

Many materials and product design tasks are governed by specification satisfaction rather than single-property optimization or constraint.
In these settings, a design is valid so long as its properties lie within prescribed ranges~\cite{chitturi2024targeted,tian2025materials}.
We refer to these acceptable property intervals as ``target ranges.''
For example, polymers used in flexible electronics and aerospace composites are often designed around application-specific glass-transition windows~\cite{qian2019glass,kim2019active}; photovoltaic absorbers perform best when their band gaps are near values suited for spectral matching~\cite{yu2012identification,kim2020upper,eperon2016perovskite}; and lubricants are viable within suitable viscosity ranges~\cite{bhushan2013principles,chen2024investigation}.
This range-based perspective is also common in broader design practice, including Ashby-style materials selection, where property ranges guide candidate screening before objective-based ranking~\cite{ashby1993materials}; pharmaceutical quality-by-design, where critical quality attributes are specified by acceptable ranges~\cite{schofield2015critical}; and product development, where needs are translated into target specifications with ideal and marginally acceptable values~\cite{ulrich2020product}.
Because several candidates may satisfy the same prescribed range yet differ in synthesis, cost, robustness, scalability, or processability, the utility of design campaigns can be enhanced should they return a portfolio of valid options rather than a single solution~\cite{ashby2004selection,zeni2025generative,renz2024diverse}.
Thus, the relevant search problem is often not to optimize a single property but to recover, under limited evaluation budgets, distinct candidates whose properties fall within prescribed ranges.

Bayesian optimization (BO) is well suited to materials design because it enables sample-efficient exploration through probabilistic surrogate modeling~\cite{mockus1974bayesian,jones1998efficient,snoek2012practical}.
It has been applied successfully to reaction screening~\cite{shields2021bayesian,taylor2023accelerated}, alloy formulation~\cite{pedersen2021bayesian,khatamsaz2023bayesian}, and polymer or protein design~\cite{an2024active,wilding2025integrating,dalal2024polymer,jiang2025generative}.
Standard BO acquisition functions, including Expected Improvement (EI)~\cite{jones1998efficient}, Probability of Improvement (PI)~\cite{kushner1964method}, and confidence-bound methods~\cite{srinivas2010gaussian}, are primarily designed to identify extrema rather than valid regions defined by specifications.
Multi-objective BO methods~\cite{knowles2006parego,emmerich2011hypervolumebased,daulton2020differentiable} extend this paradigm to Pareto-front discovery but still prioritize frontier points rather than candidates that lie within acceptable property ranges. Even when Pareto points happen to fall within a target range, interior valid designs that are dominated by frontier points are not preferentially recovered, although they may be equally desirable in practice.
Consequently, conventional BO formulations are often misaligned with specification-driven design.

Recent work has extended BO toward goal-seeking and feasibility-aware tasks.
Level-set estimation~\cite{gotovos2013active} and contour-finding methods~\cite{bryan2005active,marques2018contour}, including the Expected Feasibility Function (EFF)~\cite{bichon2008efficient}, prioritize threshold crossings or boundary structure rather than discovery of distinct candidates within prescribed ranges.
Bayesian Algorithm Execution (BAX) and its multi-target extension MultiBAX~\cite{neiswanger2021bayesian,chitturi2024targeted} encode general target-seeking utilities through posterior sampling but can incur substantial computational cost in continuous or high-dimensional domains and may propose points outside the target range under high predictive uncertainty~\cite{tian2025materials}.
Target-constrained approaches such as t-EGO~\cite{tian2025materials} focus sampling around a desired scalar target value but do not naturally address vector-valued specifications or multiple independent target ranges.
Constrained BO formulations~\cite{gardner2014bayesian,gelbart2014bayesian} treat feasibility as a multiplier on a separate objective acquisition, subordinating range-satisfaction probability to an external optimization target.
Hierarchical scalarization methods such as Chimera~\cite{hase2018chimera} let the user rank the objectives and assign each a tolerance, then fold them into a single objective function. Because the search is driven by this one aggregate score, such methods pursue the objectives in priority order rather than tracking whether each criterion lies within its target range.
Together, these limitations motivate a BO framework in which range satisfaction is the primary search signal.

In this work, we develop a range-aware BO framework for specification-driven materials design, built around two range-aware, sampling free acquisition functions: the Tolerance Ball (TB) and the Heaviside (HV).
Unlike point-target methods, which preferentially sample near a specified value, or constrained BO approaches, which treat feasibility as a modifier of a separate objective, our framework uses the posterior probability of satisfying a target range as the acquisition objective itself.
Because many design campaigns must pursue multiple specifications under a shared experimental or computational budget~\cite{attia2020closed,kusne2020fly,szymanski2023autonomous,ren2018accelerated,tamasi2022machine,an2024active,jiang2025generative}, including different product grades or operating windows within the same design space, our framework naturally extends to parallel target search across multiple target ranges.
We compare the proposed acquisition functions against existing goal-seeking and target-driven methods based on their ability to recover diverse valid designs across target ranges of varying width and difficulty.
Finally, beyond idealized benchmarks, we demonstrate the framework in two practical design settings. These include polymerization reaction condition design using kinetic Monte Carlo simulations and sequence-defined oligomer design for prescribed optical absorption bands supported by quantum chemical calculations.

\section{Methods}

\subsection{Problem Formulation}

\subsubsection{Single-Target Inverse Design}
We consider the problem of identifying designs $\mathbf{x} \in \mathcal{X} \subset \mathbb{R}^M$ whose $K$-dimensional property vector lies within a prescribed Euclidean tolerance of the target $\mathbf{y}^{\mathrm{tgt}} \in \mathbb{R}^K$. Here, $\mathcal{X}$ denotes the design space, $\mathbb{R}^M$ is the $M$-dimensional real-valued space of possible designs, and $f \colon \mathcal{X} \to \mathbb{R}^K$ denotes the property map from a design to its predicted properties.
A design $\mathbf{x}$ is considered valid if
\begin{equation}
    \|f(\mathbf{x}) - \mathbf{y}^{\mathrm{tgt}}\|_2^2 \le \varepsilon^2,
    \label{eq:valid_solution}
\end{equation}
where $\|\cdot\|_2$ denotes the Euclidean norm and $\varepsilon > 0$ is the tolerance radius.
When target properties differ in scale or admissible tolerance radius, each output dimension is normalized before applying Eq.~\ref{eq:valid_solution}.
We use the Euclidean norm in the normalized output space because, after normalization, it treats deviations across output dimensions symmetrically and provides a simple measure of overall mismatch from the target.
For a fixed target, we denote by $\mathcal{S} = \{\mathbf{x}_1, \mathbf{x}_2, \dots, \mathbf{x}_N\}$ the set of valid designs identified during the search, where $N = |\mathcal{S}|$ is the cardinality of $\mathcal{S}$ (i.e., the number of members of the set). We denote by $n_{\mathrm{eval}}$ the total number of function evaluations allocated to that target.

\subsubsection{Parallel Search for Multiple Targets}
We extend the single-target formulation to a collection of $T$ targets, $\mathcal{Y}^{\mathrm{tgt}} = \{\mathbf{y}_1^{\mathrm{tgt}}, \dots, \mathbf{y}_T^{\mathrm{tgt}}\}$.
For each target $t \in \{1,\dots,T\}$, we then define the corresponding valid set by
\begin{equation}
    \mathcal{S}_t = \left\{ \mathbf{x} \in \mathcal{X} : \|f(\mathbf{x}) - \mathbf{y}_t^{\mathrm{tgt}}\|_2^2 \le \varepsilon^2 \right\},
\end{equation}
and let $N_t = |\mathcal{S}_t|$ denote the number of valid designs identified for target $t$. Throughout and for convenience, we use a common tolerance $\varepsilon$ for all targets, although the formulation readily accommodates target-specific tolerances.

\subsection{Performance Metrics}

We quantify search performance by the diversity of valid designs discovered per function evaluation. Diversity is used as an evaluation metric only, not as an explicit term in the acquisition functions~\cite{konakovic2020diversity,maus2022discovering}.

\subsubsection{Continuous Design Spaces}
For continuous design spaces, diversity is measured using $\delta$-uniqueness. A subset $\mathcal{S}_{\delta} \subseteq \mathcal{S}$ is $\delta$-unique if
\begin{equation}
    \|\mathbf{x}_i - \mathbf{x}_j\|_2^2 > \delta^2
    \label{eq:delta_unique}
\end{equation}
for all distinct $\mathbf{x}_i, \mathbf{x}_j \in \mathcal{S}_{\delta}$.

Let $N_{\mathrm{u}}(\delta)$ denote the maximum cardinality of a $\delta$-unique subset of $\mathcal{S}$.
We define the raw diversity score as the area under the uniqueness curve,
\begin{equation}
    A_{\mathrm{u}} = \int_0^{\delta_{\max}} N_{\mathrm{u}}(\delta)\, d\delta,
    \label{eq:auc_metric}
\end{equation}
where $\delta_{\max}$ is fixed at $0.1\phi$, with $\phi$ denoting the diameter of the design space. For the normalized design space in which each of the $M$ design variables ranges from 0 to 1, $\phi = \sqrt{M}$.
The normalized diversity score for continuous design spaces is then defined as
\begin{equation}
    D_\mathrm{c} = \frac{A_{\mathrm{u}}}{n_{\mathrm{eval}}\,\delta_{\max}}.
\end{equation}
When performance is aggregated across multiple target tolerance ranges, we report the geometric mean of the corresponding $D_\mathrm{c}$ values.

\subsubsection{Discrete Design Spaces}

For discrete design spaces, the raw diversity is the number of distinct valid designs found, $N = |\mathcal{S}|$. Let $N_{\mathrm{valid}}^{\mathrm{all}}$ denote the total number of valid designs in the full discrete candidate set. Because each function evaluation can identify at most one previously unseen valid design, the maximum attainable number of valid discoveries after $n_{\mathrm{eval}}$ evaluations is
\begin{equation}
    N_{\max} = \min \left( n_{\mathrm{eval}},\, N_{\mathrm{valid}}^{\mathrm{all}} \right).
\end{equation}
If $N_{\mathrm{valid}}^{\mathrm{all}}$ is unknown, we set $N_{\max} = n_{\mathrm{eval}}$. We then define the normalized diversity score for discrete design spaces as
\begin{equation}
    D_\mathrm{d} = \frac{N}{N_{\max}}.
\end{equation}

\subsection{Gaussian Process Surrogate Model}

We model the mapping from candidates to properties using Gaussian process (GP) regression~\cite{williams2006gaussian}. For a $K$-dimensional property vector, we fit one GP independently to each output dimension:
\begin{equation}
    f_k(\mathbf{x}) \sim \mathcal{GP}\bigl(m_k(\mathbf{x}),\, \kappa_k(\mathbf{x},\mathbf{x}')\bigr),
    \qquad k = 1,\dots,K,
\end{equation}
using a zero prior mean, $m_k(\mathbf{x}) \equiv 0$.
Each GP uses a Mat\'ern 5/2 kernel with signal variance $\sigma_k^2$ and length scale $l_k$:
\begin{equation}
    \kappa_k(\mathbf{x},\mathbf{x}') =
    \sigma_k^2
    \left(
    1 + \frac{\sqrt{5}\|\mathbf{x}-\mathbf{x}'\|_2}{l_k}
    + \frac{5\|\mathbf{x}-\mathbf{x}'\|_2^2}{3l_k^2}
    \right)
    \exp\left(
    -\frac{\sqrt{5}\|\mathbf{x}-\mathbf{x}'\|_2}{l_k}
    \right).
\end{equation}
The kernel hyperparameters are re-estimated after each BO iteration by maximizing the marginal likelihood. The Mat\'ern 5/2 kernel is a standard and robust default for Bayesian optimization~\cite{snoek2012practical}; its twice-differentiable sample paths impose weaker smoothness than the squared-exponential (RBF) kernel~\cite{rasmussen2006gaussian} and thereby better accommodate the moderately rough structure--property relationships common in molecular data~\cite{aldeghi2022roughness}.

Given observations $\{(\mathbf{x}_i, y_{ik})\}_{i=1}^{n}$ for output $k$, the posterior predictive mean and variance at a new point $\mathbf{x}_*$ are
\begin{equation}
\begin{aligned}
    \mu_k(\mathbf{x}_*) &= \mathbf{k}_{k,*}^{\top}\left(\mathbf{K}_k + \sigma_{n,k}^2 \mathbf{I}\right)^{-1}\mathbf{y}_k, \\
    s_k^2(\mathbf{x}_*) &= \kappa_k(\mathbf{x}_*, \mathbf{x}_*) -
    \mathbf{k}_{k,*}^{\top}\left(\mathbf{K}_k + \sigma_{n,k}^2 \mathbf{I}\right)^{-1}\mathbf{k}_{k,*},
\end{aligned}
\end{equation}
where $\mathbf{y}_k = [y_{1k},\dots,y_{nk}]^\top$, $\mathbf{k}_{k,*} \in \mathbb{R}^{n}$ has entries
\begin{equation}
    (\mathbf{k}_{k,*})_i = \kappa_k(\mathbf{x}_*, \mathbf{x}_i),
\end{equation}
and $\mathbf{K}_k \in \mathbb{R}^{n \times n}$ has entries
\begin{equation}
    (\mathbf{K}_k)_{ij} = \kappa_k(\mathbf{x}_i,\mathbf{x}_j).
\end{equation}

\subsection{Bayesian Optimization Procedure}

Bayesian optimization is used to sequentially select candidates for evaluation. For a target $\mathbf{y}^{\mathrm{tgt}}$, we define the squared discrepancy
\begin{equation}
    d(\mathbf{x}) = \|f(\mathbf{x}) - \mathbf{y}^{\mathrm{tgt}}\|_2^2
\end{equation}
and the posterior-mean discrepancy
\begin{equation}
    \Delta^2(\mathbf{x}) = \|\boldsymbol{\mu}(\mathbf{x}) - \mathbf{y}^{\mathrm{tgt}}\|_2^2.
\end{equation}

For a single target, the BO loop proceeds as follows. First, the surrogate model is initialized with $n_{\mathrm{init}}$ candidates generated by Latin hypercube sampling (LHS)~\cite{mckay1979comparison}. The GP surrogate is then fit to the observed data, the acquisition function is maximized over $\mathcal{X}$, the selected candidates is evaluated, and the surrogate is updated. This process is repeated until the evaluation budget is exhausted. Unless otherwise noted, we use a total budget of 50 evaluations per run, and duplicate evaluations are not allowed.
 
For parallel target search, all targets are modeled by a single GP surrogate and a shared observation set. Search proceeds as a synchronous batch procedure. At the beginning of each outer iteration, the GP is fit once to all observed data. Each target then proposes one candidate by maximizing a target-specific acquisition function under the shared posterior. The resulting batch of proposed candidates is evaluated, and the GP is updated only after all evaluations in the batch are complete. The posterior therefore remains fixed while the batch is constructed. If two targets propose the same unevaluated candidate, that candidate is evaluated only once and assigned to the earlier target under a fixed ordering. The later target then selects the highest-ranked unevaluated alternative under the same posterior. This duplicate-removal step prevents redundant evaluation while preserving a batch size of $T$ whenever sufficiently many unevaluated candidates remain.

\subsection{Acquisition Functions}
We compare six acquisition strategies: Expected Improvement (EI), Lower Confidence Bound (LCB), the Tolerance Ball (TB) and Heaviside (HV) acquisitions, a posterior-mean BAX baseline (BAX), and random sampling (RS).

\subsubsection{Expected Improvement}
Expected Improvement (EI) \cite{jones1998efficient} favors selecting candidates that are expected to improve upon the smallest discrepancy observed so far, with consideration also of the magnitude of that improvement. To obtain a closed-form expression for vector-valued targets, we adopt the isotropic predictive-variance approximation introduced by Uhrenholt et al.~\cite{uhrenholt2019efficient}, treating the predictive distribution of $\mathbf{f}(\mathbf{x})$ given the current $n$ observations $\mathcal{D}_n = \{(\mathbf{x}_i, \mathbf{y}_i)\}_{i=1}^n$ as:
\begin{equation}
    \mathbf{f}(\mathbf{x}) \mid \mathcal{D}_n
    \approx
    \mathcal{N}\left(
    \boldsymbol{\mu}(\mathbf{x}),
    \eta^2(\mathbf{x}) \mathbf{I}_K
    \right),
\end{equation}
where
\begin{equation}
    \eta^2(\mathbf{x}) = \frac{1}{K}\sum_{k=1}^{K} s_k^2(\mathbf{x}).
\end{equation}
Under this approximation,
\begin{equation}
    \frac{d(\mathbf{x})}{\eta^2(\mathbf{x})}
    \sim
    \chi^2_K\bigl(\lambda(\mathbf{x})\bigr),
\end{equation}
with noncentrality parameter
\begin{equation}
    \lambda(\mathbf{x}) = \frac{\Delta^2(\mathbf{x})}{\eta^2(\mathbf{x})},
\end{equation}
where $d(\mathbf{x})$ is the squared discrepancy and $\Delta^2(\mathbf{x})$ is the posterior-mean squared discrepancy defined above. We then let
\begin{equation}
    d_{\min} = \min_{i=1,\dots,n} d(\mathbf{x}_i)
\end{equation}
denote the smallest observed discrepancy, and define
\begin{equation}
    r = \frac{d_{\min}}{\eta^2(\mathbf{x})}.
\end{equation}
The EI acquisition is then
\begin{equation}
    A_{\mathrm{EI}}(\mathbf{x}) =
    d_{\min}\, F_{K,\lambda}(r)
    -
    \eta^2(\mathbf{x})
    \left[
    K\,F_{K+2,\lambda}(r)
    +
    \lambda(\mathbf{x})\,F_{K+4,\lambda}(r)
    \right],
\end{equation}
where $F_{K,\lambda}(\cdot)$ denotes the cumulative distribution function of the noncentral $\chi^2$ distribution with $K$ degrees of freedom and noncentrality parameter $\lambda(\mathbf{x})$.

\subsubsection{Lower Confidence Bound}
As a complementary baseline, we use the closed-form lower confidence bound (LCB) acquisition~\cite{uhrenholt2019efficient},
\begin{equation}
    A_{\mathrm{LCB}}(\mathbf{x};\beta)
    =
    -\bigl(\alpha - \beta \rho\bigr)^{1/\ell}
    \bigl(K + \lambda(\mathbf{x})\bigr)\eta^2(\mathbf{x}),
\end{equation}
where $\beta = 1$ controls the exploration--exploitation trade-off. The auxiliary quantities are
\begin{equation}
\begin{aligned}
    \ell &= 1 - \frac{r_1 r_3}{3 r_2^2}, \\
    r_s &= 2^{s-1}(s-1)!\,\bigl(K + s\,\lambda(\mathbf{x})\bigr),
    \qquad s = 1,2,3, \\
    \alpha &= 1 + \ell(\ell-1)\left[
    \frac{r_2}{2 r_1^2}
    -
    (2-\ell)(1-3\ell)\frac{r_2^2}{8 r_1^4}
    \right], \\
    \rho &= \frac{\ell r_2^2}{r_1}
    \left[
    1 - \frac{\ell(1-3\ell)}{4 r_1^2} r_2
    \right].
\end{aligned}
\end{equation}

\subsubsection{Tolerance-Aware Acquisitions}
We consider two tolerance-aware, sampling-free acquisitions computed directly from the GP posterior. The first, \emph{Tolerance Ball} (TB), is the posterior probability that a candidate satisfies the output-space target range. Using the same isotropic predictive-variance approximation as in the EI derivation, we obtain
\begin{equation}
    A_{\mathrm{TB}}(\mathbf{x})
    =
    \mathbb{P}\left[d(\mathbf{x}) \le \varepsilon^2\right]
    =
    F_{K,\lambda(\mathbf{x})}
    \left(
    \frac{\varepsilon^2}{\eta^2(\mathbf{x})}
    \right).
\end{equation}
For $K=1$, this reduces to the probability that the scalar posterior lies within a tolerance interval around the target; for $K>1$, it is the corresponding probability that the posterior vector lies inside a Euclidean ball. This is analogous to probability-of-feasibility criteria widely used in constrained Bayesian optimization \cite{gardner2014bayesian,gelbart2014bayesian,griffiths2020constrained}, with feasibility here defined as membership in the prescribed output-space tolerance region.

The second, \emph{Heaviside} (HV), prioritizes candidates whose posterior mean lies inside the tolerance range while retaining a smooth transition near the boundary. We define
\begin{equation}
    A_{\mathrm{HV}}(\mathbf{x})
    =
    \bigl(1 - w(\mathbf{x})\bigr)
    +
    w(\mathbf{x})\,A_{\mathrm{TB}}(\mathbf{x}),
\end{equation}
where
\begin{equation}
    w(\mathbf{x}) =
    \frac{1}{2}
    \left[
    1 + \tanh\left(
    \frac{\Delta^2(\mathbf{x}) - \varepsilon^2}{\tau}
    \right)
    \right]
\end{equation}
and $\tau = 10^{-3}\varepsilon^2$ is a smoothing parameter that controls the width of the transition region around the tolerance boundary.
Thus, $A_{\mathrm{HV}}(\mathbf{x}) \approx 1$ for candidates whose posterior mean lies well inside the tolerance range $\bigl(\Delta^2(\mathbf{x}) \ll \varepsilon^2\bigr)$, whereas $A_{\mathrm{HV}}(\mathbf{x}) \approx A_{\mathrm{TB}}(\mathbf{x})$ for candidates whose posterior mean lies well outside it $\bigl(\Delta^2(\mathbf{x}) \gg \varepsilon^2\bigr)$. At the boundary, $\Delta^2(\mathbf{x}) = \varepsilon^2$, so $w(\mathbf{x}) = 1/2$ and    $A_{\mathrm{HV}}(\mathbf{x})= \frac{1}{2}(1+A_{\mathrm{TB}}(\mathbf{x}))$.

Notably, TB and HV are pointwise feasibility acquisitions; they do not explicitly penalize similarity to previously discovered valid candidates. Any diversity observed in the discovered set therefore arises implicitly from posterior uncertainty, the geometry of the feasible regions, and duplicate-evaluation exclusion. The approaches would thus be complementary to other strategies to promote solution diversity.

\subsubsection{Posterior-Mean BAX Baseline}

As a baseline for tolerance-constrained design discovery, we include a feasible-set exploration strategy based on a posterior-mean variant of BAX~\cite{neiswanger2021bayesian,chitturi2024targeted}. 
Given that the corresponding feasible set of valid candidates is
\begin{equation}
    \mathcal{T}
    =
    \left\{
    \mathbf{x} \in \mathcal{X} :
    d(\mathbf{x}) \le \varepsilon^2
    \right\},
\end{equation}
we approximate~\cite{neiswanger2021bayesian,chitturi2024targeted} this set by thresholding the GP posterior mean, resulting in
\begin{equation}
    \bar{\mathcal{T}}
    =
    \left\{
    \mathbf{x} \in \mathcal{X} :
    \Delta^2(\mathbf{x}) \le \varepsilon^2
    \right\}.
\end{equation}

For continuous design spaces, $\bar{\mathcal{T}}$ is approximated on a set of 1,000 candidate points sampled uniformly from $\mathcal{X}$ at each BO iteration. For discrete design spaces, the feasible-set approximation is evaluated over the full candidate set.
 The resulting BAX-style acquisition is defined as
\begin{equation}
    A_{\mathrm{BAX}}(\mathbf{x}) =
    \begin{cases}
        \dfrac{1}{K}\sum_{k=1}^{K} s_k(\mathbf{x}),
        & \text{if } \mathbf{x} \in \bar{\mathcal{T}} \setminus \mathcal{X}_{\mathrm{obs}}, \\
        0,
        & \text{otherwise}
    \end{cases}
\end{equation}
where $\mathcal{X}_{\mathrm{obs}}$ is the set of previously evaluated candidates.
In the first logic branch, `$\setminus$' indicates the set-difference operator, such that the expression is evaluated over the set of candidates in $\bar{\mathcal{T}}$ that have not already been evaluated.
If $\bar{\mathcal{T}} \setminus \mathcal{X}_{\mathrm{obs}} = \varnothing$ (i.e., there are no unevaluated candidates in the target set), the acquisition defaults to global uncertainty sampling:
\begin{equation}
    A_{\mathrm{US}}(\mathbf{x})
    =
    \frac{1}{K}\sum_{k=1}^{K} s_k(\mathbf{x}).
\end{equation}

\subsubsection{Random Sampling}
Random sampling selects the next unevaluated candidate at random from the available candidate set with uniform probability. For continuous design spaces, the candidate set consists of 1,000 points drawn with uniform probabilities from $\mathcal{X}$ at each iteration. For discrete design spaces, the candidate set consists of all remaining unevaluated candidates.

\subsection{Datasets}
We evaluated the BO methods on four classes of inverse-design tasks. In each, the goal is to find candidates whose properties fall within a prescribed tolerance of a target specification: (i) synthetic benchmark functions (Branin, Hartmann-3, Ackley-5, Layeb-6) on continuous domains, for which the target is a scalar function value; (ii) pool-based datasets of pre-characterized candidates from nanoparticle synthesis, intrinsically disordered proteins, polymer topology, and small-molecule libraries, for which targets are measured or computed physicochemical properties; (iii) kinetic Monte Carlo (KMC) styrene polymerization, for which the design variables are reaction conditions and the target is the shape of the polymer molecular weight distribution; and (iv) sequence-defined conjugated oligomers, for which the design variables are molecular building-block choices and the target is the shape of the UV--vis absorption band.
In the following, we describe these design tasks and the underlying functions in more detail.

\subsubsection{Synthetic Functions}
\begin{figure}[h!]
    \centering
    \includegraphics[width=\linewidth]{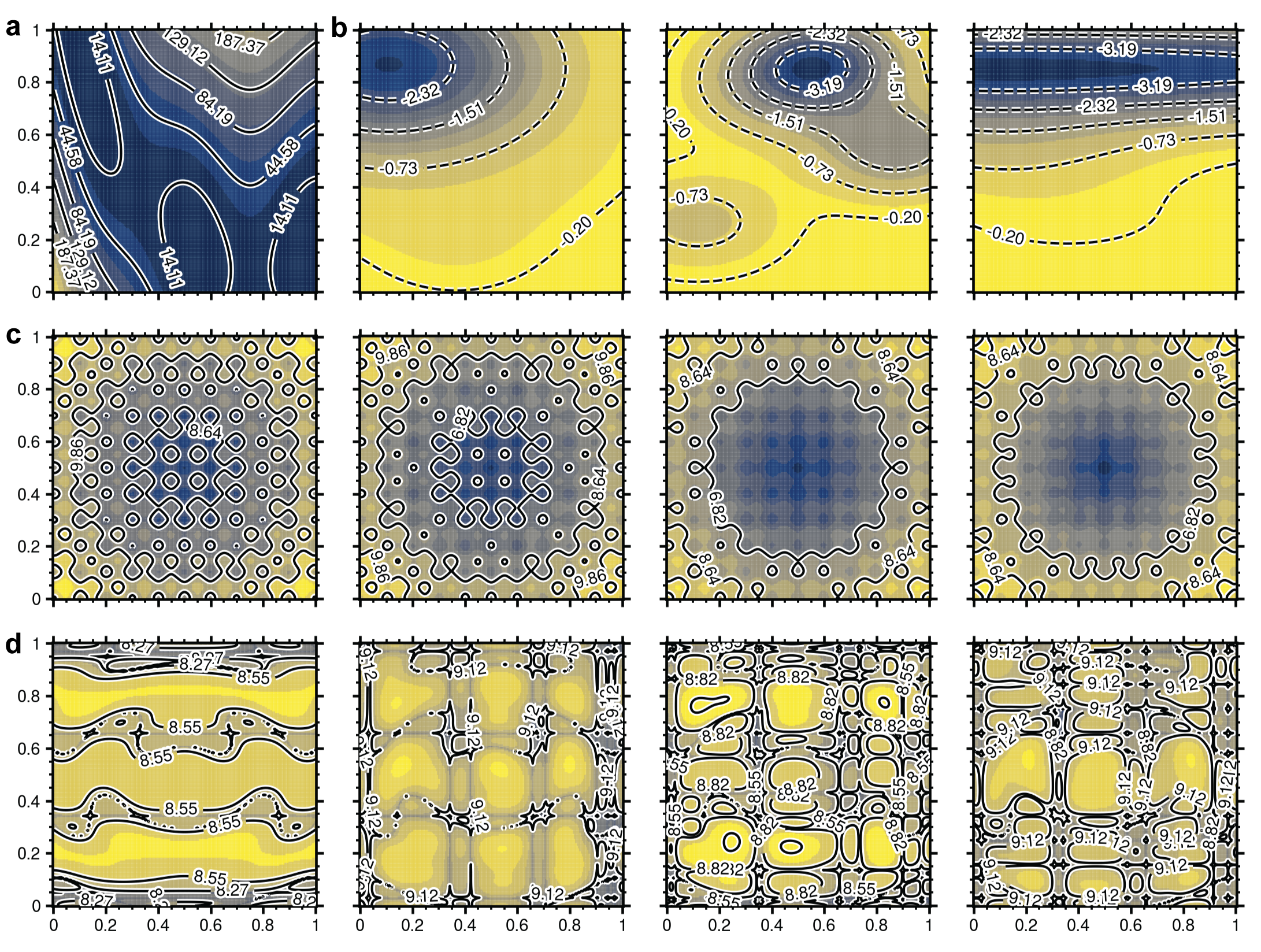}
    \caption{\textbf{Synthetic functions and targets.}
    (a) Branin function ($\mathbb{R}^2 \to \mathbb{R}$).
    (b) Hartmann-3 function ($\mathbb{R}^3 \to \mathbb{R}$), shown as two-dimensional slices in which the first, second, and third coordinates (left to right) are fixed at 0.
    (c) Ackley-5 function ($\mathbb{R}^5 \to \mathbb{R}$), shown as two-dimensional slices over $(x_1, x_2)$ with $x_4 = x_5 = 0$ and $x_3$ fixed at 0, 0.25, 0.5, and 0.75 (left to right).
    (d) Layeb-6 function ($\mathbb{R}^6 \to \mathbb{R}$), shown as two-dimensional slices along randomly selected coordinate pairs, with the remaining coordinates fixed at randomly selected values.
    Contours indicate the target levels in all panels.}
    \label{fig:syn_func}
\end{figure}

We considered four analytic benchmark functions, Branin ($\mathbb{R}^2 \to \mathbb{R}$), Hartmann-3 ($\mathbb{R}^3 \to \mathbb{R}$), Ackley-5 ($\mathbb{R}^5 \to \mathbb{R}$), and Layeb-6 ($\mathbb{R}^6 \to \mathbb{R}$). The analytic expressions and domains of all four functions are given in the SI Section S1. 
Fig.~\ref{fig:syn_func} shows representative two-dimensional slices with the target levels overlaid. 
These benchmark functions span a range of input dimensionalities and landscape characteristics, including differing degrees of multimodality and roughness, and they induce inverse-design tasks with multiple valid solutions that may lie in distinct regions of the design space \cite{jamil2013literature}.

\subsubsection{Pool-Based Datasets}
For pool-based experiments, the design space is a fixed finite candidate set. At each BO iteration, one previously unevaluated candidate is selected from the pool and evaluated. 
The datasets used were nanoparticle synthesis (NS), intrinsically disordered proteins (IDP), TopoRg, BACE, ESOL, FreeSolv, Lipophilicity (Lipo), and QM9. 

SI Table S1 summarizes the number of candidates, the design-space representation, and the target property or properties for each dataset.
The NS dataset comprises 1,997 candidate synthesis conditions, each defined by four normalized design variables, Ti(TEOA)$_2$ concentration, TEOAH$_3$ concentration, pH, and temperature~\cite{pellegrino2020machine}. The target outputs are nanoparticle radius and polydispersity index.
The IDP dataset includes 2,031 polypeptide sequences of length 20--50 residues. Each sequence is characterized by three normalized phase-behavior properties, radius of gyration ($\tilde{R}_g$), second virial coefficient ($\tilde{B}_2$), and expenditure density ($\rho^*$)~\cite{oliver2025b}.
The TopoRg dataset consists of 1,340 candidates represented by eight-dimensional latent vectors obtained from a variational autoencoder trained on graph representations of polymer architectures. The target property is the mean single-chain radius of gyration~\cite{jiang2024property}.
The remaining datasets (BACE, ESOL, FreeSolv, Lipophilicity, and QM9) are molecular libraries with experimentally measured or computationally derived physicochemical properties~\cite{mobley2014freesolv,delaney2004esol,ramakrishnan2014quantum,gallagher2025data,wu2018moleculenet}. The specific target property used for each dataset is listed in Table~S1.

\subsubsection{Kinetic Monte Carlo Styrene Polymerization}

We considered a task in which the objective is to identify reaction conditions that reproduce a prescribed molecular weight distribution (MWD). To this end, we modeled styrene polymerization using a simplified two-stage radical polymerization process inspired by the sequential controlled/conventional strategy reported by Lenzi et al.~\cite{lenzi2005producing}. The first stage is a nitroxide-mediated polymerization (NMP) step, and the second stage is a conventional free-radical polymerization (FRP) step. This model is intended as a stylized batch, bulk, isothermal benchmark that captures the mechanistic contrast between controlled radical growth in stage one and conventional radical propagation in stage two, rather than as an exact reproduction of the heterogeneous reactor configurations used experimentally by Lenzi et al.~\cite{lenzi2005producing,fierens2015mama}.

The complete reaction networks and kinetic parameters for both stages are listed in SI Tables~S2--S4, using literature-based approximations for styrene NMP and FRP kinetics. The KMC scheme follows the stochastic simulation algorithm originally developed by Gillespie
\cite{gillespie1976general,gillespie1977exact}, with the polymerization formulation reported by Romero et al.~\cite{romero2025benchmark}.
Each candidate is represented by a five-dimensional design vector
\begin{equation}
    \mathbf{x} =
    \left(
    X_{\mathrm{NMP}},
    T_{\mathrm{NMP}},
    C_{\mathrm{init,NMP}},
    T_{\mathrm{FRP}},
    C_{\mathrm{init,FRP}}
    \right),
\end{equation}
where $X_{\mathrm{NMP}}$ denotes the prescribed monomer conversion at the end of the NMP stage, $T_{\mathrm{NMP}}$ and $T_{\mathrm{FRP}}$ are the reaction temperatures in the two stages, and $C_{\mathrm{init,NMP}}$ and $C_{\mathrm{init,FRP}}$ are the initiator concentrations in the NMP and FRP stages, respectively. Temperatures are reported in units of Kelvin (K), and initiator concentrations are reported in molar units (mol\,L$^{-1}$). The FRP stage was terminated at an overall conversion of 95\%.

To generate the polymerization design space, we enumerated a full-factorial grid over these five variables. The prescribed NMP conversion $X_{\mathrm{NMP}}$ was varied from 0.4 to 0.8 using 8 uniformly spaced values. The NMP temperature $T_{\mathrm{NMP}}$ was varied from 353 to 403 K using 10 uniformly spaced values, and the FRP temperature $T_{\mathrm{FRP}}$ was varied from 313 to 363 K using 10 uniformly spaced values. The initial monomer concentration was 8.56 mol\,L$^{-1}$, and the KMC simulation was initialized with a control volume of $8.86\times10^{-16}$ L. The initiator concentrations $C_{\mathrm{init,NMP}}$ and $C_{\mathrm{init,FRP}}$ were determined from logarithmically spaced monomer-to-initiator ratios of 50 to 200 for NMP and 500 to 2000 for FRP, respectively, yielding 8 values for each stage. This corresponds to $C_{\mathrm{init,NMP}}$ values from $4.28\times10^{-2}$ to $1.712\times10^{-1}$ mol\,L$^{-1}$ and $C_{\mathrm{init,FRP}}$ values from $4.28\times10^{-3}$ to $1.712\times10^{-2}$ mol\,L$^{-1}$. The resulting full-factorial grid comprised 51,200 unique reaction conditions.

For a given candidate $\mathbf{x}$, the KMC simulation returns the final MWD in weight-fraction form. We represent this output as a 100-dimensional vector $\mathbf{y}$, where each component $y_b$ is the polymer weight fraction within the $b$-th bin. The molecular weight of a chain with degree of polymerization $n$ is $M = M_\mathrm{M}\,n$, where $M_\mathrm{M} = 104.15$~g\,mol$^{-1}$ is the styrene monomer molar mass (SI Table~S2). The bins are defined on a logarithmic molecular-weight axis spanning $M_\mathrm{M}$ to $10^{5}M_\mathrm{M}$ (approximately $10^{2}$ to $10^{7}$~g\,mol$^{-1}$), with 100 uniformly spaced intervals in log-10 space. Specifically, bin $b$ corresponds to the interval
\begin{equation}
    \log_{10}(M) \in [\xi_b,\xi_{b+1}),
\end{equation}
where $\{\xi_b\}_{b=1}^{101}$ are uniformly spaced between $\log_{10}(M_\mathrm{M})$ and $5 + \log_{10}(M_\mathrm{M})$. The output vector is normalized such that
\begin{equation}
    \sum_{b=1}^{100} y_b = 1.
\end{equation}
This binned weight-fraction output serves as the property vector used for inverse design. 
We treat the 100-bin vector as a fixed numerical descriptor of the MWD rather than as a full probabilistic representation.

\subsubsection{Sequence-Defined Conjugated Oligomers}

We considered a task defined over a library of sequence-defined conjugated oligomers. The library was constructed from 21 synthetically accessible oligomer cores and 100 commercially available pinacol boronic esters that serve as terminal caps. 
Combining each core with each cap yields 2,100 candidate structures; after removing duplicate SMILES, which arise when distinct core--cap pairings are structurally equivalent, 1,980 unique oligomers remain and constitute the optimization library (Fig.~\ref{fig:oligomer}a and SI Fig.~S15).

The objective is the shape of the simulated UV--vis absorption band. For each oligomer we computed a broadened absorption spectrum over 240--500~nm by time-dependent density functional theory (TD-DFT)~\cite{runge1984density,casida1995time} and represented it by a compact, physically interpretable parametrization. Working in photon-energy space (eV), where the electronic bands are well described as Gaussian, we fit each spectrum with $K=5$ Gaussian components,
\begin{equation}
    S(E) = \sum_{k=1}^{K} A_k \exp\!\left[-\frac{(E-\mu_k)^2}{2\sigma^2}\right],
    \label{eq:gaussian_spectrum}
\end{equation}
where $A_k$ and $\mu_k$ are the amplitude and center energy of band $k$, and the width $\sigma = 0.12$~eV is held fixed for all bands and all molecules, matching the broadening used to construct the spectra. Spectra contain between one and five resolved bands, which we encode in a fixed-length representation by ordering the components by descending amplitude. When a spectrum has fewer than five bands, each unused component is assigned zero amplitude ($A_k = 0$) together with a single canonical center energy common to all molecules ($3.35$~eV). Because this canonical center is constant across the library, it has zero variance and is therefore uninformative to the surrogate, and it cancels exactly in the tolerance distance whenever both a target and a candidate lack the corresponding band. Moreover, $S(E)$ is independent of $\mu_k$ when $A_k = 0$, such that the center of an absent band has no effect on the absorption profile being matched; $\mu_k$ influences the objective only when band $k$ is physically present ($A_k > 0$).
 
Each spectrum is peak-normalized such that its most intense band has unit amplitude ($A_1 \equiv 1$). Because $A_1$ is constant across the library, it is discarded. The resulting nine-dimensional optimization target $(\mu_1, A_2, \mu_2, A_3, \mu_3, A_4, \mu_4, A_5, \mu_5)$ comprises the position of the dominant band together with the relative amplitudes and positions of the four secondary bands. This representation captures both where a molecule absorbs and the overall shape of its absorption envelope, allowing specifications to be posed over an entire spectral band rather than over an isolated $\lambda_{\max}$. Before optimization the target vector is scaled group-wise to $[0,1]$, with all band amplitudes sharing one scaling and all band energies sharing another, such that the dimensionless amplitudes and the eV-scale energies contribute comparably to the tolerance distance rather than being normalized per dimension. The five resulting targets $T_1$ through $T_5$ are listed in SI Table S6 and displayed in Fig.~S16.
 
\paragraph{Time-Dependent Density Functional Theory Calculations.}
For each oligomer, long alkyl side chains were truncated to methyl groups to reduce computational cost~\cite{nozaki2025impact}, explicit hydrogen atoms were added, and 100 three-dimensional conformers were generated in RDKit~\cite{rdkit} using the ETKDGv3 embedding procedure~\cite{wang2020etkdgv3}, each subsequently optimized with the MMFF94 force field~\cite{halgren1996mmff94}; the lowest-energy conformer was retained. To promote planarity of the conjugated backbone while preserving side-chain flexibility, dihedral-angle constraints ($0^\circ$ or $180^\circ$, chosen from the force-field-optimized conformer) were applied only to the single bonds linking distinct ring or fused-ring systems~\cite{roseli2017origin}. Constrained geometry optimizations were then performed in ORCA~\cite{neese2020orca} using GFN2-xTB~\cite{bannwarth2019gfn2} with ALPB implicit solvation~\cite{ehlert2021alpb} in dichloromethane. Vertical excitation energies and oscillator strengths for the 15 lowest singlet excited states were computed for the optimized geometries by TD-DFT at the $\omega$B97X-3c/CPCM(dichloromethane) level~\cite{muller2023wb97x3c,barone1998cpcm} in ORCA. To obtain a continuous absorption profile rather than a discrete stick spectrum, only transitions above 240~nm were broadened (each by a Gaussian of width $\sigma = 0.12$~eV) and summed on a common grid spanning 240--500~nm, such that intense deep-UV bands below the cutoff did not contaminate the profile. The resulting broadened spectrum is the quantity subsequently parametrized by the five-Gaussian fit of Eq.~\ref{eq:gaussian_spectrum}.

\subsubsection{Input Representation and Preprocessing}

All input variables were mapped to the unit hypercube $[0,1]^M$ using min--max normalization performed separately for each dataset. 
For the molecular datasets and the oligomer case study, each candidate was represented by the full set of molecular descriptors computed with Mordred from its chemical structure~\cite{moriwaki2018mordred}, without manual feature selection. Descriptor columns with zero variance were removed, and principal component analysis (PCA)~\cite{jolliffe2002principal} was then performed separately for each dataset, retaining the number of components needed to explain 95\% of the total variance; the surrogate therefore operates on these principal components rather than on individual descriptors.
For pool-based datasets, output properties were standardized to zero mean and unit variance separately for each property before target selection and tolerance definition. This standardization makes tolerance values comparable across datasets and across property dimensions. For analytic benchmark functions and case studies without a predefined output scale, no output normalization was applied.

\subsubsection{Target and Tolerance Selection}

For each dataset, we selected five representative targets that span the output space. For pool-based datasets, we applied $k$-medoids clustering~\citep{Kaufman1990} with $k=5$ to the standardized property vectors of all candidates and used the resulting medoids as targets. For synthetic functions, we first sampled 20,000 inputs uniformly from the domain, evaluated the corresponding function values, and then applied $k$-means clustering~\citep{MacQueen1967} with $k=5$ in output space. The resulting cluster centers were used as targets. We defined a dataset-specific base tolerance $\varepsilon_0$ as the geometric mean of the pairwise Euclidean distances between the five targets. Performance was then evaluated at three tolerance levels, obtained by scaling this base tolerance by a tolerance ratio $r$,
\begin{equation}
    \varepsilon = r\,\varepsilon_0, \qquad r \in \{0.2,\,0.4,\,0.6\},
\end{equation}
where a larger tolerance ratio $r$ yields a wider acceptance range and less stringent specification.

\section{Results}

\subsection{Tolerance--Ball Acquisition Achieves the Best Overall Discovery Performance}

\begin{figure}[h!]
    \centering
    \includegraphics[width=\linewidth]{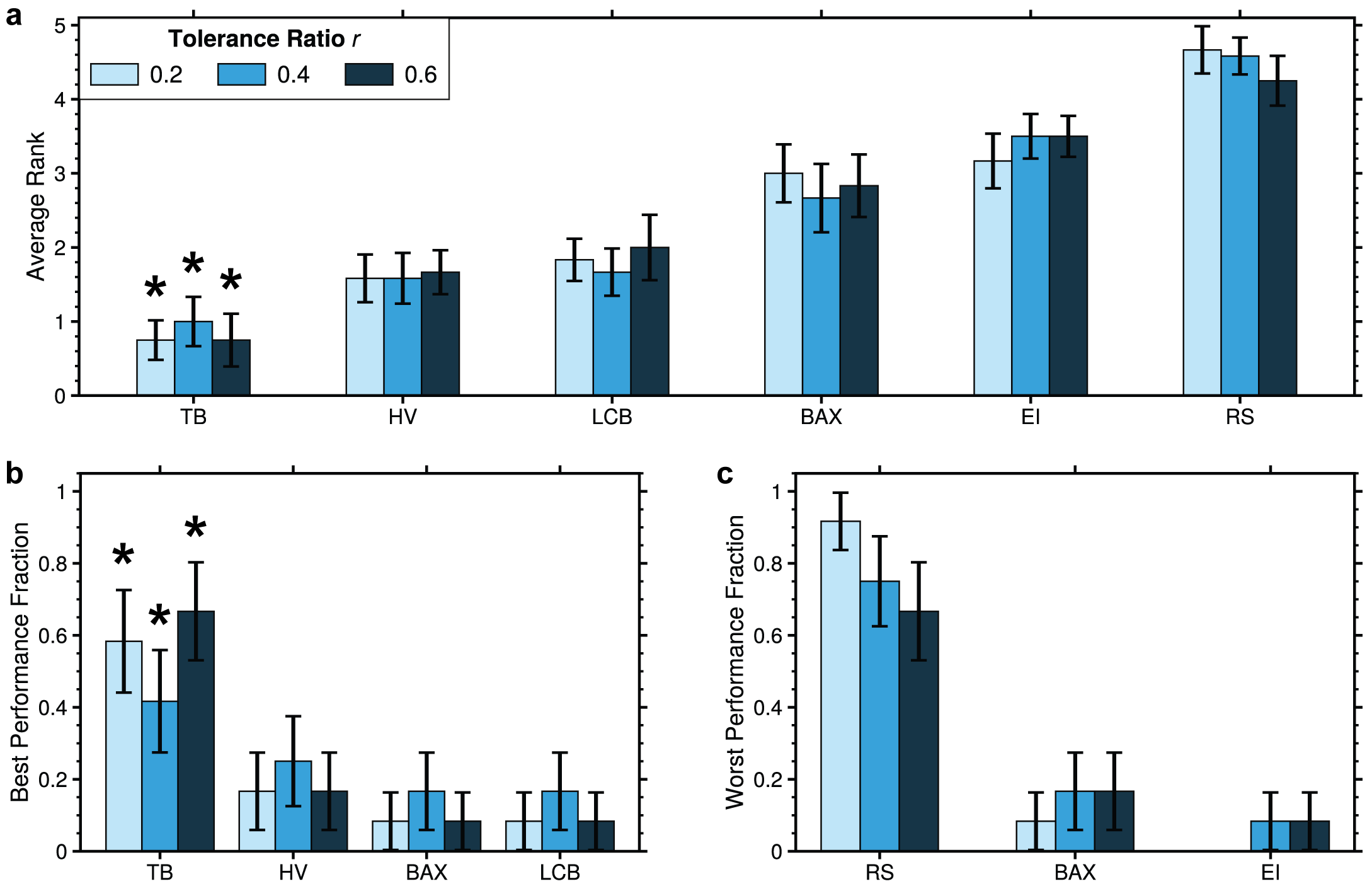}
    \caption{\textbf{Performance comparison of acquisition functions across tolerance ratios.} (a) Average rank (lower is better) of each acquisition function across the benchmark tasks, including the four analytical functions and the eight pool-based datasets, at tolerance ratios $r=0.2$, $0.4$, $0.6$. Within each task the six methods are ranked from $0$ (best) to $5$ (worst) by aggregate normalized diversity score, and these ranks are averaged over the twelve tasks. An asterisk marks the best-performing method at each tolerance ratio. (b) Fraction of tasks where each method achieved the best performance. (c) Fraction of tasks where each method achieved the worst performance. Methods that never attained the best (or worst) outcome are omitted from panel b (or c). The tolerance-ratio shading defined in panel a applies to all three panels. Bars show the mean across benchmark tasks; error bars show the standard error of the mean.}
    \label{fig:metrics}
\end{figure}

We first compare the acquisition functions on the benchmark suite, comprising the four analytical functions and the eight pool-based datasets, in terms of how effectively they recover multiple distinct valid designs under a fixed evaluation budget. This is the central criterion in our setting because specification-driven design often admits more than one acceptable solution, and a useful method should uncover a broad set of valid candidates rather than repeatedly refine a single promising region. Accordingly, performance here refers to the normalized diversity score ($D_\mathrm{c}$ for continuous tasks, $D_\mathrm{d}$ for discrete tasks) of valid discoveries per evaluation, aggregated across tasks and tolerance ratios $r$.

Figure~\ref{fig:metrics} shows that TB exhibits the strongest overall discovery performance across the benchmark suite. It achieves the lowest average rank at all three tolerance ratios (Fig.~\ref{fig:metrics}a) and the highest fraction of best-performing tasks (Fig.~\ref{fig:metrics}b), ranging from 0.42 at $r=0.4$ to 0.67 at $r=0.6$. LCB and HV form a competitive second tier but neither matches TB on either aggregate metric. In contrast, RS performs worst overall and accounts for the majority of worst-performing outcomes (Fig.~\ref{fig:metrics}c), with EI and BAX also producing occasional last-place outcomes among the non-random methods.

The performance gap between TB and HV highlights the importance of retaining probabilistic resolution within and near the tolerance range. Both acquisitions are designed for targeting ranges of valid discoveries, but HV trails TB at every tolerance ratio. One likely reason is that HV assigns similar acquisition values to many candidates whose posterior means lie inside the range, even when those candidates have different posterior probabilities of satisfying the range. This behavior can reduce its ability to prioritize the candidates with the highest probability of range satisfaction. This loss of resolution parallels the classical trade-off between Probability of Improvement (PI) and Expected Improvement (EI), in which PI saturates once a candidate is likely to improve and therefore disregards the magnitude of improvement that EI captures~\cite{R:2001_Jones_Taxonomy}. TB avoids this loss of resolution by directly scoring the posterior probability of range satisfaction, which provides a finer ranking of candidates according to their likelihood of meeting the specification.

The stronger performance of LCB relative to EI further indicates that extremum-seeking improvement is poorly aligned with range discovery. Although both methods use target discrepancy, EI evaluates improvement relative to the best discrepancy observed so far. This criterion is well suited to refining a single incumbent solution, but it can concentrate search near an already promising region after an acceptable candidate has been found. LCB combines posterior mean and uncertainty, allowing several plausible target-consistent regions to remain competitive during the search. This broader exploration behavior is more compatible with the goal of discovering multiple acceptable candidates, which likely explains why LCB consistently outperforms EI across tolerance ratios.

BAX underperforms relative to TB and LCB despite its set-oriented objective. This likely reflects its reliance on a posterior-mean approximation of the feasible region. When that approximation is incomplete, the acquisition may favor uncertainty reduction over candidates with high probability of satisfying the target range. Consequently, evaluations can be allocated to refining the feasible-set estimate rather than to recovering valid designs. This behavior is consistent with the weaker aggregate performance of BAX in Fig.~\ref{fig:metrics}.

\subsection{Tolerance--Ball Acquisition Maintains Balanced Discovery Across Targets}

\begin{figure}[h!]
    \centering
    \includegraphics[width=\linewidth]{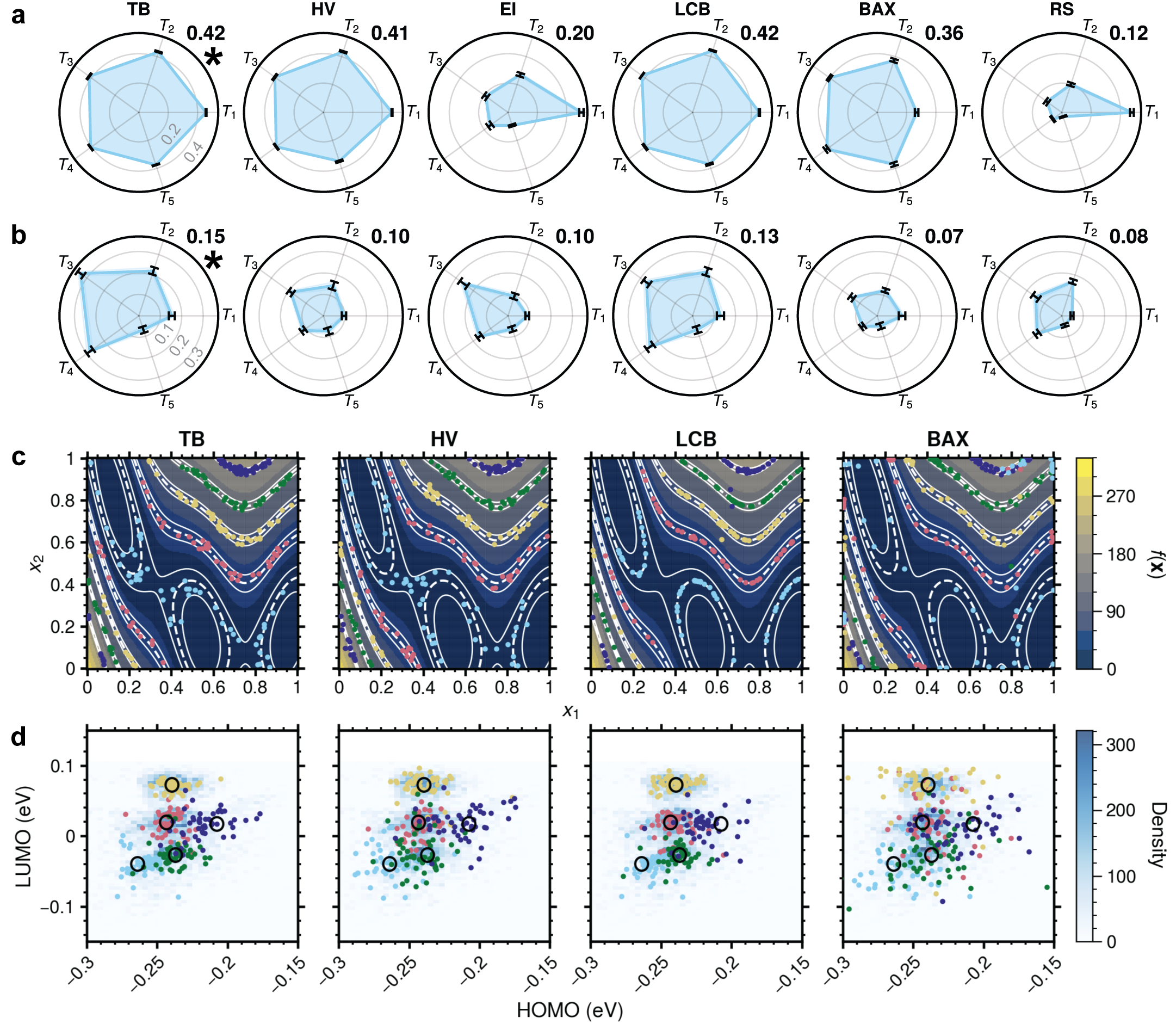}
   \caption{\textbf{Comparison of acquisition functions across the Branin and QM9 benchmarks.} (a, b) Radar plots of the normalized diversity score ($D_\mathrm{c}$ for Branin, $D_\mathrm{d}$ for QM9) for five targets $T_1$ through $T_5$ in the (a) Branin and (b) QM9 tasks at $r=0.4$, with the mean score reported on each plot. Each spoke corresponds to one target; larger radial values indicate higher normalized diversity scores, corresponding to a larger number of unique valid designs. Error bars indicate the standard error of the mean. An asterisk marks the best-performing method. (c) Comparison of solutions on the Branin function landscape. Background shading indicates the function value. White dashed contours denote the five target levels, and white solid contours denote the corresponding tolerance ranges. Colored markers indicate discovered points, with color denoting the associated target. (d) Comparison of solution on the QM9 property space defined by HOMO and LUMO. Background shading indicates the density of samples in the dataset. Open black circles mark the five target locations, and colored markers indicate discovered points with color denoting the associated target.}
    \label{fig:acq}
\end{figure}

Figure~\ref{fig:acq} examines the per-target structure of discovery performance using the Branin benchmark and the QM9 molecular dataset as representative examples from the continuous and discrete settings, respectively; radar plots for the remaining benchmark tasks are provided in SI Section S4.
The aggregate advantage of TB arises from its ability to sustain strong discovery across multiple targets rather than concentrating performance on only a subset. At an intermediate tolerance ratio ($r = 0.4$), TB achieves a high mean normalized diversity score $D_\mathrm{c}$ of 0.42 across the five targets in the Branin task (Fig.~\ref{fig:acq}a), with comparable coverage across all five targets. HV and LCB also perform well in this low-dimensional setting, with mean scores of 0.41 and 0.42, whereas EI achieves a mean of 0.20 and shows pronounced concentration on $T_1$. BAX achieves an intermediate mean score of 0.36. A similar pattern appears in the QM9 inverse-design task (Fig.~\ref{fig:acq}b), where TB again combines the highest mean score (0.15) with comparatively even performance across the full target set, while EI, BAX, and RS achieve lower scores and concentrate on a subset of targets. This balance is reflected in the across-target variance of $D_\mathrm{c}$ and $D_\mathrm{d}$ (SI Fig. S1). RS and EI show the largest across-target variation at moderate to high tolerance ratios, consistent with discovery concentrated on a subset of targets, whereas TB falls below both at $r = 0.4$ and $r = 0.6$. At $r = 0.2$, the methods are comparable. These results indicate that TB not only improves overall discovery but also tends to distribute discovery outcomes more evenly among targets.

The discovered-point plots further indicate that TB recovers candidates that are both valid and well distributed across the tolerance regions. In the Branin example (Fig.~\ref{fig:acq}c), TB identifies candidates across multiple disjoint acceptable regions while keeping the discovered points closely confined to the prescribed tolerance ranges. HV also finds candidates in multiple regions, although its discoveries are less consistently confined to the admissible ranges. LCB selects many points near the target contours, but several fall outside the tolerance regions, consistent with a search bias toward the target contour rather than the full range-satisfaction criterion. BAX yields an even more diffuse distribution of discovered points, with a larger fraction of points outside the admissible regions.

The QM9 property-space visualization reveals the same qualitative pattern in a discrete molecular dataset. In Fig.~\ref{fig:acq}d, TB produces a denser and more localized occupancy around the target locations, whereas LCB and HV generate more diffuse point clouds and BAX is the most dispersed. These visualizations support the interpretation that TB improves discovery by prioritizing candidates with high posterior probability of satisfying the tolerance criterion, rather than by broadly sampling near the target region. Additional radar plots in SI Section S4 show consistent behavior across the remaining benchmark tasks.

\subsection{Off-Target Hit Dynamics Reveal Target Fidelity During Parallel Target Search}

\begin{figure}[h!]
    \centering
    \includegraphics[width=0.5\linewidth]{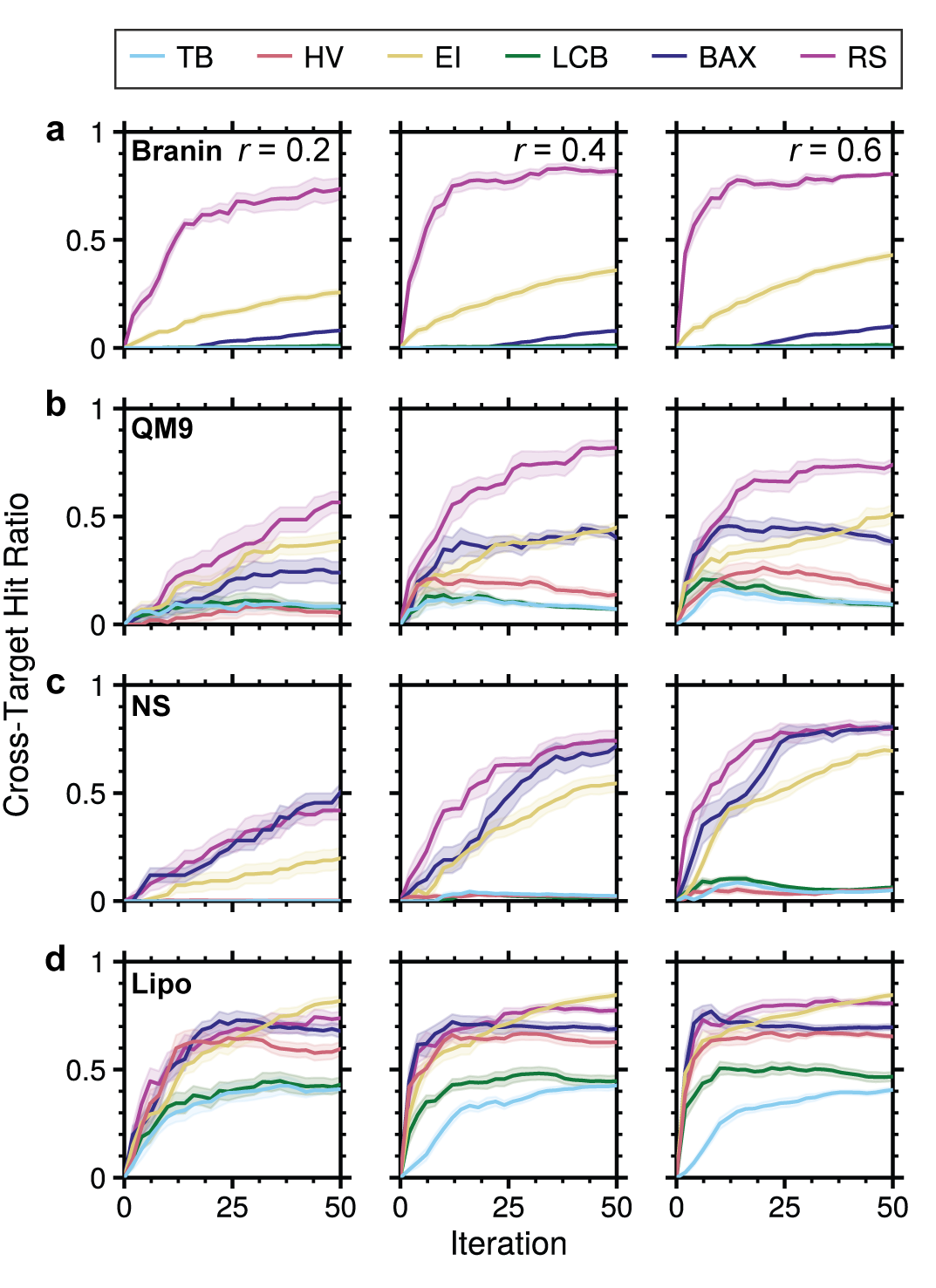}
    \caption{\textbf{Off-target hit dynamics during parallel target Bayesian optimization.} Cross-target hit ratio as a function of BO iteration for the (a) Branin, (b) QM9, (c) Nanoparticle sysnthesis (NS), and (d) Lipophilicity (Lipo) tasks at tolerance ratios $r=0.2$, $0.4$, $0.6$. Solid lines show the mean across repeated runs. Shaded regions show the standard error of the mean.}
    \label{fig:progress}
\end{figure}

In principle, algorithms for parallel target search should yield not only efficient discovery of valid candidates but also faithful allocation of those discoveries to their intended targets. 
In other words, the efficacy of an algorithm should not necessarily hinge on ``accidental'' discoveries where suitable candidates for one objective originate from the search for another.
Figure~\ref{fig:progress} tracks this property through the cross-target hit ratio. For each candidate proposed for a given target, we recorded whether the resulting property vector fell inside the tolerance range of another target. Because the target ranges considered here are disjoint, such an event indicates that the acquisition selected a valid design for the wrong target. The evolution of this ratio over BO iterations therefore provides a measure of target fidelity. Methods with low off-target hit ratios more consistently preserve the intended target, whereas methods with high ratios more frequently recover valid designs from other target ranges.

Target fidelity reflects both the acquisition function and the structure of the function landscape. In the Branin task (Fig.~\ref{fig:progress}a), TB, HV, LCB, and BAX maintain near-zero off-target hit ratios across all tolerance ratios, whereas RS rises rapidly to a high plateau, which is expected since its search strategy is not actually targeted, and EI increases more gradually. 
The QM9 task (Fig.~\ref{fig:progress}b) shows the same qualitative ordering between RS and the model-guided methods but the separation is smaller, and BAX and EI also drift upward at larger tolerance ratios. This behavior suggests that the design regions associated with different targets are less distinctly separated in the chosen representation. The nanoparticle synthesis (NS) task (Fig.~\ref{fig:progress}c) similarly preserves the advantage of TB, HV, and LCB, although BAX and EI here climb closer to RS. In the Lipophilicity (Lipo) task (Fig.~\ref{fig:progress}d), all methods show more rapid growth in off-target hit ratio, indicating that the property landscape or candidate distribution makes target-specific discrimination intrinsically more difficult. Even in this more challenging setting, TB and LCB attain the lowest off-target hit ratios.

Higher off-target hit ratios are associated with weaker target-specific discovery in these experiments. RS produces the highest off-target hit ratios and also shows the weakest overall discovery performance, whereas TB combines the lowest off-target hit ratios with the strongest aggregate discovery (Fig.~\ref{fig:metrics}). This relationship suggests that the advantage of TB is unlikely to arise from opportunistically recovering valid designs from any target range but instead from selecting candidates with high posterior probability of satisfying the intended target range. The off-target analysis therefore provides additional evidence that explicit range-satisfaction scoring can improve both discovery efficiency and target fidelity in parallel target search.
 
\subsection{Case Study: Polymer Molecular Weight Distribution Design}

\begin{figure}[p]
    \centering
    \includegraphics[width=\linewidth]{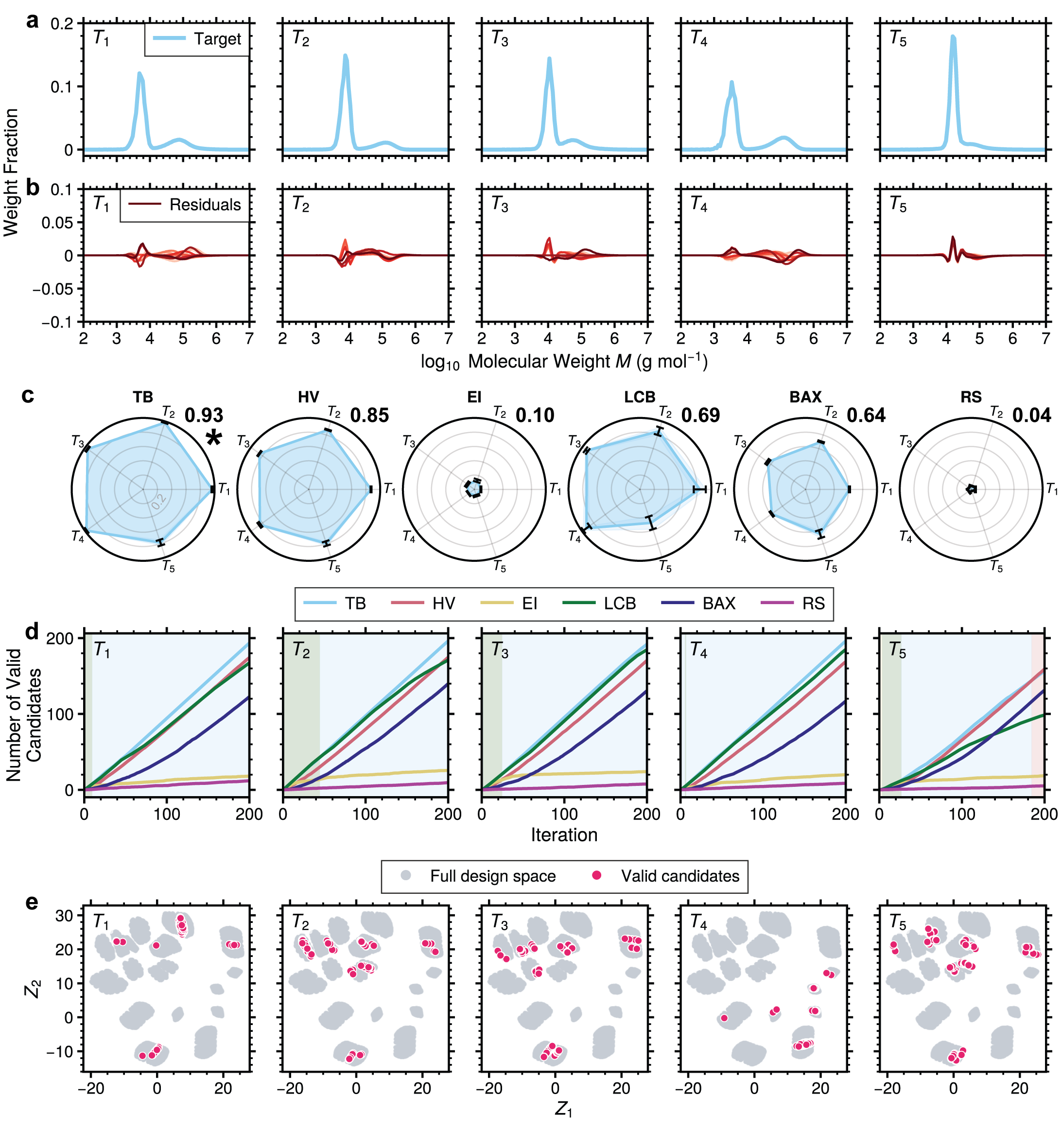}
    \caption{\textbf{Polymer molecular weight distribution design.} (a) Five target MWDs and (b) the corresponding residuals between target and recovered distributions for valid designs discovered by TB, plotted as a function of $\log_{10}$ molecular weight. (c) Radar plot of normalized diversity score $D_\mathrm{d}$ across the five targets $T_1$ through $T_5$ for all acquisition functions, with the mean score reported on each plot and error bars indicating the standard error of the mean. An asterisk marks the best-performing method. (d) Cumulative number of valid designs identified as a function of BO iteration for each target $T_1$ through $T_5$, with shaded background indicating the dominant acquisition function for each target. (e) UMAP projection of the five-dimensional reaction-condition space and the valid designs discovered for each target $T_1$ through $T_5$, with axes $Z_1$ and $Z_2$ denoting the two UMAP components.}
    \label{fig:kmc}
\end{figure}

Whereas the preceding benchmark tasks are drawn from existing datasets, we now turn to case studies designed to more closely reflect real-world considerations.
The KMC polymerization task addresses inverse polymer design for a prescribed full molecular weight distribution (MWD).
Rather than optimizing a scalar property such as number-average molecular weight or dispersity, the objective here is to identify reaction conditions that reproduce a target MWD.
Designing the full MWD is well motivated, because its breadth, skew, and modality influence polymer properties beyond what summary statistics capture~\cite{gentekos2016beyond,gentekos2019controlling}.
This goal has been pursued through controlled-synthesis and flow strategies~\cite{yoo1999molecular,li2018tuning,walsh2020general,liu2020comprehensive} and more recently through machine-learning approaches, including reinforcement learning and KMC-based active learning and multi-objective reverse engineering~\cite{liu2020inverse,mackenzie2024computer,zhou2024active,fiosina2025evolutionary,fang2026unified}.
However, recovering reaction conditions for a target MWD under an explicit, range-aware tolerance criterion remains comparatively less explored.
Here, each specification is a 100-bin weight-fraction distribution, and a design counts as valid when its simulated distribution falls within the prescribed Euclidean tolerance of the target. 
Because the design space is substantially larger (51,200 candidates) and valid regions are sparser than in the pool-based benchmarks, we extended the evaluation budget to 200 evaluations for this case study. The five target distributions span a range of shapes, including unimodal profiles centered at different molecular weights and bimodal profiles with a dominant peak and a weaker high-molecular-weight shoulder (Fig.~\ref{fig:kmc}a). Figure~\ref{fig:kmc}b also shows that the residuals between target and recovered distributions remain small in magnitude across the full molecular-weight range, indicating that the Euclidean tolerance criterion captures the principal features of the distribution in this fixed representation.

TB provides the strongest and most balanced recovery of valid polymerization conditions. In the radar plot of normalized diversity score (Fig.~\ref{fig:kmc}c), TB achieves the highest mean $D_\mathrm{d}$ of 0.93 across the five targets, followed by HV at 0.85, LCB at 0.69, and BAX at 0.64. EI and RS perform poorly, with mean scores of 0.10 and 0.04 respectively, reflecting that valid reaction conditions occupy only a small region of the five-dimensional design space and are unlikely to be recovered by extremum-seeking or unguided search. The target-resolved discovery trajectories further show that TB identifies valid designs more rapidly and consistently than the other methods. Figure~\ref{fig:kmc}c reports the cumulative number of valid designs found for each target over 200 BO iterations. TB rises steeply during the early stage of the search and reaches the largest, or near-largest, number of valid designs for most targets by the end of the evaluation budget. LCB and HV also accumulate valid designs at competitive rates, whereas BAX trails the leading methods. EI and RS remain near zero throughout the search.

The valid designs discovered by TB are also diverse in reaction-condition space, even when they produce similar MWDs. For each target, a representative pair of valid designs that yield nearly identical distributions is reported in SI Fig.~S14; the two designs differ across all five reaction variables (SI Table~S5), confirming that distinct combinations of conditions can satisfy the same product-level specification. The UMAP~\cite{mcinnes2018umap} projection in Fig.~\ref{fig:kmc}d illustrates this, with the valid designs for every target occupying multiple separated regions of the embedding. This degeneracy is valuable for process design because it reveals multiple reaction pathways to the same product specification and provides flexibility for subsequent optimization of cost, robustness, scalability, or experimental feasibility.

\subsection{Case Study: Sequence-Defined Conjugated Oligomers Discovery}

\begin{figure}[p]
    \centering
    \includegraphics[width=\linewidth]{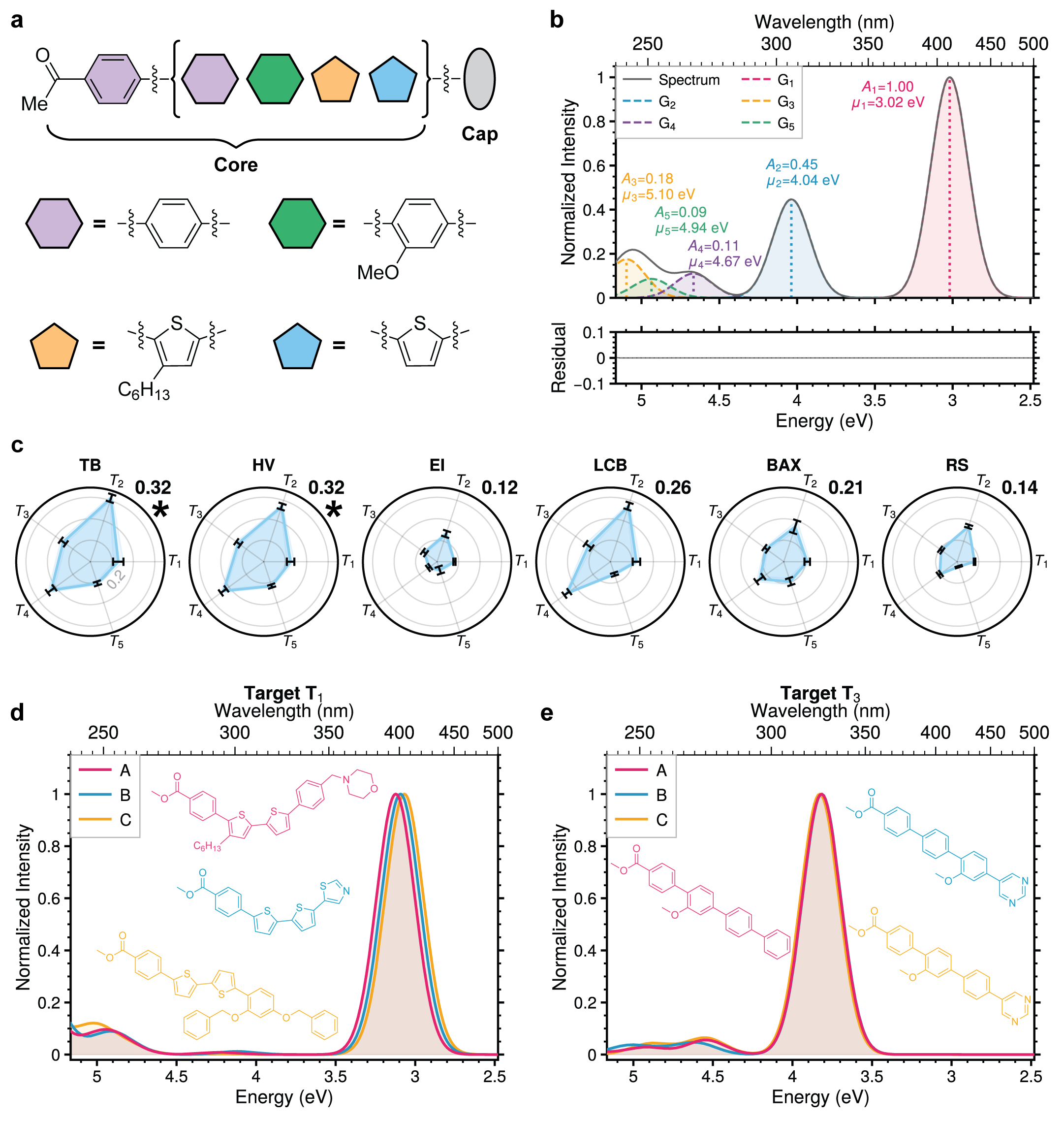}
    \caption{\textbf{Tolerance-aware discovery of sequence-defined conjugated oligomers.} (a) Schematic of the oligomer architecture, comprising a methyl ester head group, a conjugated core assembled from four monomer types, and a terminal cap. (b) Representative simulated absorption spectrum decomposed in energy space into five Gaussian components $G_1$--$G_5$, each parametrized by an amplitude $A_k$ and center $\mu_k$ (with fixed width); the fit residual is shown below and the top axis gives the corresponding wavelength. (c) Radar plot of normalized diversity score $D_\mathrm{d}$ across the five targets $T_1$ through $T_5$ for all acquisition functions, with the mean score reported on each plot and error bars indicating the standard error of the mean. An asterisk marks the best-performing method. (d, e) Three chemically distinct oligomers (A, B, C) discovered by TB for targets $T_1$ (d) and $T_3$ (e), whose simulated spectra are similar.}
    \label{fig:oligomer}
\end{figure}

The sequence-defined oligomer task tests range-aware discovery in a finite molecular library, where discrete changes in core sequence or terminal-cap chemistry reshape the entire optical absorption band rather than merely shifting its peak.  
Accordingly, each target is specified not by a single wavelength but as a region of the spectral-shape space introduced above, that is, the nine-dimensional target vector $(\mu_1, A_2, \mu_2, A_3, \mu_3, A_4, \mu_4, A_5, \mu_5)$ comprising the dominant-band energy together with the relative amplitudes and positions of the four secondary bands; a representative spectrum and its five-Gaussian decomposition are shown in Fig.~\ref{fig:oligomer}b.
The residual between the simulated spectrum and this five-Gaussian fit (Fig.~\ref{fig:oligomer}b, bottom) remains small and largely structureless, confirming that the compact parametrization reproduces the absorption band.
The five targets $T_1$--$T_5$ were obtained by $k$-medoids clustering of the Gaussian spectral vectors and span the dominant-band variation observed in the library (see SI Table~S6 and Fig.~S16).

Performance differences among acquisition functions are less pronounced in this discrete molecular setting than in the polymerization case study (Fig.~\ref{fig:oligomer}c). 
TB attains the highest mean normalized diversity score ($D_\mathrm{d} = 0.32$) and remains among the strongest methods across the five target bands, but the separation among methods is modest and random sampling is competitive for several targets. 
The target-resolved discovery trajectories in SI Fig.~S17 corroborate this view.
From this data, our observation is that no single acquisition function dominates all bands across the full evaluation budget, although TB and HV sustain higher discovery rates.

The TB discoveries nevertheless reveal a chemically important form of degeneracy, where structurally diverse oligomers can lead to comparable absorption profiles. Figures~\ref{fig:oligomer}d and~\ref{fig:oligomer}e present three TB-discovered oligomers for targets $T_1$ and $T_3$, respectively. Within each panel the molecules differ in conjugated-core sequence and terminal-cap chemistry, yet their simulated spectra are similar, sharing an almost identical dominant-band position to within the prescribed tolerance. This many-to-one mapping from molecular structure to optical response is valuable from a design perspective because it provides multiple synthetic routes to a single absorption specification, leaving room for subsequent selection according to criteria outside the present objective, such as synthetic accessibility or stability.

\section{Conclusion}

Many materials and product design problems are guided by specification satisfaction. A candidate is valuable when its properties fall within prescribed tolerance ranges, rather than when a single objective reaches an optimum. We reformulated this challenge as a parallel target-range discovery problem and derived a closed-form Tolerance Ball (TB) acquisition that directly scores the posterior probability of range satisfaction through a noncentral $\chi^2$ approximation. Because the acquisition is sampling-free and analytically tractable, it offers a computationally efficient alternative to extremum-seeking strategies and sampling-based set-estimation methods.
 
Across synthetic functions, molecular libraries, and case studies, TB consistently achieved the strongest and most balanced recovery of valid designs, outperforming standard Bayesian optimization baselines and recent goal-oriented approaches at multiple tolerance ratios. This advantage likely arises because TB explicitly scores range satisfaction for the intended target, yielding low cross-target hit ratios and high target fidelity. Improvement-based or uncertainty-driven methods, in contrast, can drift toward off-target ranges or concentrate on refining a single incumbent solution. In the large, sparse design space of the kinetic Monte Carlo polymerization task, TB recovered distinct reaction conditions that accurately reproduced prescribed molecular weight distributions (MWDs). This distributional target class has practical relevance for optimizing mechanical, rheological, and processing behavior. A similar structure--property degeneracy emerged in the sequence-defined oligomer library, where chemically distinct molecules produced nearly identical absorption bands, providing multiple molecular candidates for a single optical specification.
 
Although TB is not always the top-ranked method for every individual target or dataset, it achieves strong performance across the full range of settings considered, making it a reliable default choice for specification-driven search when the structure of the feasible region is not known in advance. The benefit of surrogate-guided search also depends on the structure of the design space: when many valid molecules fall within the target ranges in a discrete library, the performance gap between TB and random sampling narrows, indicating that model guidance adds the most value when valid regions are sparse or disconnected. Future work could incorporate explicit diversity incentives into the acquisition, relax the isotropic variance approximation through full multi-output Gaussian process models, and introduce adaptive tolerances that reflect evolving design specifications. Integrating range-aware optimization with automated synthesis and characterization pipelines could enable closed-loop, specification-driven materials discovery.

\section{Acknowledgments}\label{sec:acknowledgement}
S.J. and M.A.W. acknowledge support from the National Science Foundation under Grant No. 2118861. 
Simulations and analyses were performed using resources from Princeton Research Computing at Princeton University, which is a consortium led by the Princeton Institute for Computational Science and Engineering (PICSciE) and Office of Information Technology's Research Computing.
All code required to recreate the results above can be obtained at \texttt{\href{https://github.com/webbtheosim/target-range-optimization}{https://github.com/webbtheosim/target-range-optimization}}.

\bibliography{ref,group_references}

@Article{R:2001_Jones_Taxonomy,
  author    = {Donald R. Jones},
  journal   = {Journal of Global Optimization},
  title     = {A Taxonomy of Global Optimization Methods Based on Response Surfaces},
  year      = {2001},
  number    = {4},
  pages     = {345--383},
  volume    = {21},
  doi       = {10.1023/a:1012771025575},
  publisher = {Springer Science and Business Media {LLC}},
}

@inproceedings{mockus1974bayesian,
  author       = {Močkus, J.},
  title        = {On Bayesian Methods for Seeking the Extremum},
  booktitle    = {Optimization Techniques {IFIP} Technical Conference Novosibirsk, July 1--7, 1974},
  series       = {Lecture Notes in Computer Science},
  volume       = {27},
  pages        = {400--404},
  year         = {1975},
  publisher    = {Springer},
  address      = {Berlin, Heidelberg},
  editor       = {Marchuk, G. I.},
  doi          = {10.1007/3-540-07165-2_55},
  note         = {Published in 1975, proceedings from the 1974 conference}
}

@article{jones1998efficient,
  author    = {Jones, Donald R. and Schonlau, Matthias and Welch, William J.},
  title     = {Efficient Global Optimization of Expensive Black-Box Functions},
  journal   = {Journal of Global Optimization},
  volume    = {13},
  number    = {4},
  pages     = {455--492},
  year      = {1998},
  month     = {December},
  publisher = {Kluwer Academic Publishers},
  doi       = {10.1023/A:1008306431147},
  note      = {The well-known EGO paper}
}

@inproceedings{snoek2012practical,
  author       = {Snoek, Jasper and Larochelle, Hugo and Adams, Ryan P.},
  title        = {Practical {Bayesian} Optimization of Machine Learning Algorithms},
  booktitle    = {Advances in Neural Information Processing Systems 25 ({NeurIPS} 2012)},
  pages        = {2951--2959},
  year         = {2012},
  editor       = {Pereira, Fernando and Burges, Christopher J. C. and Bottou, Léon and Weinberger, Kilian Q.},
  publisher    = {Curran Associates, Inc.},
  doi          = {10.5555/2999325.2999464},
  note         = {arXiv:1206.2944 [stat.ML]}
}

@article{wilding2025integrating,
  author    = {Wilding, Clarissa Y. P. and Bourne, Richard A. and Warren, Nicholas J.},
  title     = {Integrating Mechanistic Modelling with Bayesian Optimisation: Accelerated Self-Driving Laboratories for {RAFT} Polymerisation},
  journal   = {Digital Discovery},
  volume    = {4},
  number    = {10},
  pages     = {2797--2803},
  year      = {2025},
  month     = {August},
  publisher = {Royal Society of Chemistry},
  doi       = {10.1039/D5DD00258C},
  note      = {Open Access}
}

@inproceedings{gotovos2013active,
  author       = {Gotovos, Alkis and Casati, Nathalie and Hitz, Gregory and Krause, Andreas},
  title        = {Active Learning for Level Set Estimation},
  booktitle    = {Proceedings of the Twenty-Third International Joint Conference on Artificial Intelligence (IJCAI 2013)},
  pages        = {1344--1350},
  year         = {2013},
  publisher    = {AAAI Press},
}

@inproceedings{bryan2005active,
  author       = {Bryan, Brent and Schneider, Jeff G. and Nichol, Robert C. and Miller, Christopher J. and Genovese, Christopher R. and Wasserman, Larry A.},
  title        = {Active Learning for Identifying Function Threshold Boundaries},
  booktitle    = {Advances in Neural Information Processing Systems 18 ({NeurIPS} 2005)},
  pages        = {163--170},
  year         = {2005},
  editor       = {Weiss, Yair and Sch{\"o}lkopf, Bernhard and Platt, John C.},
  publisher    = {MIT Press},
}

@inproceedings{neiswanger2021bayesian,
  author       = {Neiswanger, Willie and Wang, Ke Alexander and Ermon, Stefano},
  title        = {Bayesian Algorithm Execution: Estimating Computable Properties of Black-Box Functions Using Mutual Information},
  booktitle    = {Proceedings of the 38th International Conference on Machine Learning},
  volume       = {139},
  pages        = {8005--8015},
  year         = {2021},
  editor       = {Meila, Marina and Zhang, Tong},
  series       = {Proceedings of Machine Learning Research},
  month        = {18--24 Jul},
  publisher    = {PMLR},
}

@article{chitturi2024targeted,
  author    = {Chitturi, Sathya R. and Ramdas, Akash and Wu, Yue and Rohr, Brian and Ermon, Stefano and Dionne, Jennifer and Jornada, Felipe H. da and Dunne, Mike and Tassone, Christopher and Neiswanger, Willie and Ratner, Daniel},
  title     = {Targeted Materials Discovery Using {Bayesian} Algorithm Execution},
  journal   = {npj Computational Materials},
  volume    = {10},
  number    = {1},
  pages     = {156},
  year      = {2024},
  month     = {July},
  publisher = {Springer Nature},
  doi       = {10.1038/s41524-024-01326-2},
  note      = {Open access}
}

@article{tian2025materials,
  author    = {Tian, Yuan and Li, Tongtong and Pang, Jianbo and Zhou, Yumei and Xue, Dezhen and Ding, Xiangdong and Lookman, Turab},
  title     = {Materials Design with Target-Oriented {Bayesian} Optimization},
  journal   = {npj Computational Materials},
  volume    = {11},
  number    = {1},
  pages     = {209},
  year      = {2025},
  month     = {July},
  publisher = {Springer Nature},
  doi       = {10.1038/s41524-025-01704-4}
}

@book{williams2006gaussian,
  author    = {Rasmussen, Carl Edward and Williams, Christopher K. I.},
  title     = {Gaussian Processes for Machine Learning},
  year      = {2006},
  publisher = {MIT Press},
  address   = {Cambridge, MA},
  series    = {Adaptive Computation and Machine Learning},
  isbn      = {978-0-262-18253-9},
  doi       = {10.7551/mitpress/3206.001.0001}
}

@article{pellegrino2020machine,
  author    = {Pellegrino, Francesco and Isopescu, Raluca and Pelluti{\`e}, Letizia and Sordello, Fabrizio and Rossi, Andrea M. and Ortel, Erik and Martra, Gianmario and Hodoroaba, Vasile-Dan and Maurino, Valter},
  title     = {Machine Learning Approach for Elucidating and Predicting the Role of Synthesis Parameters on the Shape and Size of {TiO2} Nanoparticles},
  journal   = {Scientific Reports},
  volume    = {10},
  number    = {1},
  pages     = {18910},
  year      = {2020},
  month     = {November},
  publisher = {Springer Nature},
  doi       = {10.1038/s41598-020-75967-w}
}

@article{oliver2025b,
  author    = {Oliver, Wesley W. and Jacobs, William M. and Webb, Michael A.},
  title     = {When B2 is Not Enough: Evaluating Simple Metrics for Predicting Phase Separation of Intrinsically Disordered Proteins},
  journal   = {The Journal of Physical Chemistry B},
  volume    = {129},
  number    = {37},
  pages     = {9551--9565},
  year      = {2025},
  month     = {September},
  publisher = {American Chemical Society},
  doi       = {10.1021/acs.jpcb.5c04955}
}

@book{Kaufman1990,
  author    = {Kaufman, Leonard and Rousseeuw, Peter J.},
  title     = {Finding Groups in Data: An Introduction to Cluster Analysis},
  year      = {1990},
  publisher = {John Wiley \& Sons},
  address   = {New York},
  isbn      = {978-0-471-87876-6},
  doi       = {10.1002/9780470316801}
}

@inproceedings{MacQueen1967,
  author    = {MacQueen, James},
  title     = {Some Methods for Classification and Analysis of Multivariate Observations},
  booktitle = {Proceedings of the Fifth Berkeley Symposium on Mathematical Statistics and Probability, Volume 1: Statistics},
  editor    = {Le Cam, Lucien M. and Neyman, Jerzy},
  volume    = {1},
  pages     = {281--297},
  year      = {1967},
  publisher = {University of California Press},
  address   = {Berkeley, CA}
}

@inproceedings{uhrenholt2019efficient,
  author    = {Uhrenholt, Anders Kirk and Jensen, Bj{\o}rn Sand},
  title     = {Efficient Bayesian Optimization for Target Vector Estimation},
  booktitle = {Proceedings of the Twenty-Second International Conference on Artificial Intelligence and Statistics},
  volume    = {89},
  pages     = {2661--2670},
  year      = {2019},
  editor    = {Chaudhuri, Kamalika and Sugiyama, Masashi},
  series    = {Proceedings of Machine Learning Research},
  publisher = {PMLR}
}

@article{gallagher2025data,
  author    = {Gallagher, Quinn M. and Webb, Michael A.},
  title     = {Data Efficiency of Classification Strategies for Chemical and Materials Design},
  journal   = {Digital Discovery},
  volume    = {4},
  number    = {1},
  pages     = {135--148},
  year      = {2025},
  month     = {January},
  publisher = {Royal Society of Chemistry},
  doi       = {10.1039/D4DD00298A}
}

@article{kushner1964method,
  author    = {Kushner, Harold J.},
  title     = {A New Method of Locating the Maximum Point of an Arbitrary Multipeak Curve in the Presence of Noise},
  journal   = {Journal of Basic Engineering},
  volume    = {86},
  number    = {1},
  pages     = {97--106},
  year      = {1964},
  month     = {March},
  doi       = {10.1115/1.3653121}
}

@inproceedings{srinivas2010gaussian,
  author    = {Niranjan Srinivas and Andreas Krause and Sham M. Kakade and Matthias Seeger},
  title     = {Gaussian Process Optimization in the Bandit Setting: No Regret and Experimental Design},
  booktitle = {Proceedings of the 27th International Conference on International Conference on Machine Learning},
  series    = {ICML'10},
  year      = {2010},
  pages     = {1015--1022},
  publisher = {Omnipress},
  doi       = {10.5555/3104322.3104451}
}

@article{griffiths2020constrained,
  author    = {Ryan-Rhys Griffiths and Jos{\'e} Miguel Hern{\'a}ndez-Lobato},
  title     = {Constrained Bayesian optimization for automatic chemical design using variational autoencoders},
  journal   = {Chemical Science},
  volume    = {11},
  number    = {2},
  pages     = {577--586},
  year      = {2020},
  publisher = {Royal Society of Chemistry},
  doi       = {10.1039/C9SC04026A}
}

@article{knowles2006parego,
  author  = {Knowles, Joshua},
  title   = {{ParEGO}: A Hybrid Algorithm With On-Line Landscape Approximation for Expensive Multiobjective Optimization Problems},
  journal = {IEEE Transactions on Evolutionary Computation},
  volume  = {10},
  number  = {1},
  pages   = {50--66},
  year    = {2006},
  doi     = {10.1109/TEVC.2005.851274}
}

@inproceedings{emmerich2011hypervolumebased,
  author    = {Emmerich, Michael T. M. and Deutz, André H. and Klinkenberg, Jan Willem},
  title     = {Hypervolume-Based Expected Improvement: Monotonicity Properties and Exact Computation},
  booktitle = {2011 IEEE Congress on Evolutionary Computation (CEC)},
  year      = {2011},
  pages     = {2147--2154},
  doi       = {10.1109/CEC.2011.5949880}
}

@inproceedings{daulton2020differentiable,
  author    = {Daulton, Samuel and Balandat, Maximilian and Bakshy, Eytan},
  title     = {Differentiable Expected Hypervolume Improvement for Parallel Multi-Objective Bayesian Optimization},
  booktitle = {Advances in Neural Information Processing Systems},
  volume    = {33},
  year      = {2020},
}

@inproceedings{marques2018contour,
  author    = {Marques, Alexandre N. and Lam, Remi R. and Willcox, Karen E.},
  title     = {Contour Location via Entropy Reduction Leveraging Multiple Information Sources},
  booktitle = {Advances in Neural Information Processing Systems (NeurIPS 2018)},
  volume    = {31},
  year      = {2018},
  doi       = {10.5555/3327345.3327428}
}

@article{kim2019active,
  author  = {Kim, Chiho and Chandrasekaran, Anand and Jha, Anurag and Ramprasad, Rampi},
  title   = {Active-learning and materials design: the example of high glass transition temperature polymers},
  journal = {MRS Communications},
  volume  = {9},
  number  = {3},
  pages   = {860--866},
  year    = {2019},
  doi     = {10.1557/mrc.2019.78}
}

@article{yu2012identification,
  author  = {Yu, L. and Zunger, A.},
  title   = {Identification of Potential Photovoltaic Absorbers Based on First-Principles Spectroscopic Screening},
  journal = {Physical Review Letters},
  volume  = {108},
  pages   = {068701},
  year    = {2012},
  doi     = {10.1103/PhysRevLett.108.068701}
}

@article{qian2019glass,
  author  = {Qian, Zhiyuan and Cao, Zhiqiang and Galuska, Luke and Zhang, Song and Xu, Jie and Gu, Xiaodan},
  title   = {Glass Transition Phenomenon for Conjugated Polymers},
  journal = {Macromolecular Chemistry and Physics},
  volume  = {220},
  number  = {11},
  pages   = {1900062},
  year    = {2019},
  doi     = {10.1002/macp.201900062}
}

@article{kim2020upper,
  author  = {Kim, Sunghyun and M{\'a}rquez, Jos{\'e} A. and Unold, Thomas and Walsh, Aron},
  title   = {Upper Limit to the Photovoltaic Efficiency of Imperfect Crystals from First Principles},
  journal = {Energy \& Environmental Science},
  volume  = {13},
  number  = {5},
  pages   = {1481--1491},
  year    = {2020},
  doi     = {10.1039/D0EE00291G}
}

@book{bhushan2013principles,
  author    = {Bhushan, Bharat},
  title     = {Principles and Applications of Tribology},
  edition   = {2},
  series    = {Tribology in Practice Series},
  year      = {2013},
  publisher = {John Wiley \& Sons},
  doi       = {10.1002/9781118403020},
  isbn      = {978-1-119-94454-6}
}

@article{chen2024investigation,
  author  = {Chen, Yang-Yuan and Horng, Jeng-Haur},
  title   = {Investigation of Lubricant Viscosity and Third-Particle Contribution to Contact Behavior in Dry and Lubricated Three-Body Contact Conditions},
  journal = {Frontiers in Mechanical Engineering},
  volume  = {10},
  pages   = {1390335},
  year    = {2024},
  doi     = {10.3389/fmech.2024.1390335}
}

@article{jiang2024property,
  author = {Jiang, Shengli and Dieng, Adji Bousso and Webb, Michael A},
  title = {Property-guided generation of complex polymer topologies using variational autoencoders},
  journal = {npj Computational Materials},
  volume = {10},
  number = {1},
  pages = {139},
  year = {2024},
  doi = {10.1038/s41524-024-01328-0}
}

@article{wu2018moleculenet,
  author = {Wu, Zhenqin and Ramsundar, Bharath and Feinberg, Evan N and Gomes, Joseph and Geniesse, Caleb and Pappu, Aneesh S and Leswing, Karl and Pande, Vijay},
  title = {MoleculeNet: a benchmark for molecular machine learning},
  journal = {Chemical Science},
  volume = {9},
  number = {2},
  pages = {513--530},
  year = {2018},
  doi = {10.1039/C7SC02664A}
}

@article{moriwaki2018mordred,
  author = {Moriwaki, Hirotomo and Tian, Yu-Shi and Kawashita, Norihito and Takagi, Tatsuya},
  title = {Mordred: a molecular descriptor calculator},
  journal = {Journal of Cheminformatics},
  volume = {10},
  number = {1},
  pages = {4},
  year = {2018},
  doi = {10.1186/s13321-018-0258-y}
}

@article{lenzi2005producing,
  author = {Lenzi, Marcelo Kaminski and Cunningham, Michael F and Lima, Enrique Luis and Pinto, Jos{\'e} Carlos},
  title = {Producing bimodal molecular weight distribution polymer resins using living and conventional free-radical polymerization},
  journal = {Industrial \& Engineering Chemistry Research},
  volume = {44},
  number = {8},
  pages = {2568--2578},
  year = {2005},
  doi = {10.1021/ie0496479}
}

@article{romero2025benchmark,
  author = {Romero Pietrafesa, Tomás and Trigilio, Alessandro D and Marien, Yoshi W and Reyes, Pablo and Edeleva, Mariya and Asteasuain, Mariano and Van Steenberge, Paul HM and D’hooge, Dagmar R},
  title = {Benchmark cases and guidelines for kinetic Monte Carlo simulations with linear polymers},
  journal = {Industrial \& Engineering Chemistry Research},
  volume = {64},
  number = {20},
  pages = {9974--9992},
  year = {2025},
  doi = {10.1021/acs.iecr.5c00639}
}

@article{pedersen2021bayesian,
  author    = {Pedersen, Jack K. and Clausen, Christian M. and Krysiak, Olga A. and Xiao, Bin and Batchelor, Thomas A. A. and L{\"o}ffler, Tobias and Mints, Vladislav A. and Banko, Lars and Arenz, Matthias and Savan, Alan and Schuhmann, Wolfgang and Ludwig, Alfred and Rossmeisl, Jan},
  title     = {Bayesian Optimization of High-Entropy Alloy Compositions for Electrocatalytic Oxygen Reduction},
  journal   = {Angewandte Chemie International Edition},
  volume    = {60},
  number    = {45},
  pages     = {24144--24152},
  year      = {2021},
  month     = {November},
  publisher = {Wiley},
  doi       = {10.1002/anie.202108116}
}

@article{khatamsaz2023bayesian,
  author    = {Khatamsaz, Danial and Vela, Brent and Singh, Prashant and Johnson, Duane D. and Allaire, Douglas and Arróyave, Raymundo},
  title     = {Bayesian Optimization with Active Learning of Design Constraints Using an Entropy-Based Approach},
  journal   = {npj Computational Materials},
  volume    = {9},
  number    = {1},
  pages     = {49},
  year      = {2023},
  month     = {April},
  publisher = {Springer Nature},
  doi       = {10.1038/s41524-023-01006-7}
}

@article{shields2021bayesian,
  author    = {Shields, Benjamin J. and Stevens, Jason and Li, Jun and Parasram, Marvin and Damani, Farhan and Alvarado, Jesus I. Martinez and Janey, Jacob M. and Adams, Ryan P. and Doyle, Abigail G.},
  title     = {Bayesian Reaction Optimization as a Tool for Chemical Synthesis},
  journal   = {Nature},
  volume    = {590},
  number    = {7844},
  pages     = {89--96},
  year      = {2021},
  month     = {February},
  publisher = {Springer Nature},
  doi       = {10.1038/s41586-021-03213-y}
}

@article{taylor2023accelerated,
  author    = {Taylor, Connor J. and Felton, Kobi C. and Wigh, Daniel and Jeraal, Mohammed I. and Grainger, Rachel and Chessari, Gianni and Johnson, Christopher N. and Lapkin, Alexei A.},
  title     = {Accelerated Chemical Reaction Optimization Using Multi-Task Learning},
  journal   = {ACS Central Science},
  volume    = {9},
  number    = {5},
  pages     = {957--968},
  year      = {2023},
  month     = {April},
  publisher = {American Chemical Society},
  doi       = {10.1021/acscentsci.3c00050}
}

@article{dalal2024polymer,
  author    = {Dalal, Rishad J. and Oviedo, Felipe and Leyden, Michael C. and Reineke, Theresa M.},
  title     = {Polymer Design via SHAP and Bayesian Machine Learning Optimizes pDNA and CRISPR Ribonucleoprotein Delivery},
  journal   = {Chemical Science},
  volume    = {15},
  number    = {19},
  pages     = {7219--7228},
  year      = {2024},
  month     = {April},
  publisher = {Royal Society of Chemistry},
  doi       = {10.1039/D3SC06920F}
}

@article{jiang2025generative,
  author    = {Jiang, Shengli and Webb, Michael A.},
  title     = {Generative Active Learning across Polymer Architectures and Solvophobicities for Targeted Rheological Behavior},
  journal   = {npj Computational Materials},
  volume    = {12},
  number    = {1},
  pages     = {28},
  year      = {2026},
  month     = {December},
  publisher = {Springer Nature},
  doi       = {10.1038/s41524-025-01900-2}
}

@article{mckay1979comparison,
  author  = {McKay, M. D. and Beckman, R. J. and Conover, W. J.},
  title   = {A Comparison of Three Methods for Selecting Values of Input Variables in the Analysis of Output from a Computer Code},
  journal = {Technometrics},
  volume  = {21},
  number  = {2},
  pages   = {239--245},
  year    = {1979},
  doi     = {10.1080/00401706.1979.10489755}
}

@book{jolliffe2002principal,
  author     = {Jolliffe, Ian T.},
  title      = {Principal Component Analysis},
  edition    = {2},
  year       = {2002},
  publisher  = {Springer},
  address    = {New York},
  series     = {Springer Series in Statistics},
  isbn       = {978-0-387-95442-4},
  doi        = {10.1007/b98835}
}

@article{mcinnes2018umap,
  author  = {McInnes, Leland and Healy, John and Saul, Nathaniel and Grossberger, Lukas},
  title   = {{UMAP}: Uniform Manifold Approximation and Projection},
  journal = {Journal of Open Source Software},
  volume  = {3},
  number  = {29},
  pages   = {861},
  year    = {2018},
  doi     = {10.21105/joss.00861}
}

@article{gillespie1976general,
  author  = {Gillespie, Daniel T.},
  title   = {A General Method for Numerically Simulating the Stochastic Time Evolution of Coupled Chemical Reactions},
  journal = {Journal of Computational Physics},
  volume  = {22},
  number  = {4},
  pages   = {403--434},
  year    = {1976},
  doi     = {10.1016/0021-9991(76)90041-3}
}

@article{gillespie1977exact,
  author  = {Gillespie, Daniel T.},
  title   = {Exact Stochastic Simulation of Coupled Chemical Reactions},
  journal = {The Journal of Physical Chemistry},
  volume  = {81},
  number  = {25},
  pages   = {2340--2361},
  year    = {1977},
  doi     = {10.1021/j100540a008}
}

@article{runge1984density,
  author  = {Runge, Erich and Gross, E. K. U.},
  title   = {Density-Functional Theory for Time-Dependent Systems},
  journal = {Physical Review Letters},
  volume  = {52},
  number  = {12},
  pages   = {997--1000},
  year    = {1984},
  doi     = {10.1103/PhysRevLett.52.997}
}

@incollection{casida1995time,
  author    = {Casida, Mark E.},
  title     = {Time-Dependent Density Functional Response Theory for Molecules},
  booktitle = {Recent Advances in Density Functional Methods},
  volume    = {1},
  pages     = {155--192},
  year      = {1995},
  publisher = {World Scientific}
}

@article{ramakrishnan2014quantum,
  author  = {Ramakrishnan, Raghunathan and Dral, Pavlo O. and Rupp, Matthias and von Lilienfeld, O. Anatole},
  title   = {Quantum Chemistry Structures and Properties of 134 Kilo Molecules},
  journal = {Scientific Data},
  volume  = {1},
  pages   = {140022},
  year    = {2014},
  doi     = {10.1038/sdata.2014.22}
}

@article{delaney2004esol,
  author  = {Delaney, John S.},
  title   = {{ESOL}: Estimating Aqueous Solubility Directly from Molecular Structure},
  journal = {Journal of Chemical Information and Computer Sciences},
  volume  = {44},
  number  = {3},
  pages   = {1000--1005},
  year    = {2004},
  doi     = {10.1021/ci034243x}
}

@article{mobley2014freesolv,
  author  = {Mobley, David L. and Guthrie, J. Peter},
  title   = {{FreeSolv}: A Database of Experimental and Calculated Hydration Free Energies, with Input Files},
  journal = {Journal of Computer-Aided Molecular Design},
  volume  = {28},
  pages   = {711--720},
  year    = {2014},
  doi     = {10.1007/s10822-014-9747-x}
}

@inproceedings{gardner2014bayesian,
  author    = {Gardner, Jacob R. and Kusner, Matt J. and Xu, Zhixiang Eddie and Weinberger, Kilian Q. and Cunningham, John P.},
  title     = {Bayesian Optimization with Inequality Constraints},
  booktitle = {Proceedings of the 31st International Conference on Machine Learning},
  series    = {Proceedings of Machine Learning Research},
  volume    = {32},
  pages     = {937--945},
  year      = {2014},
  publisher = {PMLR}
}

@inproceedings{gelbart2014bayesian,
  author    = {Gelbart, Michael A. and Snoek, Jasper and Adams, Ryan P.},
  title     = {Bayesian Optimization with Unknown Constraints},
  booktitle = {Proceedings of the Thirtieth Conference on Uncertainty in Artificial Intelligence},
  pages     = {250--259},
  year      = {2014},
  publisher = {AUAI Press}
}

@article{fierens2015mama,
  author    = {Fierens, Stijn K. and D'hooge, Dagmar R. and Van Steenberge, Paul H. M. and Reyniers, Marie-Fran{\c{c}}oise and Marin, Guy B.},
  title     = {MAMA-SG1 Initiated Nitroxide Mediated Polymerization of Styrene: From Arrhenius Parameters to Model-Based Design},
  journal   = {Chemical Engineering Journal},
  volume    = {278},
  pages     = {407--420},
  year      = {2015},
  month     = {October},
  publisher = {Elsevier},
  doi       = {10.1016/j.cej.2014.09.024}
}

@article{chauvin2006nitroxide,
  author    = {Chauvin, Florence and Dufils, Pierre-Emmanuel and Gigmes, Didier and Guillaneuf, Yohann and Marque, Sylvain R. A. and Tordo, Paul and Bertin, Denis},
  title     = {Nitroxide-Mediated Polymerization: The Pivotal Role of the kd Value of the Initiating Alkoxyamine and the Importance of the Experimental Conditions},
  journal   = {Macromolecules},
  volume    = {39},
  number    = {16},
  pages     = {5238--5250},
  year      = {2006},
  month     = {August},
  publisher = {American Chemical Society},
  doi       = {10.1021/ma0527193}
}

@article{buback1995critically,
  author    = {Buback, Michael and Gilbert, Robert G. and Hutchinson, Robin A. and Klumperman, Bert and Kuchta, Frank-Dieter and Manders, Bart G. and O'Driscoll, Kenneth F. and Russell, Gregory T. and Schweer, Johannes},
  title     = {Critically Evaluated Rate Coefficients for Free-Radical Polymerization, 1. Propagation Rate Coefficient for Styrene},
  journal   = {Macromolecular Chemistry and Physics},
  volume    = {196},
  number    = {10},
  pages     = {3267--3280},
  year      = {1995},
  month     = {October},
  publisher = {Wiley},
  doi       = {10.1002/macp.1995.021961016}
}

@article{hui1972thermal,
  author    = {Hui, Albert W. and Hamielec, Archie E.},
  title     = {Thermal Polymerization of Styrene at High Conversions and Temperatures. An Experimental Study},
  journal   = {Journal of Applied Polymer Science},
  volume    = {16},
  number    = {3},
  pages     = {749--769},
  year      = {1972},
  month     = {March},
  publisher = {Wiley},
  doi       = {10.1002/app.1972.070160319}
}

@article{de2020roadmap,
  author    = {De Smit, Kyann and Marien, Yoshi W. and Edeleva, Mariya and Van Steenberge, Paul H. M. and D'hooge, Dagmar R.},
  title     = {Roadmap for Monomer Conversion and Chain Length-Dependent Termination Reactivity Algorithms in Kinetic Monte Carlo Modeling of Bulk Radical Polymerization},
  journal   = {Industrial \& Engineering Chemistry Research},
  volume    = {59},
  number    = {52},
  pages     = {22422--22439},
  year      = {2020},
  month     = {December},
  publisher = {American Chemical Society},
  doi       = {10.1021/acs.iecr.0c04328}
}

@article{achilias1992development,
  author    = {Achilias, D. S. and Kiparissides, C.},
  title     = {Development of a General Mathematical Framework for Modeling Diffusion-Controlled Free-Radical Polymerization Reactions},
  journal   = {Macromolecules},
  volume    = {25},
  number    = {14},
  pages     = {3739--3750},
  year      = {1992},
  month     = {July},
  publisher = {American Chemical Society},
  doi       = {10.1021/ma00040a021}
}

@article{buback1997termination,
  author    = {Buback, Michael and Kuchta, Frank-Dieter},
  title     = {Termination Kinetics of Free-Radical Polymerization of Styrene over an Extended Temperature and Pressure Range},
  journal   = {Macromolecular Chemistry and Physics},
  volume    = {198},
  number    = {5},
  pages     = {1455--1480},
  year      = {1997},
  month     = {May},
  publisher = {Wiley},
  doi       = {10.1002/macp.1997.021980513}
}

@article{gilbert1996critically,
  author    = {Gilbert, Robert G.},
  title     = {Critically-Evaluated Propagation Rate Coefficients in Free Radical Polymerizations I. Styrene and Methyl Methacrylate (Technical Report)},
  journal   = {Pure and Applied Chemistry},
  volume    = {68},
  number    = {7},
  pages     = {1491--1494},
  year      = {1996},
  month     = {July},
  publisher = {De Gruyter},
  doi       = {10.1351/pac199668071491}
}

@article{hatada1985evidence,
  author    = {Hatada, Koichi and Kitayama, Tatsuki and Masuda, Fiji},
  title     = {Evidence for Disproportionation in Termination Reaction of Styrene Polymerization by $\alpha$, $\alpha$'-Azobisisobutyronitrile},
  journal   = {Polymer Journal},
  volume    = {17},
  number    = {8},
  pages     = {985--989},
  year      = {1985},
  month     = {August},
  publisher = {Springer Nature},
  doi       = {10.1295/polymj.17.985}
}

@article{yamada1992esr,
  author    = {Yamada, Bunichiro and Kageoka, Masakazu and Otsu, Takayuki},
  title     = {ESR Study of the Radical Polymerization of Styrene: 5. Temperature Dependence of Propagation and Termination Rate Constants over a Wide Temperature Range},
  journal   = {Polymer Bulletin},
  volume    = {29},
  number    = {3},
  pages     = {385--392},
  year      = {1992},
  month     = {September},
  publisher = {Springer},
  doi       = {10.1007/BF00944835}
}

@article{kapfenstein2001novel,
  author    = {Kapfenstein-Doak, Heidi and Barner-Kowollik, Christopher and Davis, Thomas P. and Schweer, Johannes},
  title     = {A Novel Method for the Measurement of Chain Transfer to Monomer Constants in Styrene Homopolymerizations: The Pulsed Laser Rotating Reactor Assembly},
  journal   = {Macromolecules},
  volume    = {34},
  number    = {9},
  pages     = {2822--2829},
  year      = {2001},
  month     = {April},
  publisher = {American Chemical Society},
  doi       = {10.1021/ma001871w}
}

@article{eperon2016perovskite,
  author     = {Eperon, Giles E and Leijtens, Tomas and Bush, Kevin A and Prasanna, Rohit and Green, Thomas and Wang, Jacob Tse-Wei and McMeekin, David P and Volonakis, George and Milot, Rebecca L and May, Richard and others},
  title      = {Perovskite-perovskite tandem photovoltaics with optimized band gaps},
  journal    = {Science},
  volume     = {354},
  number     = {6314},
  pages      = {861--865},
  year       = {2016},
  month      = {November},
  publisher  = {American Association for the Advancement of Science},
  doi        = {10.1126/science.aaf9717}
}

@article{ashby1993materials,
  author    = {Ashby, Michael F and Cebon, David},
  title     = {Materials selection in mechanical design},
  journal   = {Le Journal de Physique IV},
  volume    = {3},
  number    = {C7},
  pages     = {C7--1--C7--9},
  year      = {1993},
  month     = {November},
  publisher = {EDP Sciences},
  doi       = {10.1051/jp4:1993701}
}

@incollection{schofield2015critical,
  author     = {Schofield, Timothy and Robbins, David and Mir{\'o}-Quesada, Guillermo},
  title      = {Critical quality attributes, specifications, and control strategy},
  booktitle  = {Quality by Design for Biopharmaceutical Drug Product Development},
  pages      = {511--535},
  year       = {2015},
  month      = {April},
  publisher  = {Springer},
  doi        = {10.1007/978-1-4939-2316-8_21}
}

@book{ulrich2020product,
  author      = {Ulrich, Karl T. and Eppinger, Steven D. and Yang, Maria C.},
  title       = {Product Design and Development},
  edition     = {7},
  year        = {2020},
  publisher   = {McGraw-Hill Education},
  address     = {New York, NY},
  isbn        = {9781260043655}
}

@article{bichon2008efficient,
  author     = {Bichon, Barron J and Eldred, Michael S and Swiler, Laura Painton and Mahadevan, Sandaran and McFarland, John M},
  title      = {Efficient global reliability analysis for nonlinear implicit performance functions},
  journal    = {AIAA Journal},
  volume     = {46},
  number     = {10},
  pages      = {2459--2468},
  year       = {2008},
  month      = {October},
  doi        = {10.2514/1.34321}
}

@article{ashby2004selection,
  author     = {Ashby, MF and Br{\'e}chet, YJM and Cebon, D and Salvo, L},
  title      = {Selection strategies for materials and processes},
  journal    = {Materials \& Design},
  volume     = {25},
  number     = {1},
  pages      = {51--67},
  year       = {2004},
  month      = {February},
  publisher  = {Elsevier},
  doi        = {10.1016/S0261-3069(03)00159-6}
}

@article{zeni2025generative,
  author     = {Zeni, Claudio and Pinsler, Robert and Z{\"u}gner, Daniel and Fowler, Andrew and Horton, Matthew and Fu, Xiang and Wang, Zilong and Shysheya, Aliaksandra and Crabb{\'e}, Jonathan and Ueda, Shoko and others},
  title      = {A generative model for inorganic materials design},
  journal    = {Nature},
  volume     = {639},
  number     = {8055},
  pages      = {624--632},
  year       = {2025},
  month      = {March},
  publisher  = {Nature Publishing Group},
  doi        = {10.1038/s41586-025-08628-5}
}

@article{renz2024diverse,
  author     = {Renz, Philipp and Luukkonen, Sohvi and Klambauer, G{\"u}nter},
  title      = {Diverse hits in de novo molecule design: Diversity-based comparison of goal-directed generators},
  journal    = {Journal of Chemical Information and Modeling},
  volume     = {64},
  number     = {15},
  pages      = {5756--5761},
  year       = {2024},
  month      = {July},
  publisher  = {ACS Publications},
  doi        = {10.1021/acs.jcim.4c00519}
}

@article{hase2018chimera,
  author     = {H{\"a}se, Florian and Roch, Lo{\"i}c M. and Aspuru-Guzik, Al{\'a}n},
  title      = {Chimera: enabling hierarchy based multi-objective optimization for self-driving laboratories},
  journal    = {Chemical Science},
  year       = {2018},
  volume     = {9},
  pages      = {7642--7655},
  doi        = {10.1039/C8SC02239A},
  publisher  = {Royal Society of Chemistry}
}

@article{konakovic2020diversity,
  author     = {Konakovic Lukovic, Mina and Tian, Yunsheng and Matusik, Wojciech},
  title      = {Diversity-guided multi-objective Bayesian optimization with batch evaluations},
  journal    = {Advances in Neural Information Processing Systems},
  volume     = {33},
  pages      = {17708--17720},
  year       = {2020},
  publisher  = {Curran Associates, Inc.},
  doi        = {}
}

@article{maus2022discovering,
  author     = {Maus, Natalie and Wu, Kaiwen and Eriksson, David and Gardner, Jacob},
  title      = {Discovering Many Diverse Solutions with Bayesian Optimization},
  journal    = {arXiv preprint arXiv:2210.10953},
  year       = {2022},
  month      = {October},
  doi        = {10.48550/arXiv.2210.10953},
  url        = {https://arxiv.org/abs/2210.10953}
}

@article{attia2020closed,
  title   = {Closed-loop optimization of fast-charging protocols for batteries with machine learning},
  author  = {Attia, Peter M. and Grover, Aditya and Jin, Norman and Severson, Kristen A. and Markov, Todor M. and Liao, Yang-Hung and Chen, Michael H. and Cheong, Bryan and Perkins, Nicholas and Yang, Zi and Herring, Patrick K. and Aykol, Muratahan and Harris, Stephen J. and Braatz, Richard D. and Ermon, Stefano and Chueh, William C.},
  journal = {Nature},
  volume  = {578},
  pages   = {397--402},
  year    = {2020},
  month   = {February},
  doi     = {10.1038/s41586-020-1994-5},
  publisher = {Springer Nature}
}

@article{kusne2020fly,
  title   = {On-the-fly closed-loop materials discovery via Bayesian active learning},
  author  = {Kusne, A. Gilad and Yu, Heshan and Wu, Changming and Zhang, Huairuo and Hattrick-Simpers, Jason and DeCost, Brian and Sarker, Suchismita and Oses, Corey and Toher, Cormac and Curtarolo, Stefano and Davydov, Albert V. and Agarwal, Ritesh and Bendersky, Leonid A. and Li, Mo and Mehta, Apurva and Takeuchi, Ichiro},
  journal = {Nature Communications},
  volume  = {11},
  number  = {1},
  pages   = {5966},
  year    = {2020},
  month   = {November},
  doi     = {10.1038/s41467-020-19597-w},
  publisher = {Springer Nature}
}

@article{szymanski2023autonomous,
  title   = {An autonomous laboratory for the accelerated synthesis of inorganic materials},
  author  = {Szymanski, Nathan J. and Rendy, Bernardus and Fei, Yuxing and Kumar, Rishi E. and He, Tanjin and Milsted, David and McDermott, Matthew J. and Gallant, Max and Cubuk, Ekin Dogus and Merchant, Amil and others},
  journal = {Nature},
  volume  = {624},
  number  = {7990},
  pages   = {86--91},
  year    = {2023},
  month   = {December},
  doi     = {10.1038/s41586-023-06734-w},
  publisher = {Springer Nature}
}

@article{ren2018accelerated,
  title   = {Accelerated discovery of metallic glasses through iteration of machine learning and high-throughput experiments},
  author  = {Ren, Fang and Ward, Logan and Williams, Travis and Laws, Kevin J. and Wolverton, Christopher and Hattrick-Simpers, Jason and Mehta, Apurva},
  journal = {Science Advances},
  volume  = {4},
  number  = {4},
  pages   = {eaaq1566},
  year    = {2018},
  month   = {April},
  doi     = {10.1126/sciadv.aaq1566},
  publisher = {American Association for the Advancement of Science}
}

@article{tamasi2022machine,
  title   = {Machine learning on a robotic platform for the design of polymer--protein hybrids},
  author  = {Tamasi, Matthew J. and Patel, Roshan A. and Borca, Carlos H. and Kosuri, Shashank and Mugnier, Heloise and Upadhya, Rahul and Murthy, N. Sanjeeva and Webb, Michael A. and Gormley, Adam J.},
  journal = {Advanced Materials},
  volume  = {34},
  number  = {30},
  pages   = {2201809},
  year    = {2022},
  month   = {May},
  doi     = {10.1002/adma.202201809},
  publisher = {Wiley-VCH}
}

@article{an2024active,
  title   = {Active learning of the thermodynamics-dynamics trade-off in protein condensates},
  author  = {An, Yaxin and Webb, Michael A. and Jacobs, William M.},
  journal = {Science Advances},
  volume  = {10},
  number  = {1},
  pages   = {eadj2448},
  year    = {2024},
  month   = {January},
  doi     = {10.1126/sciadv.adj2448},
  publisher = {American Association for the Advancement of Science}
}

@article{jamil2013literature,
  title   = {A literature survey of benchmark functions for global optimisation problems},
  author  = {Jamil, Momin and Yang, Xin-She},
  journal = {International Journal of Mathematical Modelling and Numerical Optimisation},
  volume  = {4},
  number  = {2},
  pages   = {150--194},
  year    = {2013},
  month   = {July},
  doi     = {10.1504/IJMMNO.2013.055204},
  publisher = {Inderscience Publishers Ltd}
}

@article{roseli2017origin,
  title = {Origin of the excited-state absorption spectrum of polythiophene},
  author = {Roseli, Ras Baizureen and Tapping, Patrick C. and Kee, Tak W.},
  journal = {The Journal of Physical Chemistry Letters},
  volume = {8},
  number = {13},
  pages = {2806--2811},
  year = {2017},
  month = {July},
  doi = {10.1021/acs.jpclett.7b01053},
  publisher = {ACS Publications}
}

@misc{rdkit,
  title = {RDKit: Open-source cheminformatics},
  author = {Landrum, Greg and RDKit Contributors},
  howpublished = {\url{https://www.rdkit.org}},
  year = {2006},
  note = {https://doi.org/10.5281/zenodo.591637}
}

@article{wang2020etkdgv3,
  title = {Improving Conformer Generation for Small Rings and Macrocycles Based on Distance Geometry and Experimental Torsional-Angle Preferences},
  author = {Wang, Shuzhe and Witek, Jagna and Landrum, Gregory A. and Riniker, Sereina},
  journal = {Journal of Chemical Information and Modeling},
  volume = {60},
  number = {4},
  pages = {2044--2058},
  year = {2020},
  month = {March},
  doi = {10.1021/acs.jcim.0c00025},
  publisher = {American Chemical Society}
}

@article{halgren1996mmff94,
  title = {Merck molecular force field. I. Basis, form, scope, parameterization, and performance of MMFF94},
  author = {Halgren, Thomas A.},
  journal = {Journal of Computational Chemistry},
  volume = {17},
  number = {5-6},
  pages = {490--519},
  year = {1996},
  month = {April},
  doi = {10.1002/(SICI)1096-987X(199604)17:5/6<490::AID-JCC1>3.0.CO;2-P},
  publisher = {Wiley}
}

@article{neese2020orca,
  title = {The ORCA quantum chemistry program package},
  author = {Neese, Frank and Wennmohs, Frank and Becker, Ute and Riplinger, Christoph},
  journal = {The Journal of Chemical Physics},
  volume = {152},
  number = {22},
  pages = {224108},
  year = {2020},
  month = {June},
  doi = {10.1063/5.0004608},
  publisher = {AIP Publishing}
}

@article{bannwarth2019gfn2,
  title = {GFN2-xTB—An Accurate and Broadly Parametrized Self-Consistent Tight-Binding Quantum Chemical Method with Multipole Electrostatics and Density-Dependent Dispersion Contributions},
  author = {Bannwarth, Christoph and Ehlert, Sebastian and Grimme, Stefan},
  journal = {Journal of Chemical Theory and Computation},
  volume = {15},
  number = {3},
  pages = {1652--1671},
  year = {2019},
  month = {February},
  doi = {10.1021/acs.jctc.8b01176},
  publisher = {American Chemical Society}
}

@article{ehlert2021alpb,
  title = {Robust and Efficient Implicit Solvation Model for Fast Semiempirical Methods},
  author = {Ehlert, Sebastian and Stahn, Marcel and Spicher, Sebastian and Grimme, Stefan},
  journal = {Journal of Chemical Theory and Computation},
  volume = {17},
  number = {7},
  pages = {4250--4261},
  year = {2021},
  month = {June},
  doi = {10.1021/acs.jctc.1c00471},
  publisher = {American Chemical Society}
}

@article{muller2023wb97x3c,
  title={$\omega$B97X-3c: A composite range-separated hybrid DFT method with a molecule-optimized polarized valence double-$\zeta$ basis set},
  author = {Müller, Marcel and Hansen, Andreas and Grimme, Stefan},
  journal = {The Journal of Chemical Physics},
  volume = {158},
  number = {1},
  pages = {014103},
  year = {2023},
  month = {January},
  doi = {10.1063/5.0133026},
  publisher = {AIP Publishing}
}

@article{barone1998cpcm,
  title = {Quantum Calculation of Molecular Energies and Energy Gradients in Solution by a Conductor Solvent Model},
  author = {Barone, Vincenzo and Cossi, Maurizio},
  journal = {The Journal of Physical Chemistry A},
  volume = {102},
  number = {11},
  pages = {1995--2001},
  year = {1998},
  month = {March},
  doi = {10.1021/jp9714387},
  publisher = {American Chemical Society}
}

@article{gentekos2019controlling,
  author  = {Gentekos, Dillon T. and Sifri, Renee M. and Fors, Brett P.},
  title   = {Controlling polymer properties through the shape of the molecular-weight distribution},
  journal = {Nature Reviews Materials},
  volume  = {4},
  number  = {12},
  pages   = {761--774},
  year    = {2019},
  doi     = {10.1038/s41578-019-0138-8}
}

@article{gentekos2016beyond,
  author  = {Gentekos, Dillon T. and Dupuis, Lauren N. and Fors, Brett P.},
  title   = {Beyond dispersity: Deterministic control of polymer molecular weight distribution},
  journal = {Journal of the American Chemical Society},
  volume  = {138},
  number  = {6},
  pages   = {1848--1851},
  year    = {2016},
  doi     = {10.1021/jacs.5b13565}
}

@article{li2018tuning,
  author  = {Li, Haichen and Collins, Christopher R. and Ribelli, Thomas G. and Matyjaszewski, Krzysztof and Gordon, Geoffrey J. and Kowalewski, Tomasz and Yaron, David J.},
  title   = {Tuning the molecular weight distribution from atom transfer radical polymerization using deep reinforcement learning},
  journal = {Molecular Systems Design \& Engineering},
  volume  = {3},
  number  = {3},
  pages   = {496--508},
  year    = {2018},
  doi     = {10.1039/C7ME00131B}
}

@article{walsh2020general,
  author = {Walsh, Dylan J. and Schinski, Devin A. and Schneider, Robert A. and Guironnet, Damien},
  title = {General route to design polymer molecular weight distributions through flow chemistry},
  journal = {Nature Communications},
  volume = {11},
  pages = {3094},
  year = {2020},
  doi = {10.1038/s41467-020-16874-6}
}

@article{liu2020inverse,
  author = {Liu, Hong and Xue, Yao-Hong and Zhu, You-Liang and Gu, Feng-Long and Lu, Zhong-Yuan},
  title = {Inverse Design of Molecular Weight Distribution in Controlled Polymerization via a One-Pot Reaction Strategy},
  journal = {Macromolecules},
  volume = {53},
  number = {15},
  pages = {6409--6419},
  year = {2020},
  doi = {10.1021/acs.macromol.0c01383}
}

@article{zhou2024active,
  author = {Zhou, Haifan and Fang, Yue and Gao, Hanyu},
  title = {Using Active Learning for the Computational Design of Polymer Molecular Weight Distributions},
  journal = {ACS Engineering Au},
  volume = {4},
  number = {2},
  pages = {231--240},
  year = {2024},
  doi = {10.1021/acsengineeringau.3c00056}
}

@article{fiosina2025evolutionary,
  author = {Fiosina, Jelena and Sievers, Philipp and Drache, Marco and Beuermann, Sabine},
  title = {Machine learning supported evolutionary optimization for multi-objective reverse engineering of radical polymerizations},
  journal = {Computers \& Chemical Engineering},
  volume = {199},
  pages = {109125},
  year = {2025},
  doi = {10.1016/j.compchemeng.2025.109125}
}

@article{yoo1999molecular,
  author = {Yoo, Kee-Youn and Jeong, Boong-Goon and Rhee, Hyun-Ku},
  title = {Molecular Weight Distribution Control in a Batch Polymerization Reactor Using the On-Line Two-Step Method},
  journal = {Industrial \& Engineering Chemistry Research},
  volume = {38},
  number = {12},
  pages = {4805--4814},
  year = {1999},
  doi = {10.1021/ie980799b}
}

@article{liu2020comprehensive,
  author = {Liu, Ke and Corrigan, Nathaniel and Postma, Almar and Moad, Graeme and Boyer, Cyrille},
  title = {A Comprehensive Platform for the Design and Synthesis of Polymer Molecular Weight Distributions},
  journal = {Macromolecules},
  volume = {53},
  number = {20},
  pages = {8867--8882},
  year = {2020},
  doi = {10.1021/acs.macromol.0c01954}
}

@article{fang2026unified,
  author = {Fang, Yue and Jin, Yang and Gao, Hanyu},
  title = {A unified kinetic Monte Carlo and Bayesian optimization framework for the circular design of poly(methyl methacrylate): From synthesis to recycling},
  journal = {Chemical Engineering Journal},
  volume = {538},
  pages = {176625},
  year = {2026},
  doi = {10.1016/j.cej.2026.176625}
}

@article{mackenzie2024computer,
  author = {MacKenzie, Anja and Schneider, Jakob and Meyer, Jan and Loschen, Christoph},
  title = {Computer aided recipe design: optimization of polydisperse chemical mixtures using molecular descriptors},
  journal = {Reaction Chemistry \& Engineering},
  volume = {9},
  pages = {1061--1076},
  year = {2024},
  doi = {10.1039/D3RE00601H}
}

@article{nozaki2025impact,
  title = {Impact of aromatic to quinoidal transformation on the degradation kinetics of imine-based semiconducting polymers},
  author = {Nozaki, Naoya and Uva, Azalea and Iwahashi, Takashi and Matsumoto, Hidetoshi and Tran, Helen and Ashizawa, Minoru},
  journal = {RSC Applied Polymers},
  volume = {3},
  number = {1},
  pages = {257--267},
  year = {2025},
  publisher = {Royal Society of Chemistry},
  doi = {10.1039/D4LP00310A}
}

@article{aldeghi2022roughness,
  title   = {Roughness of Molecular Property Landscapes and Its Impact on Modellability},
  author  = {Aldeghi, Matteo and Graff, David E. and Frey, Nathan and Morrone, Joseph A. and Pyzer-Knapp, Edward O. and Jordan, Kirk E. and Coley, Connor W.},
  journal = {Journal of Chemical Information and Modeling},
  year    = {2022},
  volume  = {62},
  number  = {19},
  pages   = {4660--4671},
  doi     = {10.1021/acs.jcim.2c00903}
}

@book{rasmussen2006gaussian,
  title     = {Gaussian Processes for Machine Learning},
  author    = {Rasmussen, Carl Edward and Williams, Christopher K. I.},
  year      = {2006},
  publisher = {MIT Press},
  address   = {Cambridge, MA},
  isbn      = {026218253X},
}

\end{document}


\title{Supplementary Information \\
for\\
Range-Aware Bayesian Optimization for Discovering Diverse Designs within Target Property Windows}

\author{Shengli Jiang$^1$, Jason Wu$^1$, Charles M. Schroeder$^1$, and Michael A. Webb$^{1*}$\\
\\
{\small $^1$Department of Chemical and Biological Engineering, Princeton University, Princeton, NJ 08540}\\
{\small $^*$Corresponding Author: mawebb@princeton.edu}}

\date{}
\maketitle
 
\tableofcontents
\newpage
 
\section{Synthetic Functions}
 
\paragraph{Branin function (2D).}
Defined on the domain $x_1 \in [-5,10]$ and $x_2 \in [0,15]$, the Branin function is
\begin{equation}
f(x_1, x_2) =
\left(
x_2 - \frac{5.1}{4\pi^2}x_1^2 + \frac{5}{\pi}x_1 - 6
\right)^2
+ 10\left(1 - \frac{1}{8\pi}\right)\cos(x_1) + 10.
\end{equation}
This is a classical non-convex benchmark with multiple global minima.
 
\paragraph{Hartmann-3D function (3D).}
Defined on the unit cube $[0,1]^3$, the Hartmann-3D function is
\begin{equation}
f(\mathbf{x}) =
-\sum_{i=1}^{4}\alpha_i
\exp\!\left(
-\sum_{j=1}^{3} A_{ij}(x_j - P_{ij})^2
\right),
\end{equation}
where
\[
\alpha = (1.0,\,1.2,\,3.0,\,3.2),
\qquad
A =
\begin{bmatrix}
3.0 & 10 & 30 \\
0.1 & 10 & 35 \\
3.0 & 10 & 30 \\
0.1 & 10 & 35
\end{bmatrix},
\qquad
P = 10^{-4}
\begin{bmatrix}
3689 & 1170 & 2673 \\
4699 & 4387 & 7470 \\
1091 & 8732 & 5547 \\
381  & 5743 & 8828
\end{bmatrix}.
\]
This is a multimodal benchmark with several local minima.
 
\paragraph{Ackley-5D function (5D).}
Defined on the hypercube $[-5,5]^5$, the Ackley function in $d$ dimensions is
\begin{equation}
f(\mathbf{x}) =
-a \exp\!\left(
-b \sqrt{\frac{1}{d}\sum_{i=1}^{d} x_i^2}
\right)
-
\exp\!\left(
\frac{1}{d}\sum_{i=1}^{d}\cos(c x_i)
\right)
+ a + e,
\end{equation}
with parameters $a=20$, $b=0.2$, $c=2\pi$, and here $d=5$.
It has a global minimum at the origin and a broad outer region that makes optimization challenging.
 
\paragraph{Layeb-6 function (6D).}
The Layeb-6 function in $d$ dimensions is defined by
\begin{equation}
f(\mathbf{x}) =
\sum_{i=1}^{d-1}
\left|
\cos\!\left(\sqrt{x_i^2 + x_{i+1}^2}\right)\sin(x_{i+1})
+ \cos(x_{i+1}) + 1
\right|^{0.1},
\end{equation}
and here we use $d=6$. The function couples consecutive variables through nonlinear oscillatory terms and provides a multimodal benchmark for inverse-design search.
 
\section{Pool-Based Datasets}
 
Table~\ref{tab:pool_datasets} summarizes the pool-based datasets used in this work, including the number of candidates, the input representation, and the target output property or properties used for inverse design.
 
\begin{table}[htbp]
  \centering
  \caption{\textbf{Summary of pool-based datasets used in this study.}}
  \label{tab:pool_datasets}
  \footnotesize
  \setlength{\tabcolsep}{4pt}
  \renewcommand{\arraystretch}{1.2}
  \begin{tabularx}{\textwidth}{@{}
    >{\raggedright\arraybackslash}p{1.4cm}
    >{\centering\arraybackslash}p{1.4cm}
    >{\raggedright\arraybackslash}X
    >{\centering\arraybackslash}p{1.1cm}
    >{\raggedright\arraybackslash}p{3.4cm}
    >{\centering\arraybackslash}p{1.1cm}
    >{\raggedright\arraybackslash}p{2.2cm}
    @{}}
    \toprule
    \textbf{Dataset} & \textbf{Candidates} & \textbf{Design-space representation} & \textbf{Input dim.} & \textbf{Target property} & \textbf{Output dim.} & \textbf{Output units} \\
    \midrule
    NS
      & 1{,}997
      & Normalized synthesis variables
      & 4
      & Nanoparticle radius and polydispersity index
      & 2
      & nm; -- \\

    IDP
      & 2{,}031
      & Polypeptide descriptors
      & 30
      & Radius of gyration, second virial coefficient, expenditure density
      & 3
      & \AA{}; --; g\,mL$^{-1}$ \\

    TopoRg
      & 1{,}340
      & VAE latent vector
      & 8
      & Mean single-chain radius of gyration
      & 1
      & $\sigma$ \\

    BACE
      & 1{,}513
      & Mordred descriptors with PCA
      & 62
      & pIC50
      & 1
      & -- \\

    ESOL
      & 1{,}122
      & Mordred descriptors with PCA
      & 72
      & Aqueous solubility
      & 1
      & $\log_{10}$(mol/L) \\

    FreeSolv
      & 642
      & Mordred descriptors with PCA
      & 65
      & Hydration free energy
      & 1
      & kcal/mol \\

    Lipo
      & 4{,}102
      & Mordred descriptors with PCA
      & 92
      & Octanol-water distribution coefficient
      & 1
      & $\log D$ \\

    QM9
      & 6{,}693
      & Mordred descriptors with PCA
      & 69
      & HOMO and LUMO
      & 2
      & eV \\
    \bottomrule
  \end{tabularx}
\end{table}

 \newpage
\section{Across-Target Variance of the Diversity Score}
 
\begin{figure}[h!]
    \centering
    \includegraphics[width=\linewidth]{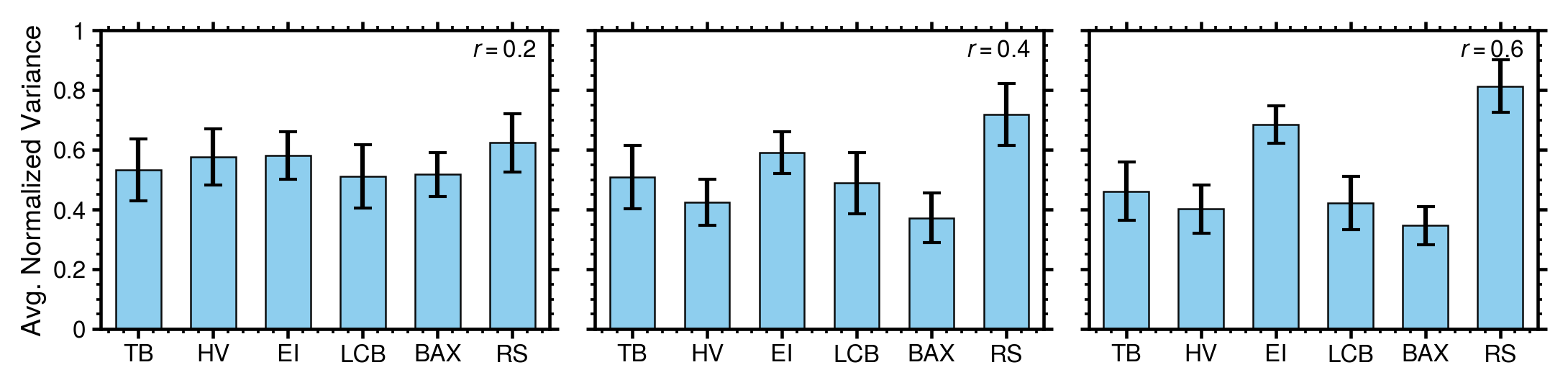}
    \caption{\textbf{Across-target variance of the normalized diversity score across acquisition functions for three tolerance ratios.}
Each panel shows, for one tolerance ratio ($r = 0.2, 0.4, 0.6$), the variance of the normalized diversity score ($D_\mathrm{c}$ for continuous tasks, $D_\mathrm{d}$ for discrete tasks) across the five targets, averaged over datasets (bars) with standard error (error bars). Values are normalized by the maximum across-target variance per dataset. Lower values indicate more even discovery across the five targets.}
    \label{fig:si-target-variance}
\end{figure}

 \newpage
\section{Additional Benchmark Results}

 \begin{figure}[p]
    \centering
    \includegraphics[width=\linewidth]{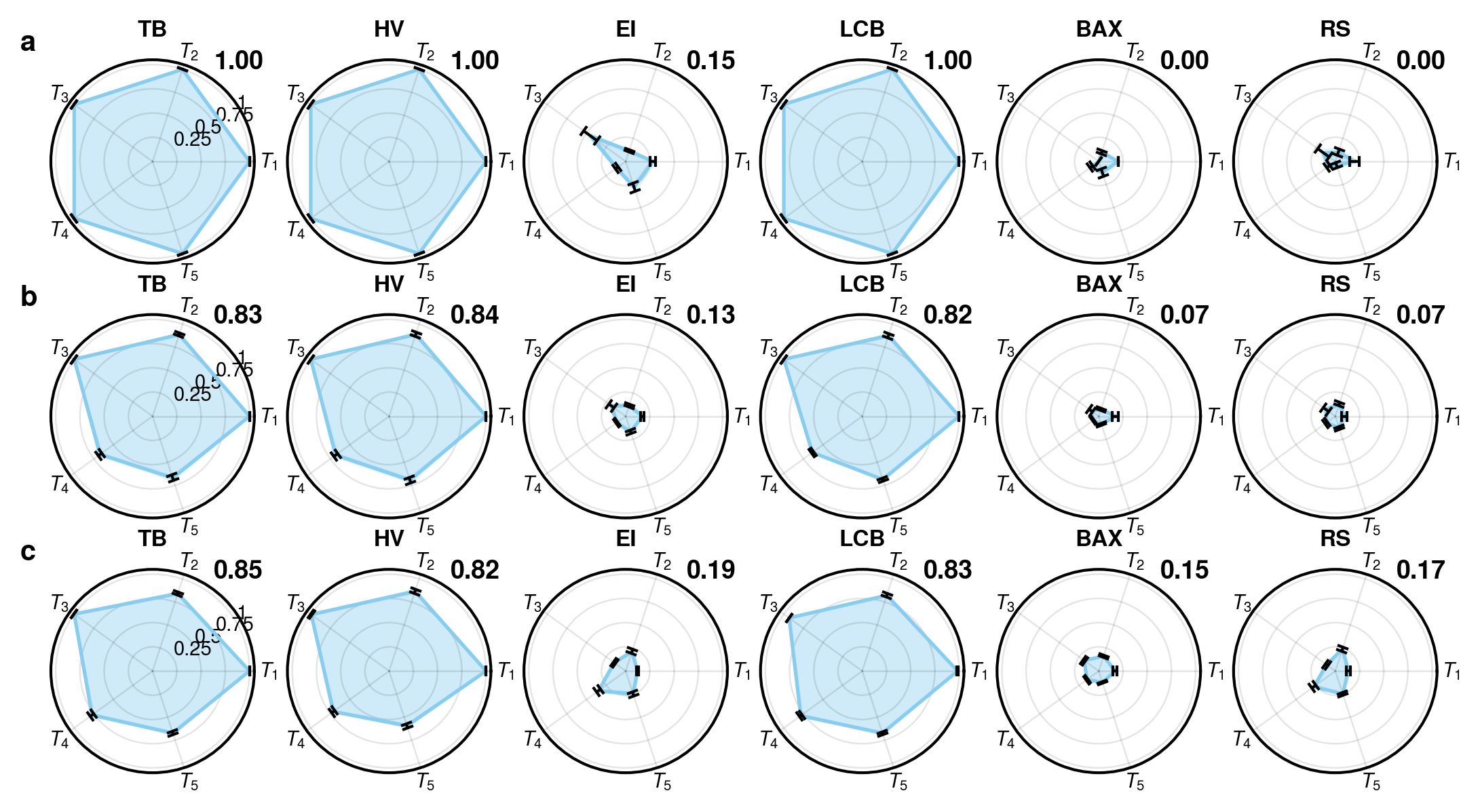}
    \caption{\textbf{Normalized diversity score across acquisition functions for the NS benchmark.}
    Radar plots of the normalized diversity score $D_\mathrm{d}$ for five targets $T_1$ through $T_5$ on the NS benchmark at tolerance ratios (a) $r = 0.2$, (b) $r = 0.4$, and (c) $r = 0.6$. }
    \label{fig:si-radar-ns}
\end{figure}

 \begin{figure}[p]
    \centering
    \includegraphics[width=\linewidth]{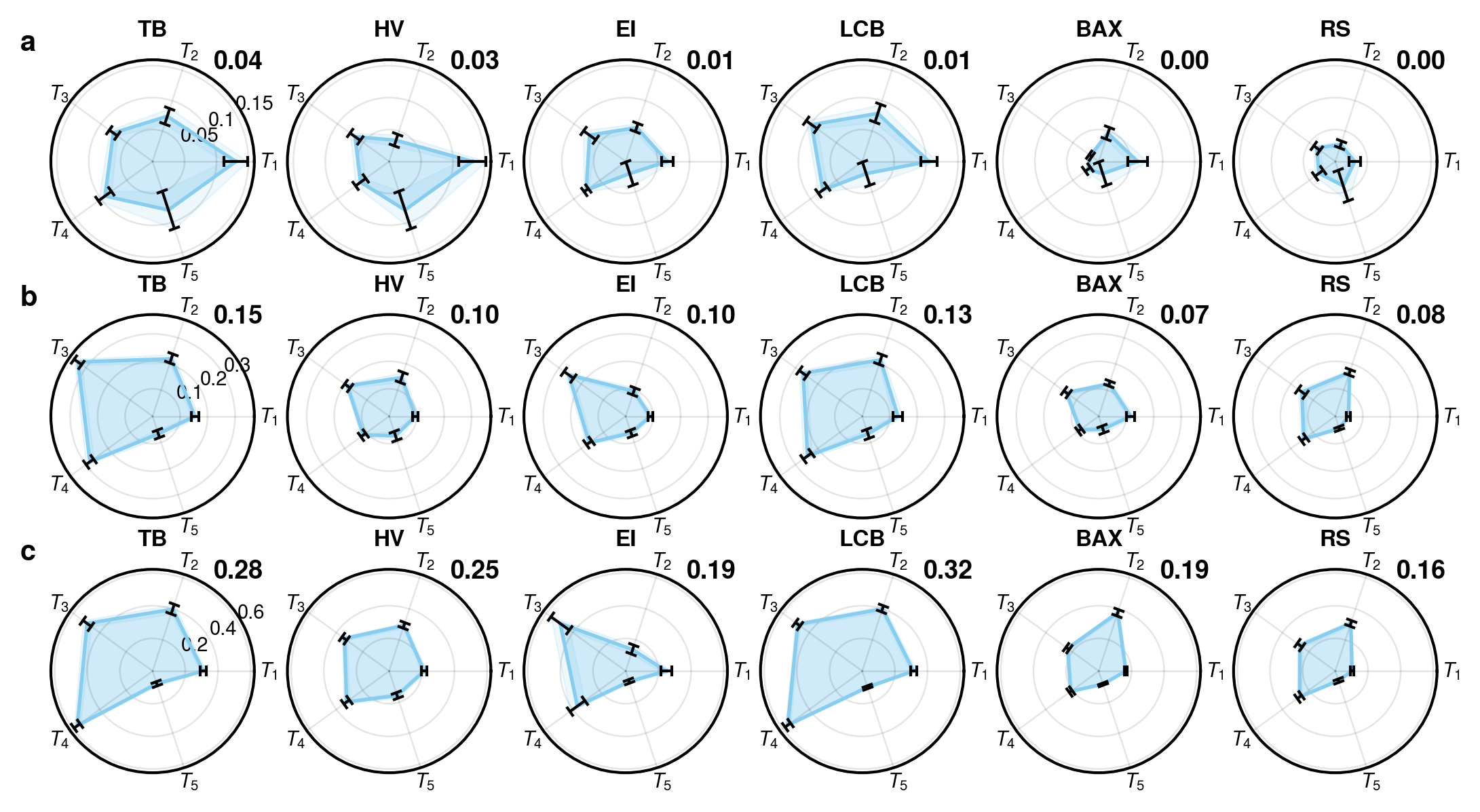}
    \caption{\textbf{Normalized diversity score across acquisition functions for the QM9 benchmark.}
    Radar plots of the normalized diversity score $D_\mathrm{d}$ for five targets $T_1$ through $T_5$ on the QM9 benchmark at tolerance ratios (a) $r = 0.2$, (b) $r = 0.4$, and (c) $r = 0.6$. }
    \label{fig:si-radar-qm9}
\end{figure}

 \begin{figure}[p]
    \centering
    \includegraphics[width=\linewidth]{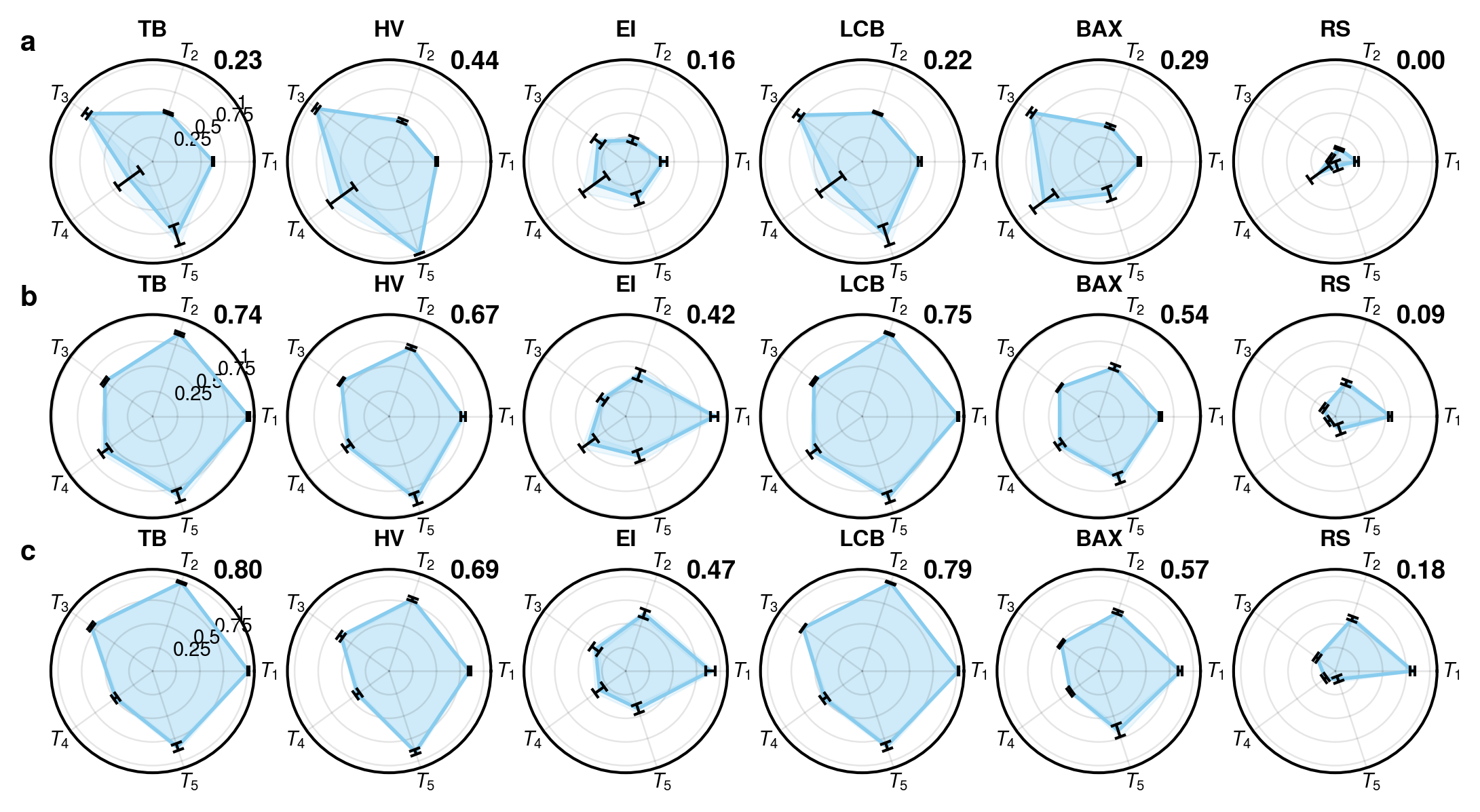}
    \caption{\textbf{Normalized diversity score across acquisition functions for the IDP benchmark.}
    Radar plots of the normalized diversity score $D_\mathrm{d}$ for five targets $T_1$ through $T_5$ on the IDP benchmark at tolerance ratios (a) $r = 0.2$, (b) $r = 0.4$, and (c) $r = 0.6$. }
    \label{fig:si-radar-propensity}
\end{figure}

 \begin{figure}[p]
    \centering
    \includegraphics[width=\linewidth]{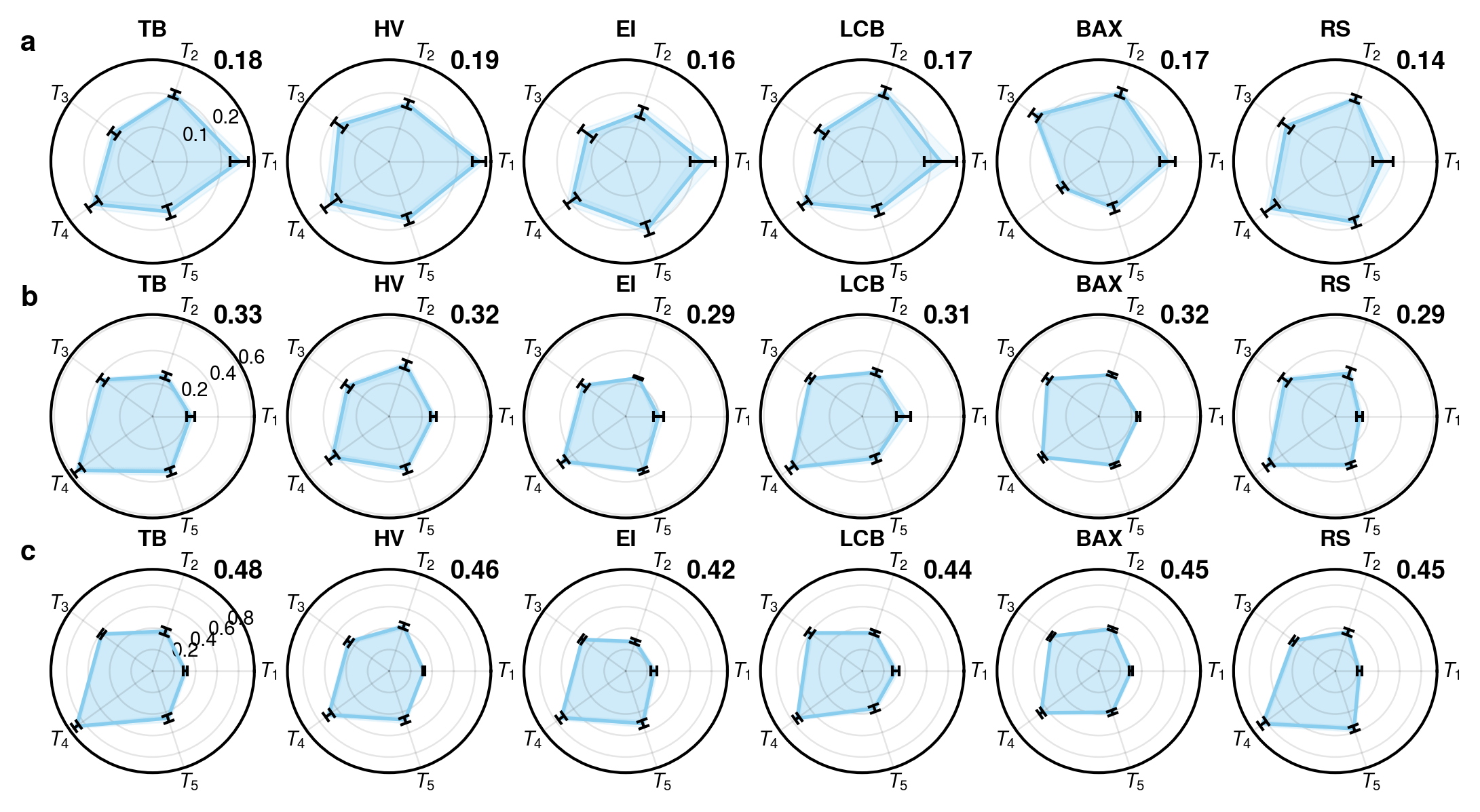}
    \caption{\textbf{Normalized diversity score across acquisition functions for the BACE benchmark.}
    Radar plots of the normalized diversity score $D_\mathrm{d}$ for five targets $T_1$ through $T_5$ on the BACE benchmark at tolerance ratios (a) $r = 0.2$, (b) $r = 0.4$, and (c) $r = 0.6$. }
    \label{fig:si-radar-bace}
\end{figure}

 \begin{figure}[p]
    \centering
    \includegraphics[width=\linewidth]{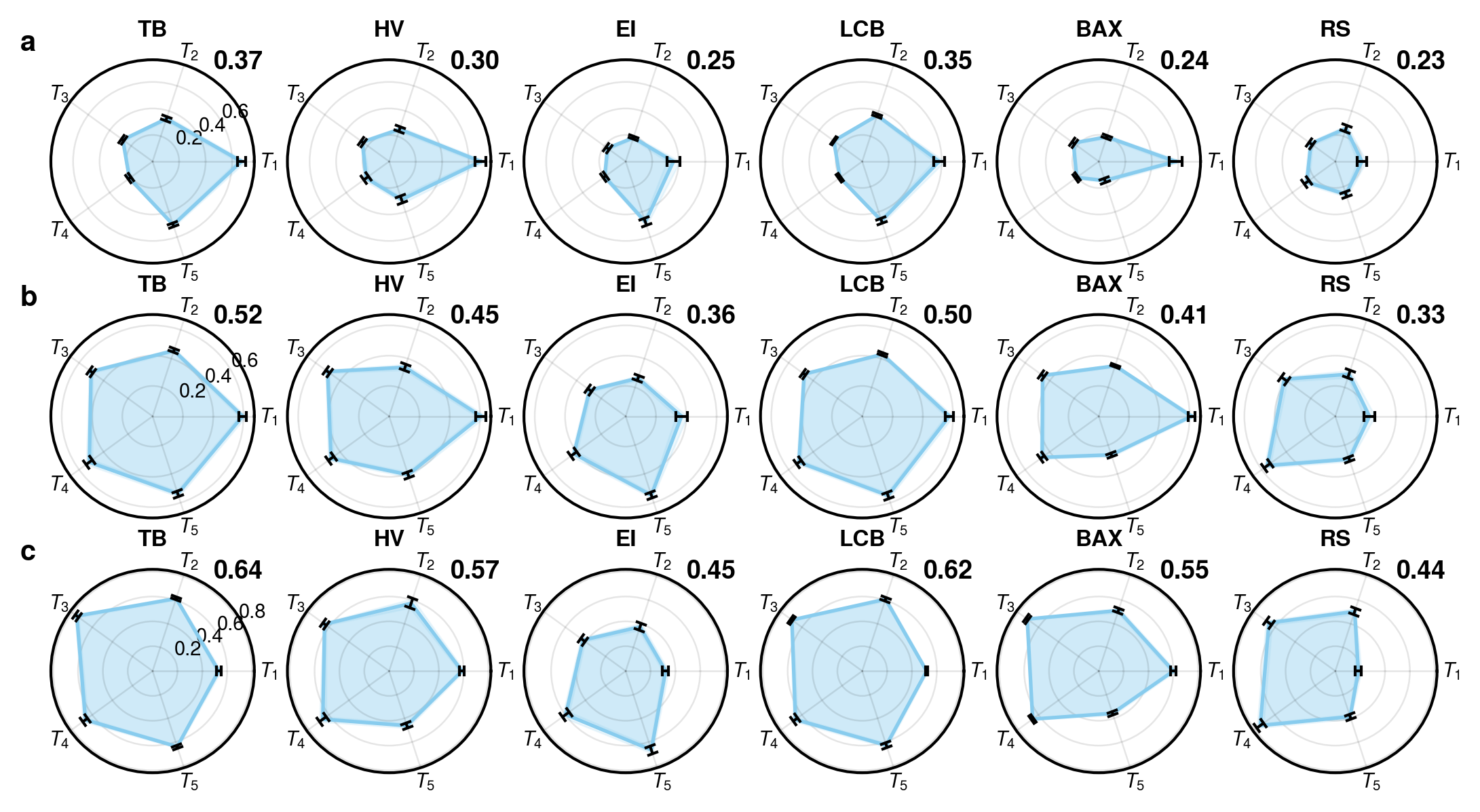}
    \caption{\textbf{Normalized diversity score across acquisition functions for the ESOL benchmark.}
    Radar plots of the normalized diversity score $D_\mathrm{d}$ for five targets $T_1$ through $T_5$ on the ESOL benchmark at tolerance ratios (a) $r = 0.2$, (b) $r = 0.4$, and (c) $r = 0.6$. }
    \label{fig:si-radar-esol}
\end{figure}

 \begin{figure}[p]
    \centering
    \includegraphics[width=\linewidth]{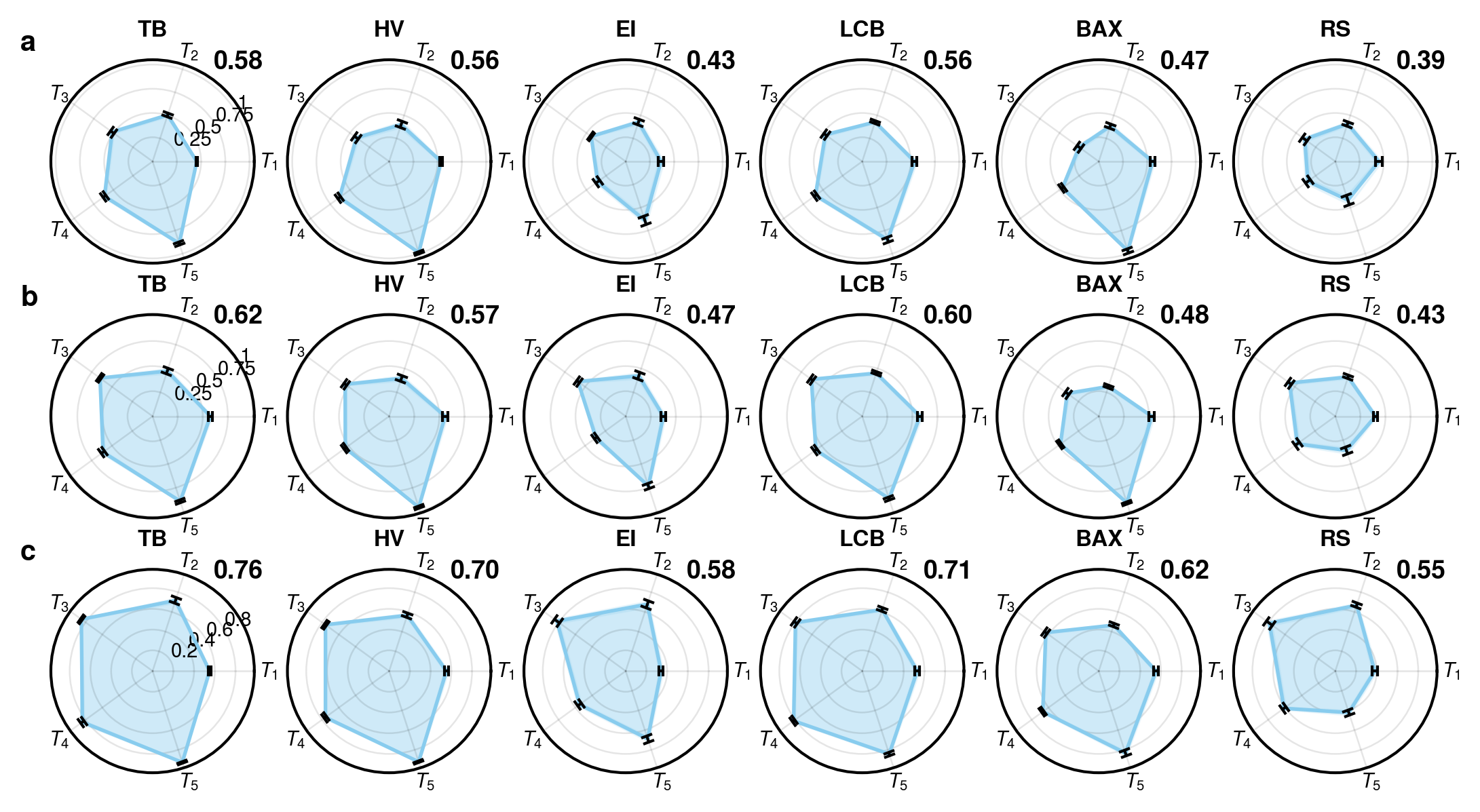}
    \caption{\textbf{Normalized diversity score across acquisition functions for the Freesolv benchmark.}
    Radar plots of the normalized diversity score $D_\mathrm{d}$ for five targets $T_1$ through $T_5$ on the Freesolv benchmark at tolerance ratios (a) $r = 0.2$, (b) $r = 0.4$, and (c) $r = 0.6$. }
    \label{fig:si-radar-freesolv}
\end{figure}

 \begin{figure}[p]
    \centering
    \includegraphics[width=\linewidth]{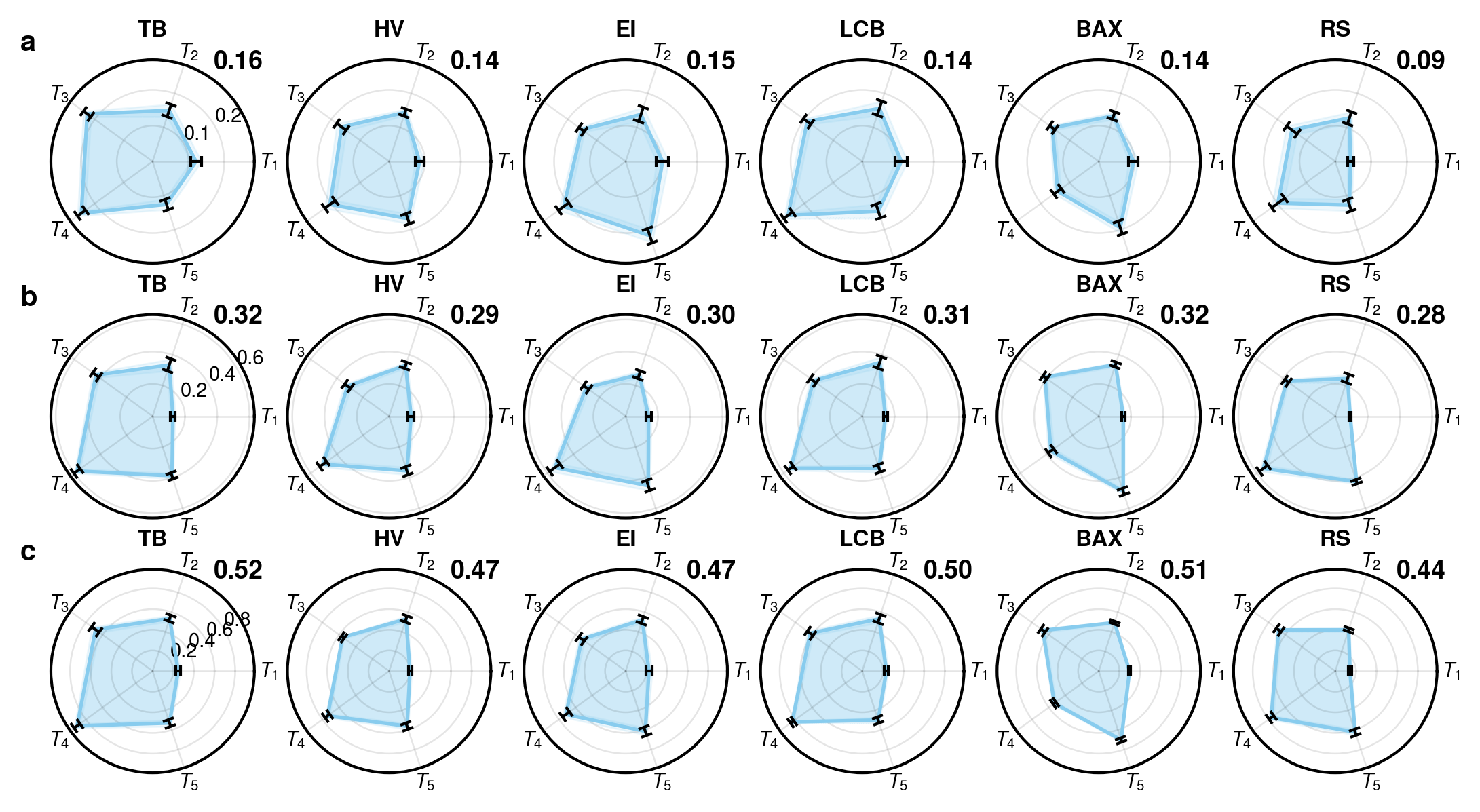}
    \caption{\textbf{Normalized diversity score across acquisition functions for the Lipo benchmark.}
    Radar plots of the normalized diversity score $D_\mathrm{d}$ for five targets $T_1$ through $T_5$ on the Lipo benchmark at tolerance ratios (a) $r = 0.2$, (b) $r = 0.4$, and (c) $r = 0.6$. }
    \label{fig:si-radar-lipo}
\end{figure}

 \begin{figure}[p]
    \centering
    \includegraphics[width=\linewidth]{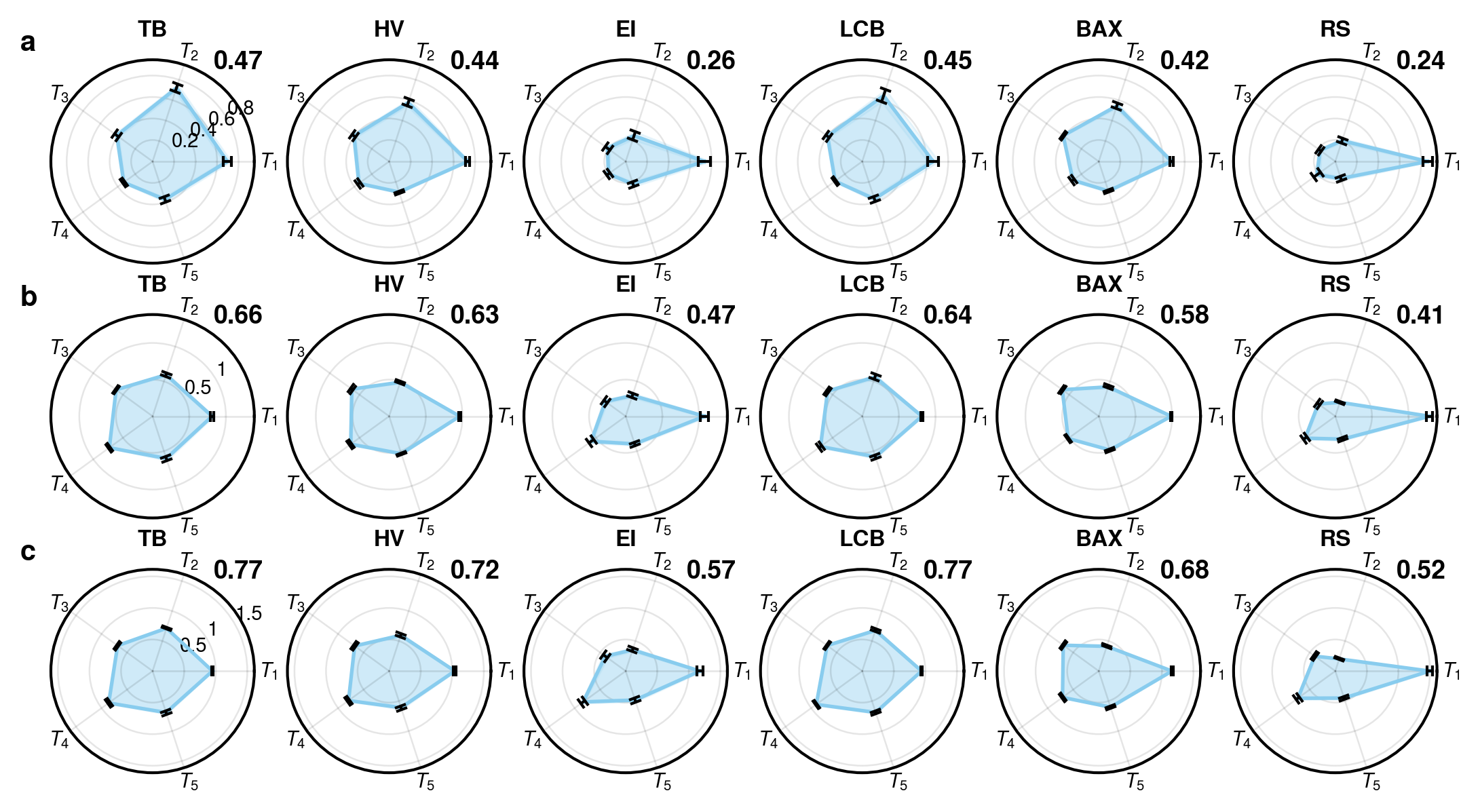}
    \caption{\textbf{Normalized diversity score across acquisition functions for the Toporg benchmark.}
    Radar plots of the normalized diversity score $D_\mathrm{d}$ for five targets $T_1$ through $T_5$ on the Toporg benchmark at tolerance ratios (a) $r = 0.2$, (b) $r = 0.4$, and (c) $r = 0.6$. }
    \label{fig:si-radar-toporg}
\end{figure}
 
\begin{figure}[p]
    \centering
    \includegraphics[width=\linewidth]{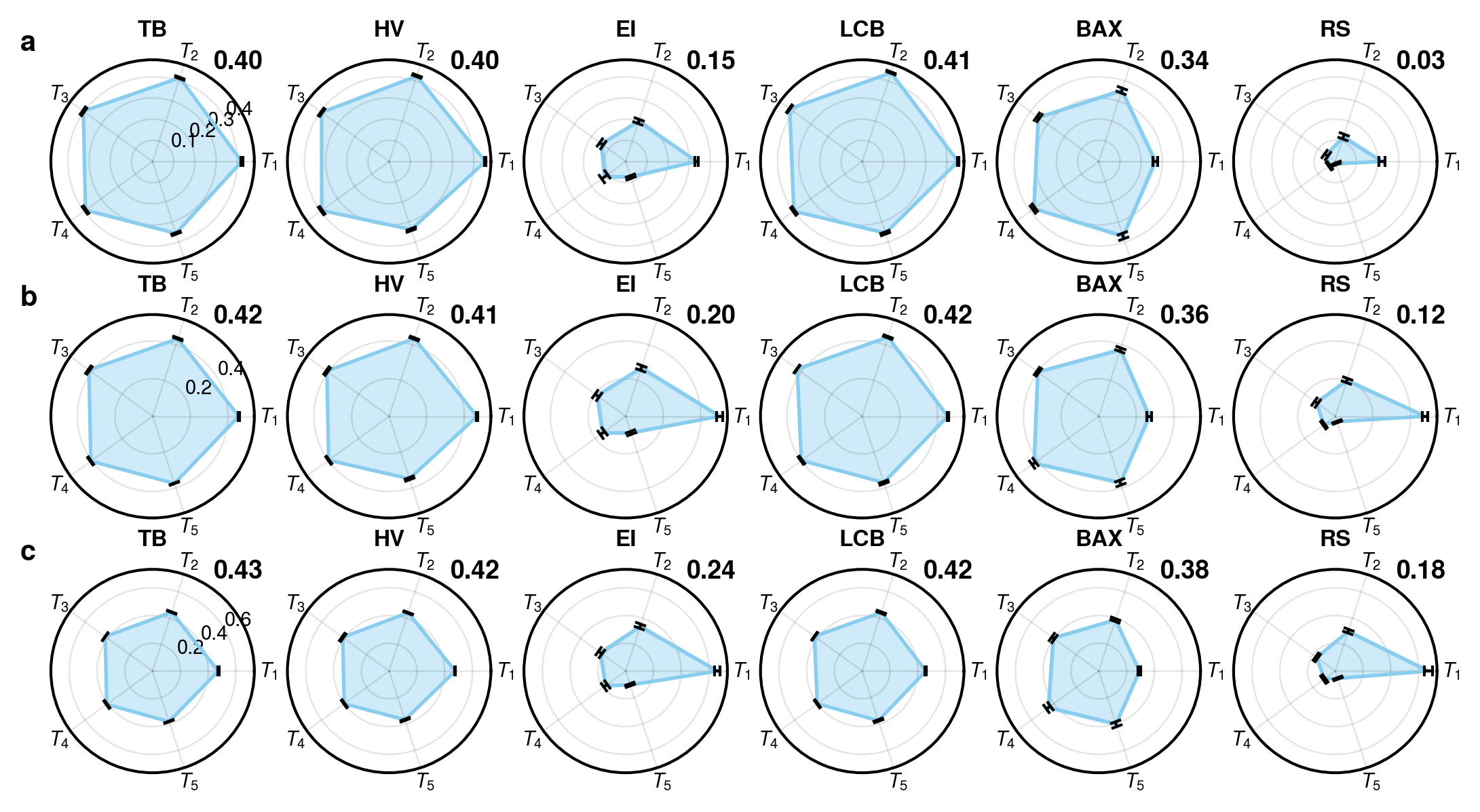}
    \caption{\textbf{Normalized diversity score across acquisition functions for the Branin benchmark.}
    Radar plots of the normalized diversity score $D_\mathrm{c}$ for five targets $T_1$ through $T_5$ on the Branin benchmark at tolerance ratios (a) $r = 0.2$, (b) $r = 0.4$, and (c) $r = 0.6$. }
    \label{fig:si-radar-branin}
\end{figure}
 
\begin{figure}[p]
    \centering
    \includegraphics[width=\linewidth]{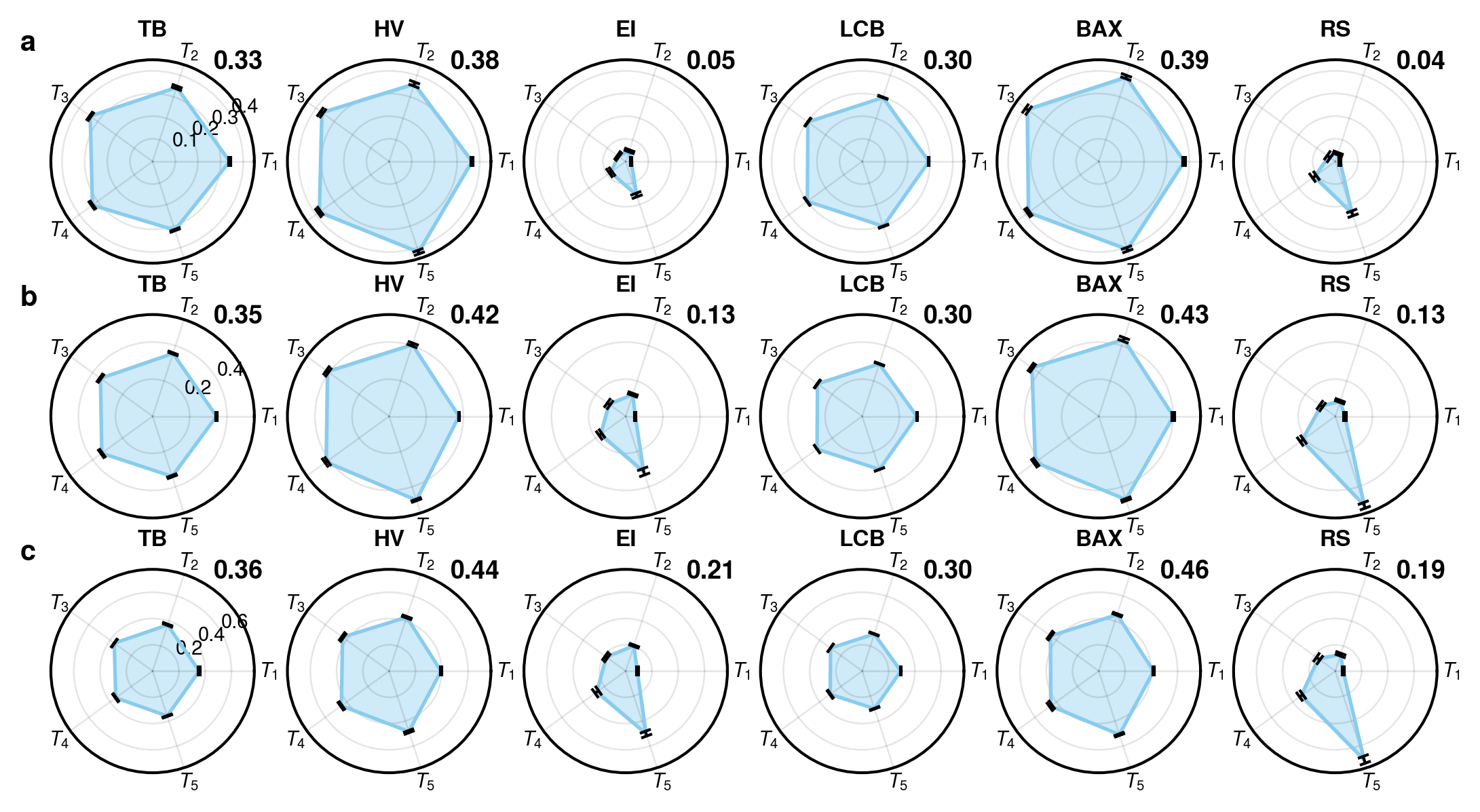}
    \caption{\textbf{Normalized diversity score across acquisition functions for the Hartmann benchmark.}
    Radar plots of the normalized diversity score $D_\mathrm{c}$ for five targets $T_1$ through $T_5$ on the Hartmann benchmark at tolerance ratios (a) $r = 0.2$, (b) $r = 0.4$, and (c) $r = 0.6$. }
    \label{fig:si-radar-hartmann}
\end{figure}

 \begin{figure}[p]
    \centering
    \includegraphics[width=\linewidth]{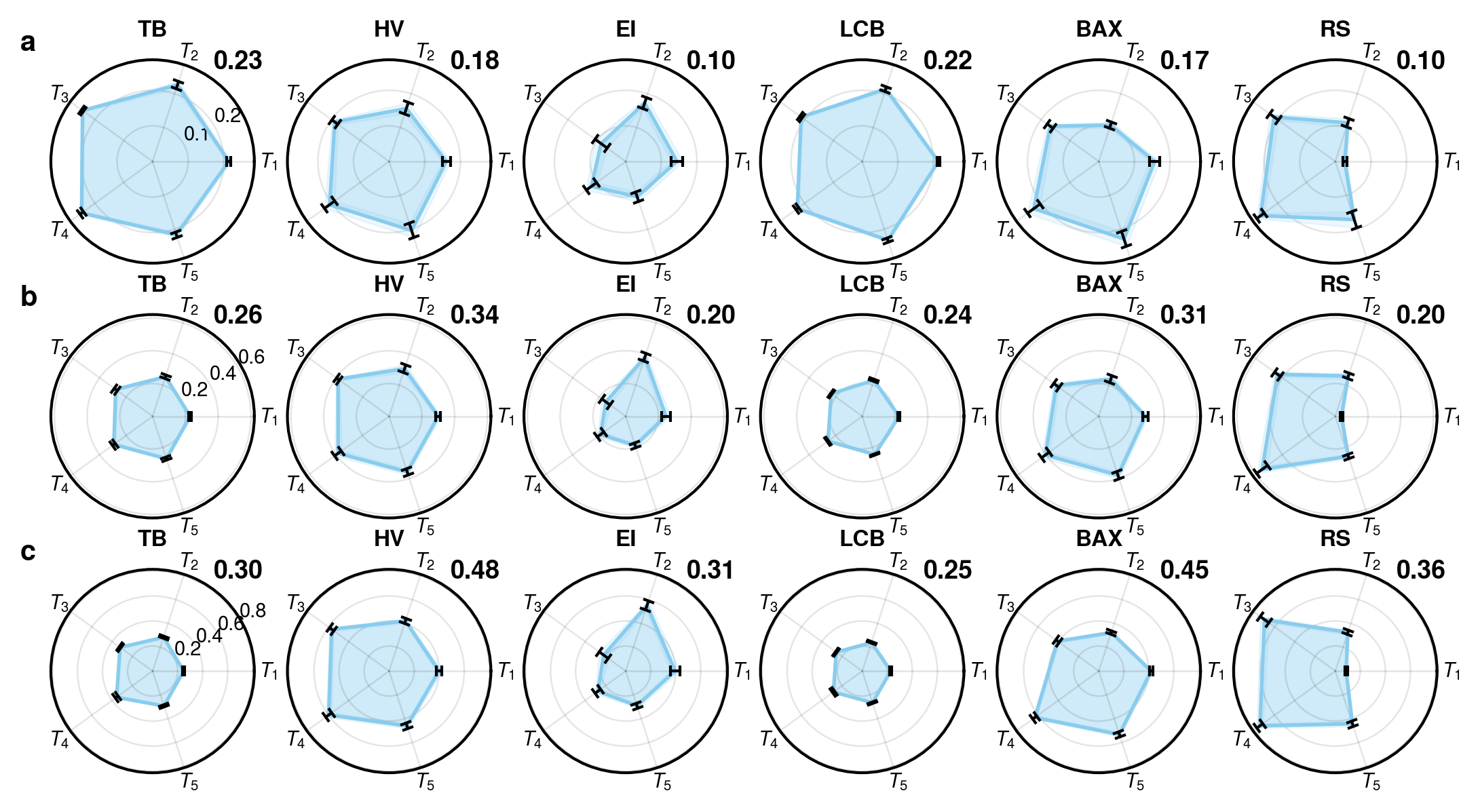}
    \caption{\textbf{Normalized diversity score across acquisition functions for the Ackley benchmark.}
    Radar plots of the normalized diversity score $D_\mathrm{c}$ for five targets $T_1$ through $T_5$ on the Ackley benchmark at tolerance ratios (a) $r = 0.2$, (b) $r = 0.4$, and (c) $r = 0.6$. }
    \label{fig:si-radar-ackley}
\end{figure}

 \begin{figure}[p]
    \centering
    \includegraphics[width=\linewidth]{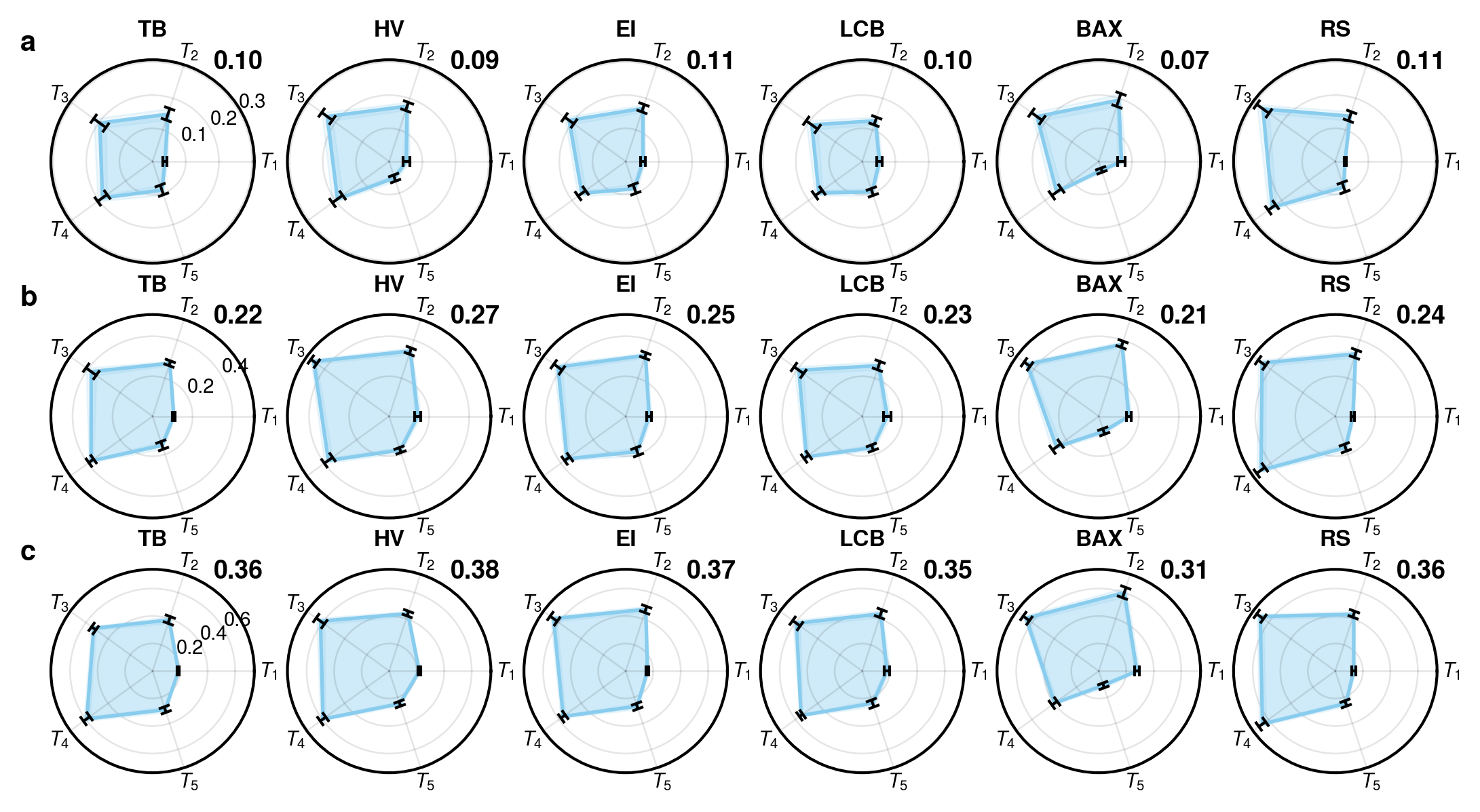}
    \caption{\textbf{Normalized diversity score across acquisition functions for the Layeb06 benchmark.}
    Radar plots of the normalized diversity score $D_\mathrm{c}$ for five targets $T_1$ through $T_5$ on the Layeb06 benchmark at tolerance ratios (a) $r = 0.2$, (b) $r = 0.4$, and (c) $r = 0.6$. }
    \label{fig:si-radar-layeb06}
\end{figure}

\newpage
\section{Kinetic Monte Carlo Styrene Polymerization}
 
\sloppy The kinetic model used in this work is a simplified bulk representation of a sequential controlled/conventional styrene radical polymerization process. Although the process design is inspired by the bimodal molecular weight distribution strategy of Lenzi et al.\cite{lenzi2005producing}, the present simulations do not attempt to reproduce the original miniemulsion or suspension system based on 2,2,6,6-tetramethylpiperidin-1-oxyl (TEMPO), benzoyl peroxide (BPO), and potassium persulfate (KPS). Instead, the controlled stage is modeled using literature kinetics for styrene nitroxide-mediated polymerization (NMP) with the SG1-based alkoxyamine initiator MAMA-SG1 (N-(2-methylpropyl)\allowbreak-N-(1-diethylphosphono\allowbreak-2,2-dimethylpropyl)\allowbreak-O-(2-carboxyprop-2-yl)\allowbreak-hydroxylamine), whereas the conventional stage is modeled using literature kinetics for 2,2'-azobis(2-methylpropionitrile) (AIBN)-initiated styrene free-radical polymerization (FRP).~\cite{lenzi2005producing,fierens2015mama}
 
The physicochemical simulation parameters for both polymerization stages are listed in Table~\ref{tab:styrene_setup}. For computational tractability, both stages were modeled as batch, bulk, and isothermal processes. A constant average termination rate coefficient was used instead of conversion- and chain-length-dependent apparent rate coefficients. Additional assumptions include (i) a constant initiator efficiency in the FRP stage, (ii) only linear polymer chains as products, and (iii) neglect of styrene thermal self-initiation in the NMP stage.
 The present KMC formulation also neglects compartmentalization, particle nucleation, and interphase transport effects associated with the miniemulsion or suspension polymerization system used by Lenzi et al.\cite{lenzi2005producing}. It should therefore be interpreted as a homogeneous kinetic surrogate rather than a reactor-resolved model.
 
\begin{table}[htbp]
  \centering
  \caption{\textbf{Physicochemical parameters and simulation settings for the stylized two-stage styrene polymerization model.}}
  \label{tab:styrene_setup}
  \small
  \setlength{\tabcolsep}{5pt}
  \renewcommand{\arraystretch}{1.15}
  \begin{tabularx}{\textwidth}{@{}
    >{\raggedright\arraybackslash}X
    >{\centering\arraybackslash}p{1.8cm}
    >{\raggedright\arraybackslash}p{4.2cm}
    >{\raggedright\arraybackslash}p{4.2cm}
    @{}}
    \toprule
    \textbf{Parameter} & \textbf{Symbol} & \textbf{NMP stage} & \textbf{FRP stage} \\
    \midrule
    Monomer & ---
      & Styrene
      & Styrene \\
 
    Initiator / controller & ---
      & MAMA-SG1
      & AIBN \\
 
    Monomer molar mass & $M_\mathrm{M}$
      & \SI{104.15}{\gram\per\mole}
      & \SI{104.15}{\gram\per\mole} \\
 
    Initiator molar mass & $M_\mathrm{I}$
      & \SI{381.45}{\gram\per\mole}
      & \SI{164.21}{\gram\per\mole} \\
 
    Monomer density & $\rho_\mathrm{M}$
      & \SI{859.2}{\gram\per\liter}
      & \SI{859.2}{\gram\per\liter} \\
 
    Polymer density & $\rho_\mathrm{P}$
      & \SI{969.2}{\gram\per\liter}
      & \SI{969.2}{\gram\per\liter} \\
 
    Initial reactor volume & $V_0$
      & \SI{8.86e-16}{\liter}
      & inherited from NMP stage \\
 
    Initial monomer concentration & $C_\mathrm{M,0}$
      & \SI{8.56}{\mole\per\liter}
      & inherited from NMP stage \\
 
    Initial initiator concentration & $C_{\mathrm{init}}$
      & input
      & input \\
 
    Stage temperature & $T$
      & input
      & input \\
 
    Stage-end conversion & $X$
      & input
      & fixed at \SI{95}{\percent} \\
 
    Initiator efficiency & $f$
      & ---
      & \num{0.50} \\
    \bottomrule
  \end{tabularx}
\end{table}
 
\begin{table}[htbp]
  \centering
  \caption{\textbf{Kinetic parameters and reaction channels for the effective SG1-alkoxyamine-mediated NMP stage of styrene.}}
  \label{tab:nmp_kinetics}
  \small
  \setlength{\tabcolsep}{5pt}
  \renewcommand{\arraystretch}{1.15}
  \begin{tabularx}{\textwidth}{@{}
    >{\raggedright\arraybackslash}X
    >{\raggedright\arraybackslash}X
    S[table-format=1.2e2]
    >{\centering\arraybackslash}p{1.8cm}
    >{\raggedright\arraybackslash}p{2.6cm}
    @{}}
    \toprule
    \textbf{Reaction channel} & \textbf{Elementary step} & \multicolumn{1}{c}{\textbf{$A$}} & \textbf{$E_a$} & \textbf{Reference} \\
    & & \multicolumn{1}{c}{([\si{\liter\per\mole}]\,\si{\per\second})} & (\si{\kilo\joule\per\mole}) & \\
    \midrule
    Activation of initiator alkoxyamine
      & $AX_0 \rightarrow R_{0,1} + X$
      & \num{1.16e13}
      & 105.3
      & \cite{fierens2015mama} \\
 
    Activation of dormant macroradical
      & $P_iX \rightarrow R_i + X$
      & \num{4.04e17}
      & 148.7
      & \cite{fierens2015mama} \\
 
    Deactivation of initiator radical
      & $R_{0,1} + X \rightarrow AX_0$
      & \num{2.80e6}
      & ---
      & \cite{fierens2015mama} \\
 
    Deactivation of propagating radical
      & $R_i + X \rightarrow P_iX$
      & \num{1.09e6}
      & ---
      & \cite{fierens2015mama} \\
 
    Chain initiation by fragment type 1
      & $R_{0,1} + M \rightarrow R_1$
      & \num{1.55e6}
      & 16.5
      & \cite{chauvin2006nitroxide} \\
 
    Chain initiation by fragment type 2
      & $R_{0,2} + M \rightarrow R_1$
      & \num{4.24e7}
      & 32.5
      & \cite{buback1995critically} \\
 
    Propagation
      & $R_i + M \rightarrow R_{i+1}$
      & \num{4.24e7}
      & 32.5
      & \cite{buback1995critically} \\
 
    Chain transfer to monomer
      & $R_i + M \rightarrow P_i + R_{0,2}$
      & \num{2.30e6}
      & 53.0
      & \cite{hui1972thermal} \\
 
    Termination by recombination
      & $R_i + R_j \rightarrow P_{i+j}$
      & \num{1.00e8}
      & ---
      & \cite{de2020roadmap} \\
    \bottomrule
  \end{tabularx}
\end{table}
 
\begin{table}[htbp]
  \centering
  \caption{\textbf{Kinetic parameters and reaction channels for the AIBN-initiated free-radical polymerization (FRP) stage of styrene.}}
  \label{tab:frp_kinetics}
  \small
  \setlength{\tabcolsep}{5pt}
  \renewcommand{\arraystretch}{1.15}
  \begin{tabularx}{\textwidth}{@{}
    >{\raggedright\arraybackslash}X
    >{\raggedright\arraybackslash}X
    S[table-format=1.2e2]
    >{\centering\arraybackslash}p{1.8cm}
    >{\raggedright\arraybackslash}p{2.6cm}
    @{}}
    \toprule
    \textbf{Reaction channel} & \textbf{Elementary step} & \multicolumn{1}{c}{\textbf{$A$}} & \textbf{$E_a$} & \textbf{Reference} \\
    & & \multicolumn{1}{c}{([\si{\liter\per\mole}]\,\si{\per\second})} & (\si{\kilo\joule\per\mole}) & \\
    \midrule
    Initiator dissociation
      & $I_2 \rightarrow 2I$
      & \num{5.88e15}
      & 133.0
      & \cite{achilias1992development} \\
 
    Chain initiation\textsuperscript{a}
      & $I + M \rightarrow R_1$
      & \num{4.27e7}
      & 32.5
      & \cite{gilbert1996critically} \\
 
    Propagation
      & $R_i + M \rightarrow R_{i+1}$
      & \num{4.27e7}
      & 32.5
      & \cite{gilbert1996critically} \\
 
    Termination by disproportionation
      & $R_i + R_j \rightarrow P_i + P_j$
      & \num{6.70e9}
      & 15.6
      & \cite{buback1997termination,hatada1985evidence,yamada1992esr} \\
 
    Termination by recombination
      & $R_i + R_j \rightarrow P_{i+j}$
      & \num{2.00e10}
      & 15.6
      & \cite{buback1997termination,hatada1985evidence,yamada1992esr} \\
 
    Chain transfer to monomer
      & $R_i + M \rightarrow P_i + R_1$
      & \num{9.38e6}
      & 54.11
      & \cite{kapfenstein2001novel,gilbert1996critically} \\
    \bottomrule
  \end{tabularx}
 
  \vspace{0.35em}
  \raggedright
  \footnotesize
  \textsuperscript{a} The chain-initiation coefficient was approximated by the styrene propagation coefficient, $k_i \approx k_p$, as a coarse-grained modeling assumption for primary-radical addition.
\end{table}

The five target MWDs were obtained by $k$-medoids clustering of the 100-dimensional weight-fraction vectors across the 51,200 design conditions. The tolerance used in the main text is $\varepsilon = 0.0444$ in the 100D weight-fraction space.

\begin{figure}[h!]
    \centering
    \includegraphics[width=.5\linewidth]{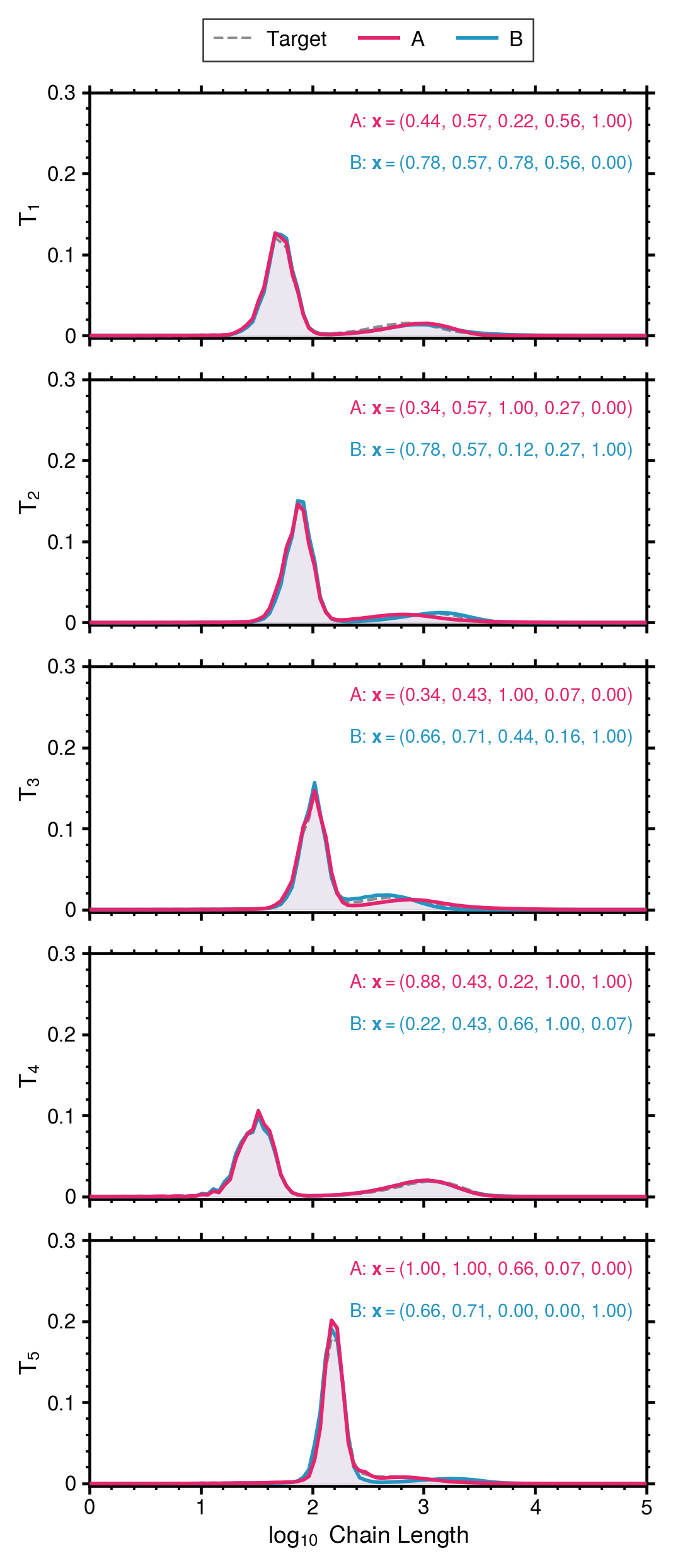}
    \caption{\textbf{Degeneracy in reaction-condition space.} For each target $T_1$ through $T_5$, two TB-discovered reaction designs that are separated in the design space yet yield similar MWDs. The annotated vectors $\mathbf{x}$ give the five design variables min-max normalized to $[0,1]$; their physical values and units are listed in Table~\ref{tab:kmc-pairs}, and the variables themselves are defined in Table~\ref{tab:styrene_setup}.}
    \label{fig:si-polymerization}
\end{figure}

\begin{table}[htbp]
  \centering
  \caption{\textbf{Physical reaction conditions for the degenerate design pairs of Fig.~\ref{fig:si-polymerization}.}}
  \label{tab:kmc-pairs}
  \small
  \renewcommand{\arraystretch}{1.2}
  \begin{tabularx}{\textwidth}{@{}
    >{\centering\arraybackslash}p{1.1cm}
    >{\centering\arraybackslash}p{1.1cm}
    >{\centering\arraybackslash}X
    >{\centering\arraybackslash}X
    >{\centering\arraybackslash}X
    >{\centering\arraybackslash}X
    >{\centering\arraybackslash}X
    @{}}
    \toprule
    \textbf{Target} & \textbf{Design} & {$T_{\mathrm{NMP}}$ (K)} & {$X_{\mathrm{NMP}}$} & {$T_{\mathrm{FRP}}$ (K)} & {$C_{\mathrm{init,NMP}}$ (mol/L)} & {$C_{\mathrm{init,FRP}}$ (mol/L)} \\
    \midrule
    $T_1$ & A & 375 & 0.63 & 324 & 0.115 & 0.0171 \\
          & B & 392 & 0.63 & 352 & 0.115 & 0.0043 \\
    \addlinespace
    $T_2$ & A & 370 & 0.63 & 363 & 0.078 & 0.0043 \\
          & B & 392 & 0.63 & 319 & 0.078 & 0.0171 \\
    \addlinespace
    $T_3$ & A & 370 & 0.57 & 363 & 0.052 & 0.0043 \\
          & B & 386 & 0.69 & 335 & 0.064 & 0.0171 \\
    \addlinespace
    $T_4$ & A & 397 & 0.57 & 324 & 0.171 & 0.0171 \\
          & B & 364 & 0.57 & 346 & 0.171 & 0.0052 \\
    \addlinespace
    $T_5$ & A & 403 & 0.80 & 346 & 0.052 & 0.0043 \\
          & B & 386 & 0.69 & 313 & 0.043 & 0.0171 \\
    \bottomrule
  \end{tabularx}
\end{table}

\newpage
\section{Sequence-Defined Conjugated Oligomers}

\begin{figure}[h!]
    \centering
    \includegraphics[width=\linewidth]{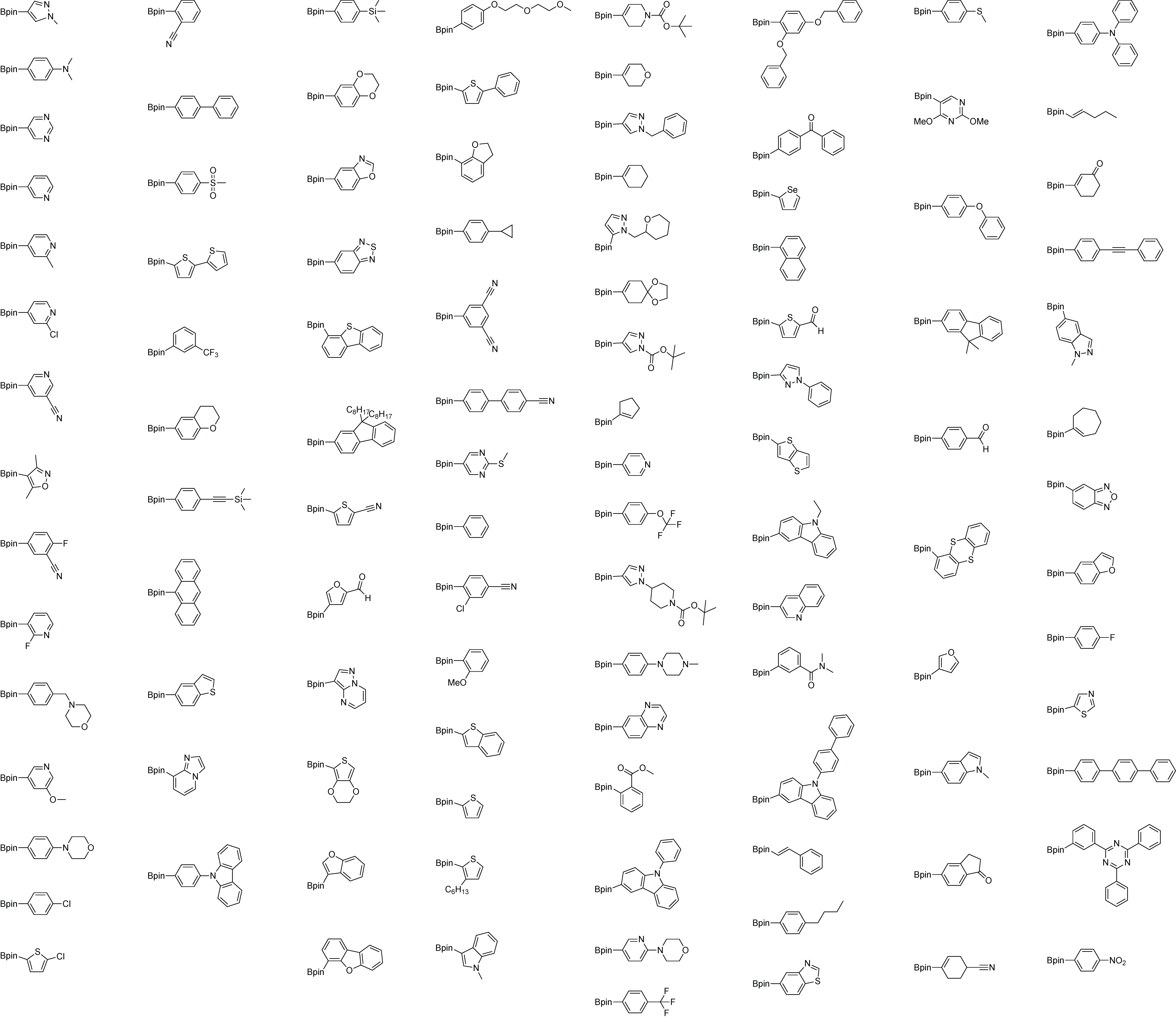}
    \caption{\textbf{Terminal caps.} The 100 terminal-cap fragments, derived from commercially available pinacol boronic esters, that were appended to each of the 21 conjugated cores to enumerate the oligomer library.}
    \label{fig:si-terminal_caps}
\end{figure}

\begin{figure}[h!]
    \centering
    \includegraphics[width=.5\linewidth]{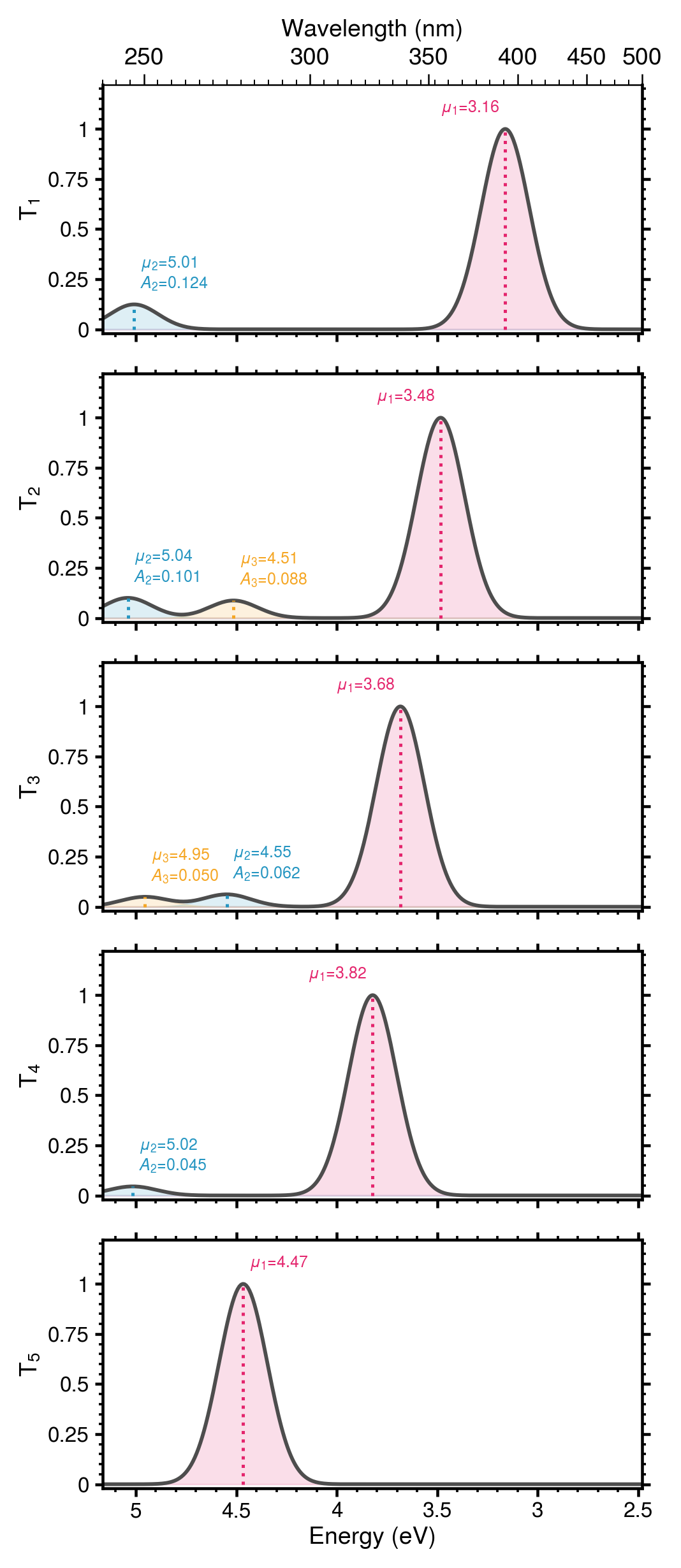}
    \caption{\textbf{Target absorption profiles.} Reconstructed spectra of the five targets $T_1$ through $T_5$ on the energy axis, with the corresponding wavelength on the top axis. Shaded Gaussian components and dotted lines mark the fitted band centers $\mu_k$ and amplitudes $A_k$ (Table~\ref{tab:oligomer-targets}). Targets are ordered by increasing dominant-band energy, from 3.16~eV ($T_1$) to 4.47~eV ($T_5$).}
    \label{fig:si-tddft-targets}
\end{figure}

The library pairs 21 conjugated cores with 100 terminal caps derived from commercially available pinacol boronic esters (Fig.~\ref{fig:si-terminal_caps}). Removing duplicate structures leaves 1{,}980 unique oligomers. The five target absorption profiles are shown in Fig.~\ref{fig:si-tddft-targets}, with their band parameters in Table~\ref{tab:oligomer-targets}.

\begin{table}[htbp]
  \centering
  \caption{\textbf{Band parameters of the five absorption targets.} The dominant band has unit amplitude ($A_1 \equiv 1$, fixed during fitting); secondary bands $G_2$--$G_5$ are given as amplitude at center energy, with a dash where the band is absent.  The final column is the group-wise scaled vector $(\mu_1, A_2, \mu_2, A_3, \mu_3, A_4, \mu_4, A_5, \mu_5)$ in $[0,1]^9$ optimized by the surrogate; absent bands take the fixed canonical center $0.324$. Scaling is group-wise rather than per dimension, where all band energies share one min--max ($\mu\in[2.48,5.17]$~eV) and all amplitudes share another. Because the dominant band is fixed at unit amplitude during fitting, the amplitude group already spans $[0,1]$, so each scaled amplitude equals its raw value (e.g., the $T_1$ secondary amplitude is $0.12$ in both columns); only the band energies are numerically rescaled.}
  \label{tab:oligomer-targets}
  \footnotesize
  \renewcommand{\arraystretch}{1.1}
  \setlength{\tabcolsep}{3pt}
  \begin{tabularx}{\textwidth}{@{}
    >{\centering\arraybackslash}p{0.8cm}
    >{\centering\arraybackslash}p{1.7cm}
    >{\centering\arraybackslash}p{2.1cm}
    >{\centering\arraybackslash}p{2.1cm}
    >{\centering\arraybackslash}p{0.9cm}
    >{\centering\arraybackslash}p{0.9cm}
    >{\raggedright\arraybackslash}X
    @{}}
    \toprule
    \textbf{Target} & \textbf{$G_1$ $\mu_1$ (eV)} & \textbf{$G_2$ ($A_2$ @ $\mu_2$)} & \textbf{$G_3$ ($A_3$ @ $\mu_3$)} & \textbf{$G_4$} & \textbf{$G_5$} & \textbf{Scaled 9D vector} \\
    \midrule
    $T_1$ & 3.16 & 0.12 @ 5.01 & --- & --- & --- &
      \scriptsize (0.254, 0.124, 0.941, 0, 0.324, 0, 0.324, 0, 0.324) \\
    $T_2$ & 3.48 & 0.10 @ 5.04 & 0.09 @ 4.51 & --- & --- &
      \scriptsize (0.373, 0.101, 0.953, 0.088, 0.757, 0, 0.324, 0, 0.324) \\
    $T_3$ & 3.68 & 0.06 @ 4.55 & 0.05 @ 4.95 & --- & --- &
      \scriptsize (0.448, 0.062, 0.769, 0.050, 0.921, 0, 0.324, 0, 0.324) \\
    $T_4$ & 3.82 & 0.05 @ 5.02 & --- & --- & --- &
      \scriptsize (0.499, 0.045, 0.944, 0, 0.324, 0, 0.324, 0, 0.324) \\
    $T_5$ & 4.47 & --- & --- & --- & --- &
      \scriptsize (0.739, 0, 0.324, 0, 0.324, 0, 0.324, 0, 0.324) \\
    \bottomrule
  \end{tabularx}

  \vspace{0.3em}
  \raggedright\footnotesize
   The tolerance used in the main text is $\varepsilon = 0.0774$ in the group-minmax scaled $[0,1]^5$ space.
\end{table}

\begin{figure}[h!]
    \centering
    \includegraphics[width=\linewidth]{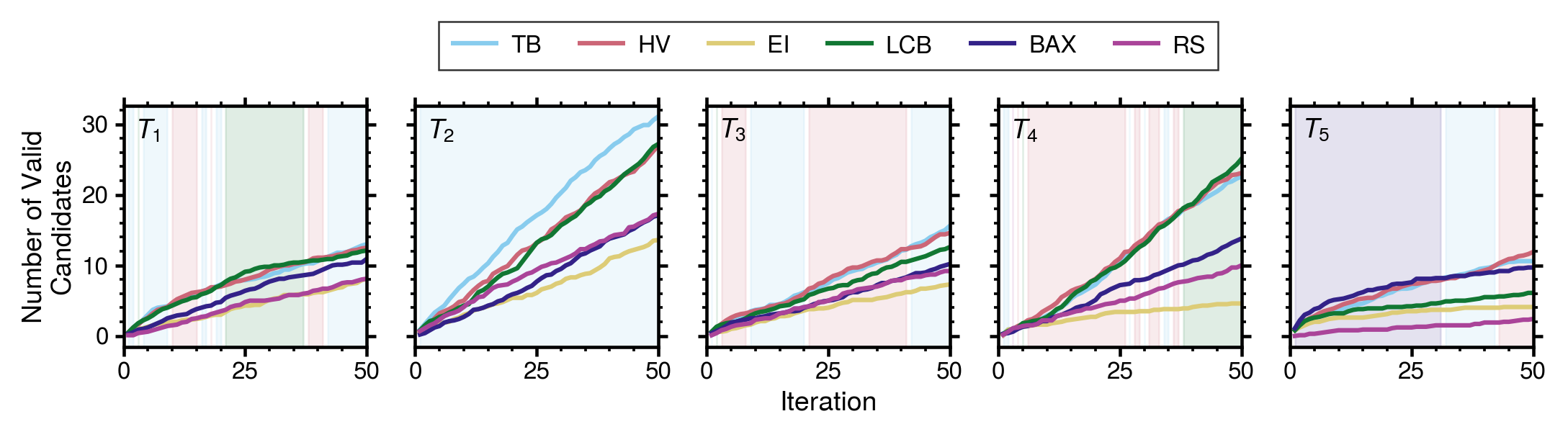}
    \caption{\textbf{Target-resolved discovery trajectories for the sequence-defined oligomer library.} Cumulative number of valid oligomers identified as a function of Bayesian optimization iteration for each target $T_1$ through $T_5$, comparing the six acquisition functions. The shaded background in each panel indicates the leading acquisition function over the corresponding iteration range.}
    \label{fig:si-oligomer-lines}
\end{figure}

\clearpage
\bibliographystyle{unsrt}
\bibliography{ref}